\documentclass[acmtog,authorversion]{acmart}
\acmSubmissionID{xxx}

\usepackage{amsmath}
\usepackage{bbm}
\usepackage{booktabs} 
\usepackage{cancel}
\usepackage[commandnameprefix=always,authormarkup=none,final]{changes}
\usepackage{enumitem}
\usepackage{epsfig}
\usepackage{graphicx}
\usepackage{makecell}
\usepackage{multirow}
\usepackage{subcaption}
\usepackage{xcolor}

\usepackage[ruled]{algorithm2e} 

\SetAlFnt{\small}
\SetAlCapFnt{\small}
\SetAlCapNameFnt{\small}
\SetAlCapHSkip{0pt}

\acmJournal{TOG}

\usepackage{amsfonts}
\usepackage{textcomp}

\usepackage[capitalize]{cleveref}
\crefname{section}{Sec.}{Secs.}
\Crefname{section}{Section}{Sections}
\Crefname{table}{Table}{Tables}
\crefname{table}{Tab.}{Tabs.}

\usepackage{scalerel}

\definechangesauthor[name=new additions, color=teal]{new}
\definechangesauthor[name=in place modifications, color=blue]{modified}
\definechangesauthor[name=deletions, color=red]{deleted}

\newcommand{\flowerbrackets}[1]{\left\{ #1 \right\}}
\newcommand{\commonbrackets}[1]{\left( #1 \right)}
\newcommand{\indicatorfunc}{\mathbbm{1}}
\newcommand{\stopgradient}{\cancel{\nabla}}
\newcommand{\degree}{{}^\circ}
\newcommand{\lossColor}{{\mathcal{L}_{ph}}}
\newcommand{\lossMainModel}{{\mathcal{L}_m}}
\newcommand{\lossAugModel}{{\mathcal{L}_a}}
\newcommand{\lossSparseDepth}{{\mathcal{L}_{sd}}}
\newcommand{\lossAugDepth}{{\mathcal{L}_{aug}}}
\newcommand{\lossAugSmoothing}{{\mathcal{L}_s}}
\newcommand{\lossAugLambertian}{{\mathcal{L}_l}}
\newcommand{\lossCoarseFineConsistency}{{\mathcal{L}_{cfc}}}
\newcommand{\lossMassConcentration}{{\mathcal{L}_{mc}}}
\newcommand{\lossWeightMainModel}{{\lambda_m}}
\newcommand{\lossWeightAugModel}{{\lambda_a}}
\newcommand{\lossWeightSparseDepth}{{\lambda_{sd}}}
\newcommand{\lossWeightAugDepth}{{\lambda_{aug}}}
\newcommand{\lossWeightCoarseFineConsistency}{{\lambda_{cfc}}}
\newcommand{\lossWeightMassConcentration}{{\lambda_{mc}}}
\newcommand{\numTensorfVoxels}[1][]{{N_{vox}^{#1}}}
\newcommand{\numTensorfAugMassConcentrationBins}{{N_{mc}}}
\newcommand{\pixelq}{\mathbf{q}}
\newcommand{\pointp}{\mathbf{p}}
\newcommand{\viewv}{\mathbf{v}}
\newcommand{\colorc}{\mathbf{c}}
\newcommand{\featureh}{\mathbf{h}}
\newcommand{\fieldf}{\mathcal{F}}
\newcommand{\mlpn}{\mathcal{N}}
\newcommand{\gridg}{\mathcal{G}}
\newcommand{\matrixm}{\mathbf{M}}
\newcommand{\vectora}{\mathbf{a}}
\newcommand{\vectorb}{\mathbf{b}}
\newcommand{\vectorv}{\mathbf{v}}
\newcommand{\hashgridh}{\mathcal{H}}
\newcommand{\realR}{\mathbb{R}}

\newcommand{\figureTeaser}{
    \begin{teaserfigure}
        \centering
        \begin{subfigure}{0.32\linewidth}
            \centering
            \includegraphics[width=\linewidth]{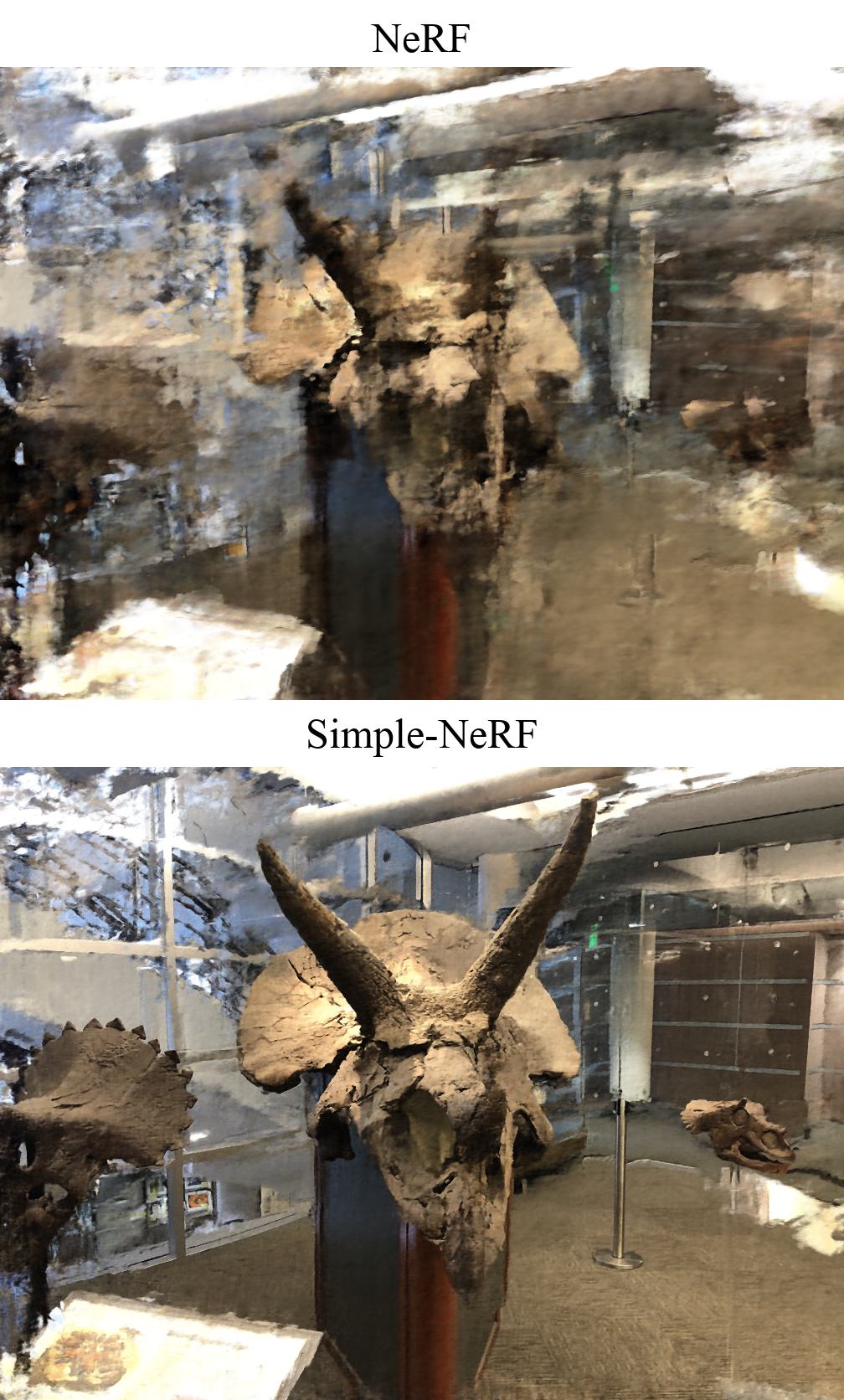}  
            \label{fig:teaser-simple-nerf}
        \end{subfigure}
        \hfill
        \begin{subfigure}{0.32\linewidth}
            \centering
            \includegraphics[width=\linewidth]{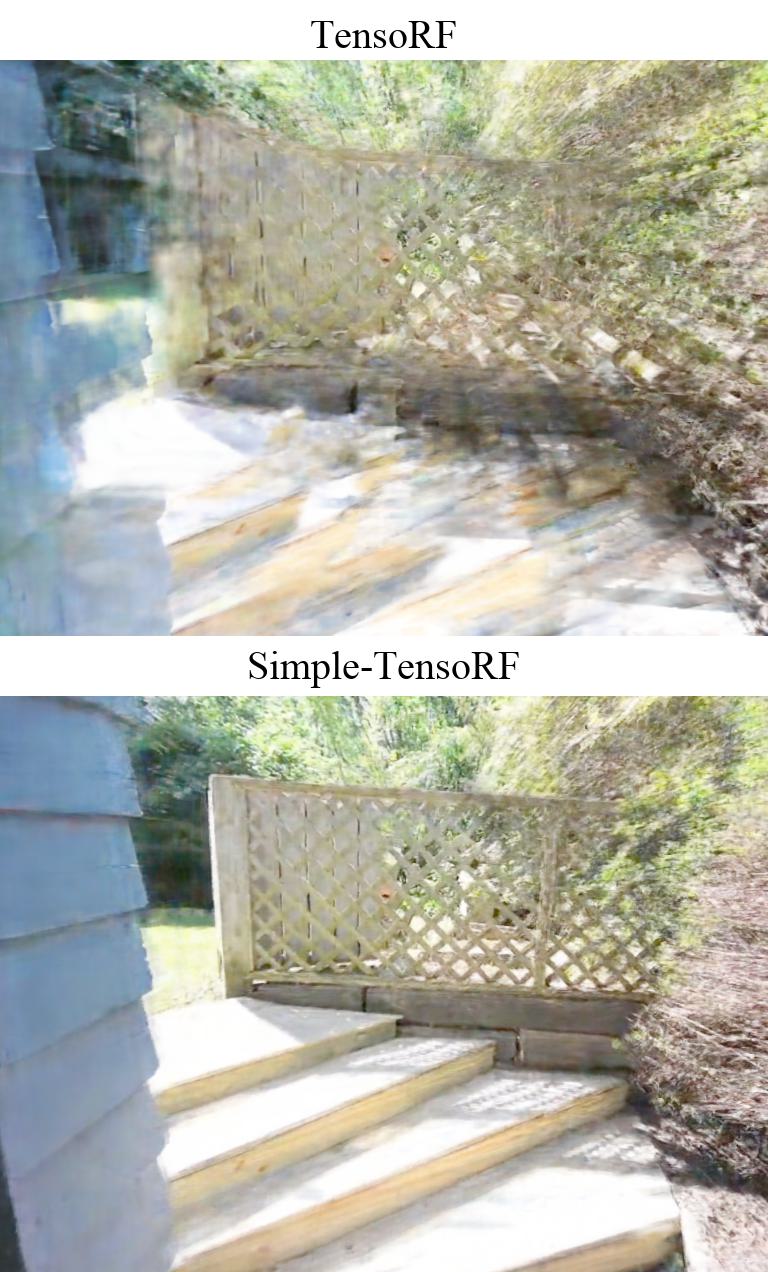}  
            \label{fig:teaser-simple-tensorf}
        \end{subfigure}
        \hfill
        \begin{subfigure}{0.32\linewidth}
            \centering
            \includegraphics[width=\linewidth]{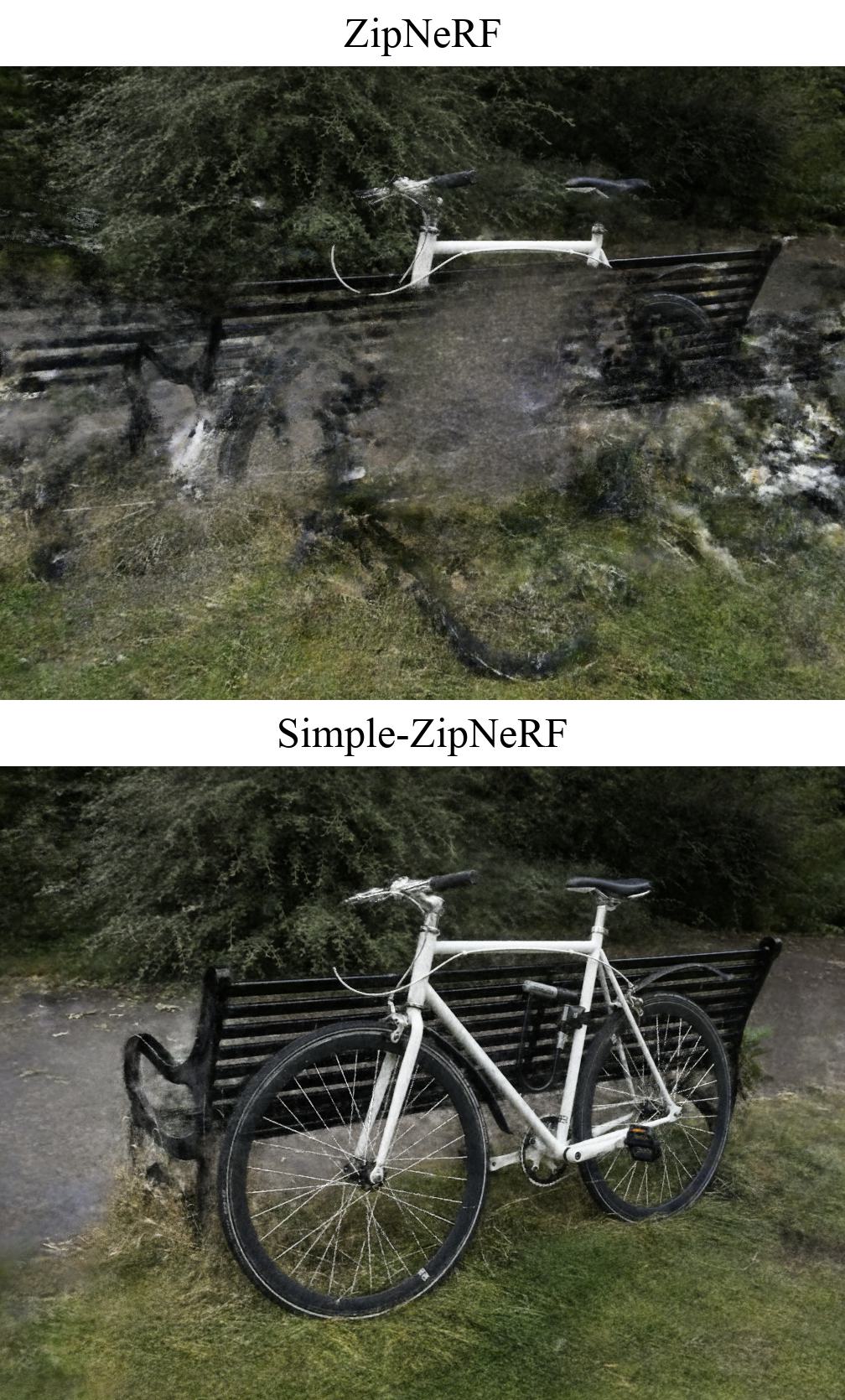}
            \label{fig:teaser-simple-zipnerf}
        \end{subfigure}
        \caption{We show the improvements achieved by our regularizations on the NeRF, TensoRF and ZipNeRF models on NeRF-LLFF, RealEstate-10K and MipNeRF360 datasets respectively.
        We observe that the vanilla radiance fields suffer from various distortions.
        Regularizing the radiance fields with simpler solutions leads to significantly better reconstructions with all the three radiance fields.
        }
        \label{fig:teaser}
    \end{teaserfigure}
}

\newcommand{\figureModelArchitectureSimpleRF}{
    \begin{figure*}
        \includegraphics[width=0.75\linewidth]{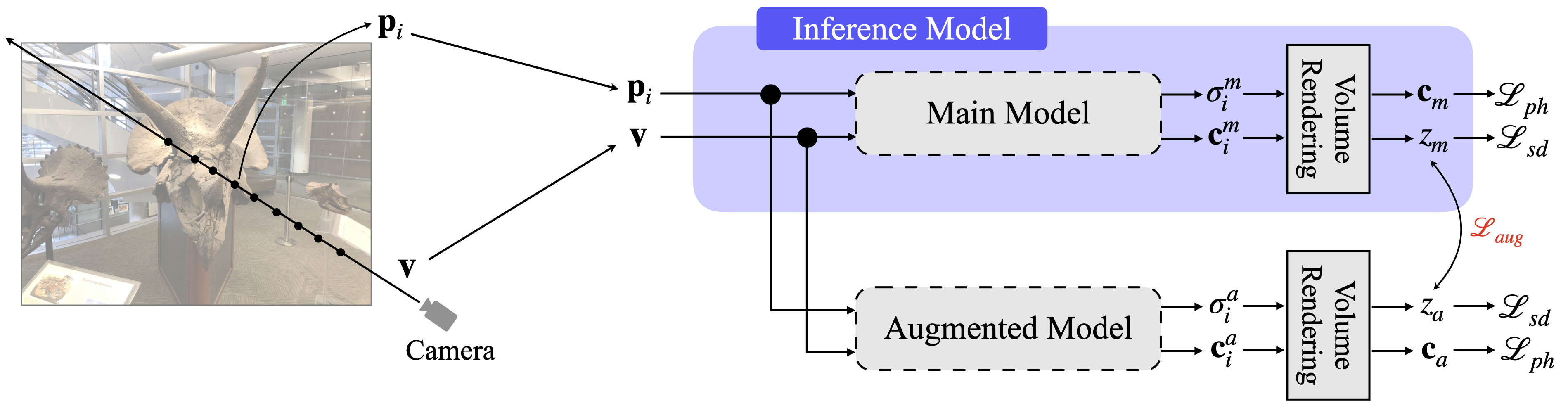}
        \caption{Architecture of Simple-RF family of models.
        We train the augmented model that only learns simpler solutions in tandem with the main model.
        The augmented models learn better depth in certain regions, which is propagated to the main model through the depth supervision loss \textcolor{red}{$\lossAugDepth$}.
        During inference, only the Main Model is employed.
        }
        \label{fig:model-architecture-simple-rf}
    \end{figure*}
}

\newcommand{\figureModelArchitectureSimpleNeRF}{
    \begin{figure*}
        \includegraphics[width=\linewidth]{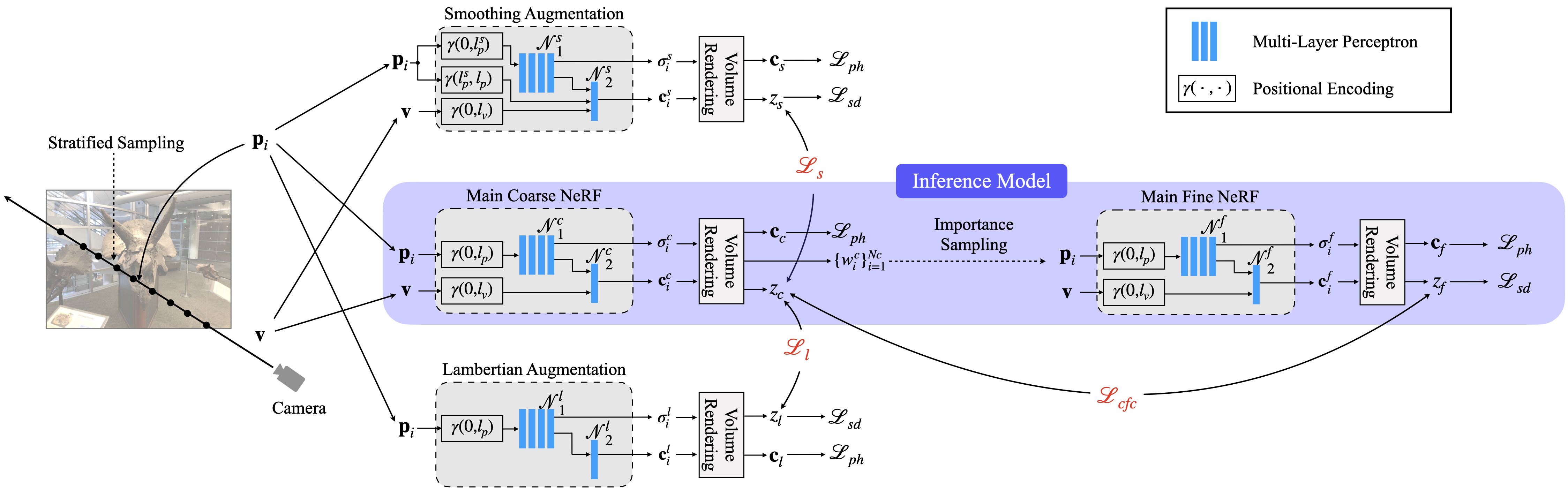}
        \caption{Architecture of Simple-NeRF.
        We train two augmented NeRF models in tandem with the main NeRF.
        In smoothing augmentation, we reduce the positional encoding frequencies that are input to $\mlpn_1^s$ and concatenate the remaining frequencies to the input of $\mlpn_2^s$.
        For Lambertian augmentation, we ask $\mlpn_2^l$ to output the color based on position alone, independent of the viewing direction.
        We add depth supervision losses \textcolor{red}{$\lossAugSmoothing$} and \textcolor{red}{$\lossAugLambertian$} between the coarse NeRFs of the main and augmented models and a consistency loss \textcolor{red}{$\lossCoarseFineConsistency$} between the coarse and fine NeRFs of the main model.
        During inference, only the Main Model is employed.
        }
        \label{fig:model-architecture-simple-nerf}
    \end{figure*}
}

\newcommand{\figureDistortionsSparseInputNeRF}{
    \begin{figure}
        \centering
        \begin{subfigure}{\linewidth}
            \centering
            \includegraphics[width=\linewidth]{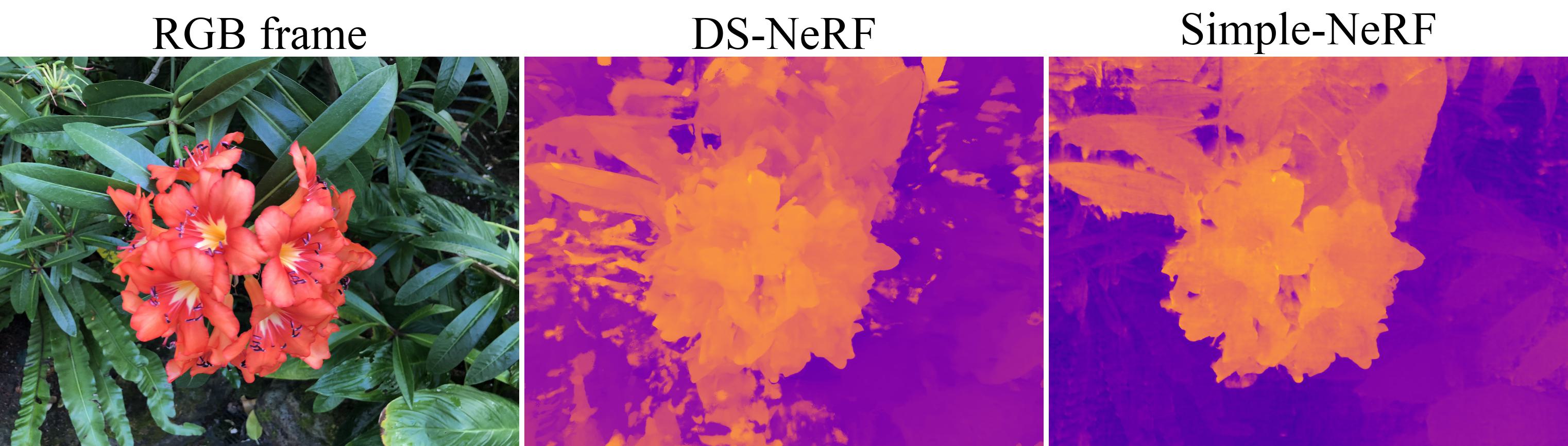}  
            \caption{\textit{Floater artifacts:}
            We visualize the depth learned by the NeRF model for an input frame from the NeRF-LLFF flower scene.
            }
            \label{fig:distortions-nerf-floaters}
        \end{subfigure}
        \begin{subfigure}{\linewidth}
            \centering
            \includegraphics[width=\linewidth]{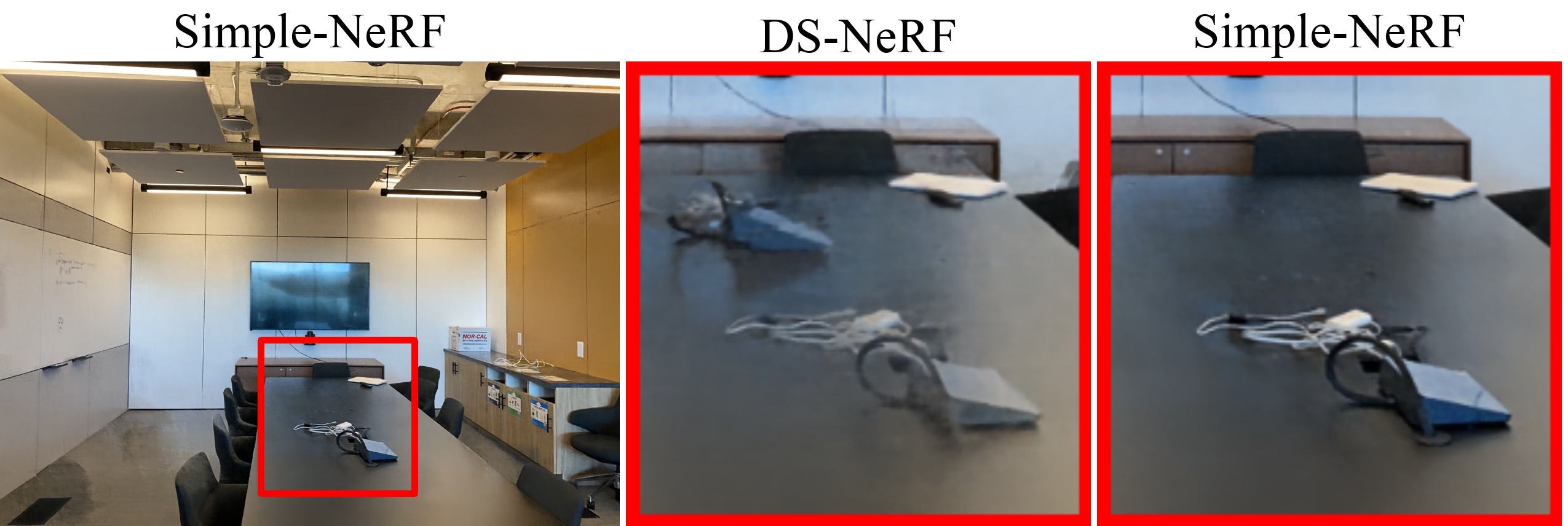}  
            \caption{\textit{Duplication artifacts:}
            To visualize the duplication artifacts that arise due to the shape-radiance ambiguity in sparse-input NeRF, we render an input frame by only changing the viewing direction.
            This is an example from the NeRF-LLFF room scene.
            }
            \label{fig:distortions-nerf-duplications}
        \end{subfigure}
        \caption{\textbf{Failure of sparse-input NeRF:}
        We show two shortcomings of the NeRF when trained with two input views on the NeRF-LLFF dataset.
        In Fig (a), we observe the floaters as small orange regions in the depth map.
        In Fig (b), we observe the duplication of the object on the table caused by the NeRF trying to blend the input images.
        Simple-NeRF introduces regularizations to mitigate these distortions as seen in both figures.
        We note that the models used to synthesize the above images include the sparse depth supervision (\cref{subsec:overall-loss}).
        }
        \label{fig:distortions-sparse-nerf}
    \end{figure}
}

\newcommand{\figureSimpleNerfHierarchicalSampling}{
    \begin{figure}
        \centering
        \begin{subfigure}{\linewidth}
            \centering
            \includegraphics[width=\linewidth]{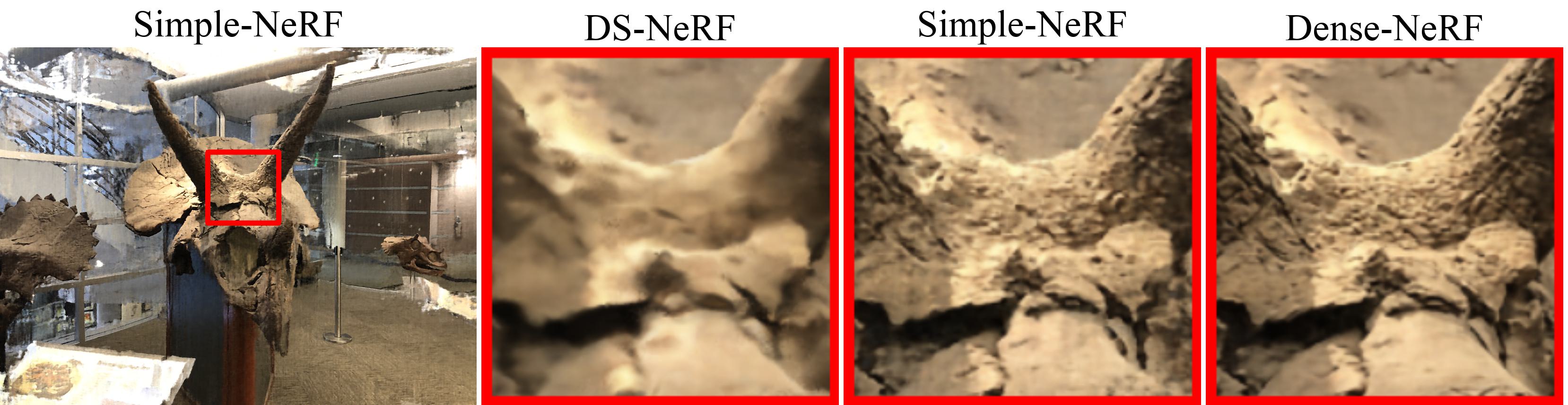}  
            \caption{
            The above images correspond to the NeRF-LLFF horns scene.
            We enlarge a small region of the frame to better observe the improvement in sharpness.
            }
            \label{fig:qualitative-simple-nerf-llff06a}
        \end{subfigure}
        \hfill
        \begin{subfigure}{\linewidth}
            \centering
            \includegraphics[width=\linewidth]{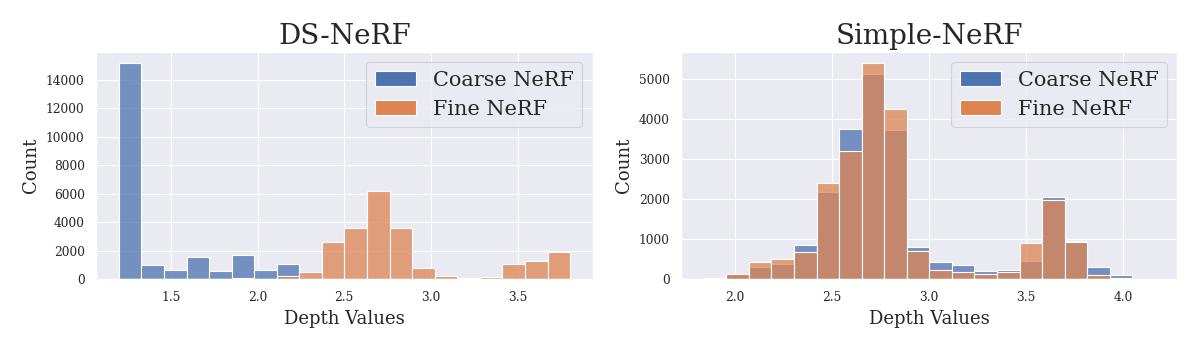}  
            \caption{
            Histogram of depth values predicted by the coarse and fine NeRF models for the image patch shown in Fig (a).
            }
            \label{fig:qualitative-simple-nerf-llff06b}
        \end{subfigure}
        \caption{\textbf{Ineffective hierarchical sampling in sparse-input NeRF:}
        Fig (b) shows that the coarse and fine models in the NeRF converge to different depth estimates when training with sparse input views.
        This leads to ineffective hierarchical sampling, resulting in blurry predictions in Fig (a).
        By predicting consistent depth estimates with the help of $\lossCoarseFineConsistency$, Simple-NeRF predicts consistent depth estimates leading to sharp reconstructions.
        We note that the models used to synthesize the above images include the sparse depth supervision (\cref{subsec:overall-loss}).
        }
        \label{fig:qualitative-simple-nerf-llff06}
    \end{figure}
}

\newcommand{\figureDistortionsSparseInputTensoRF}{
    \begin{figure}
        \centering
        \begin{subfigure}{\linewidth}
            \centering
            \includegraphics[width=\linewidth]{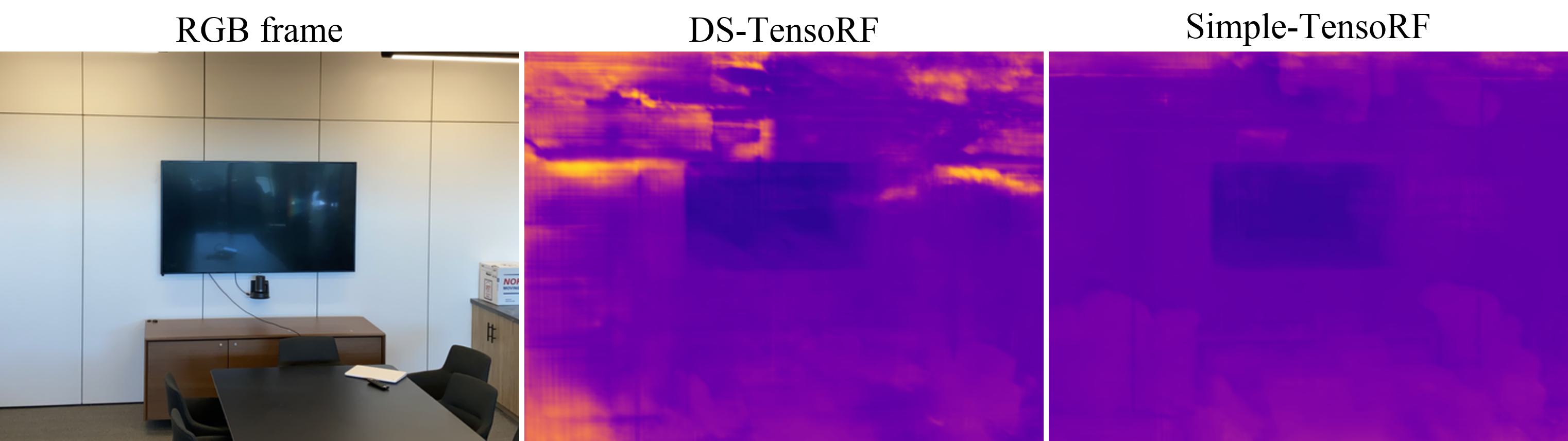}  
            \caption{\textit{Floater artifacts:}
            We visualize the depth learned by the TensoRF model for an input frame from the NeRF-LLFF room scene.
            }
            \label{fig:distortions-tensorf-floaters}
        \end{subfigure}
        \begin{subfigure}{\linewidth}
            \centering
            \includegraphics[width=\linewidth]{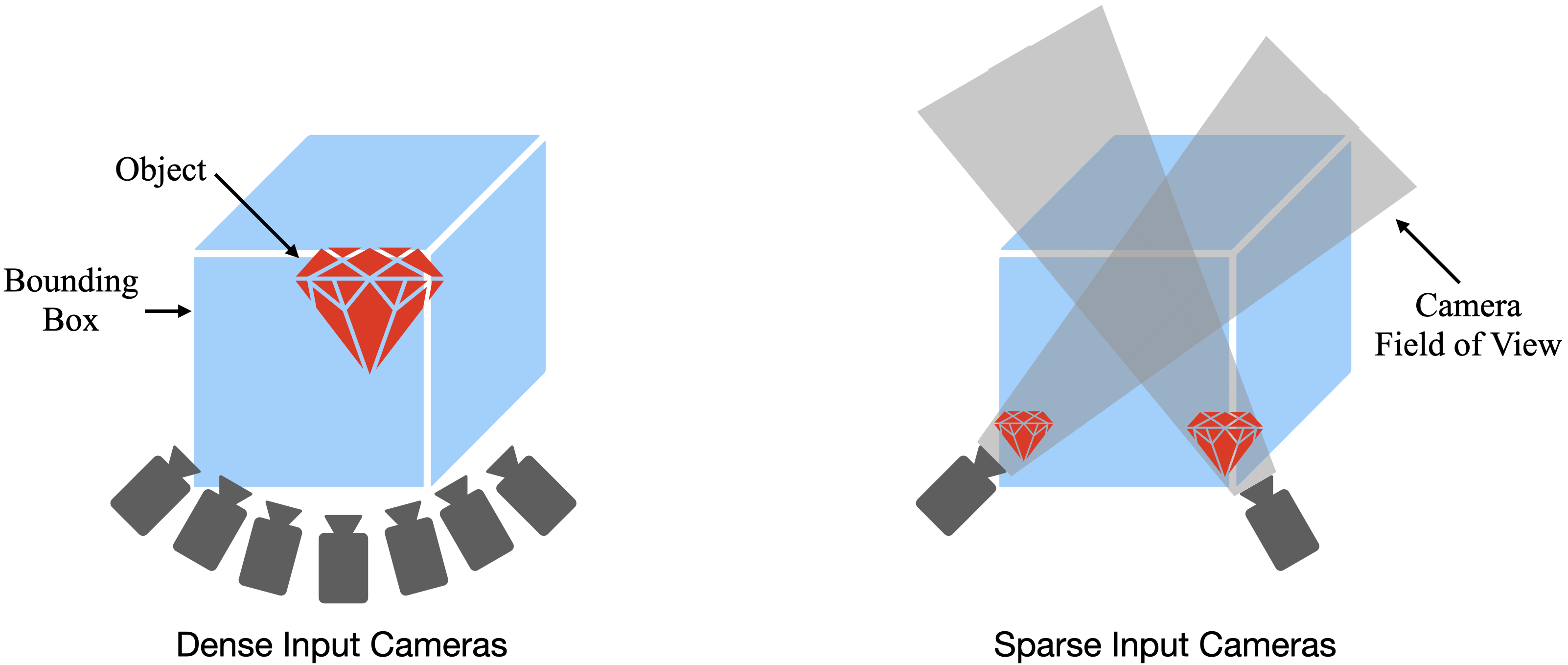}
            \caption{\textit{Objects close to camera:}
            We illustrate TensoRF incorrectly placing objects close to the cameras using a toy example.
            }
            \label{fig:distortions-tensorf-replications}
        \end{subfigure}
        \caption{\textbf{Failure of sparse-input TensoRF}:
        We show the two shortcomings of TensoRF when trained with few input views.
        In Fig (a), the orange regions indicate the floaters.
        For reference, we also show the depth learned by Simple-TensoRF, which is free from floaters.
        We note that the models used to synthesize these images include the sparse depth supervision (\cref{subsec:overall-loss}).
        In Fig (b), the image on the left depicts the true scene, which can be accurately learned by the TensoRF model provided with dense input views.
        The image on the right illustrates how TensoRF can incorrectly place the objects yet perfectly reconstruct the input views, when training with few input views.
        }
        \label{fig:distortions-sparse-tensorf}
    \end{figure}
}

\newcommand{\figureDistortionsSparseInputZipNeRF}{
    \begin{figure}
        \centering
        \begin{subfigure}{\linewidth}
            \centering
            \includegraphics[width=\linewidth]{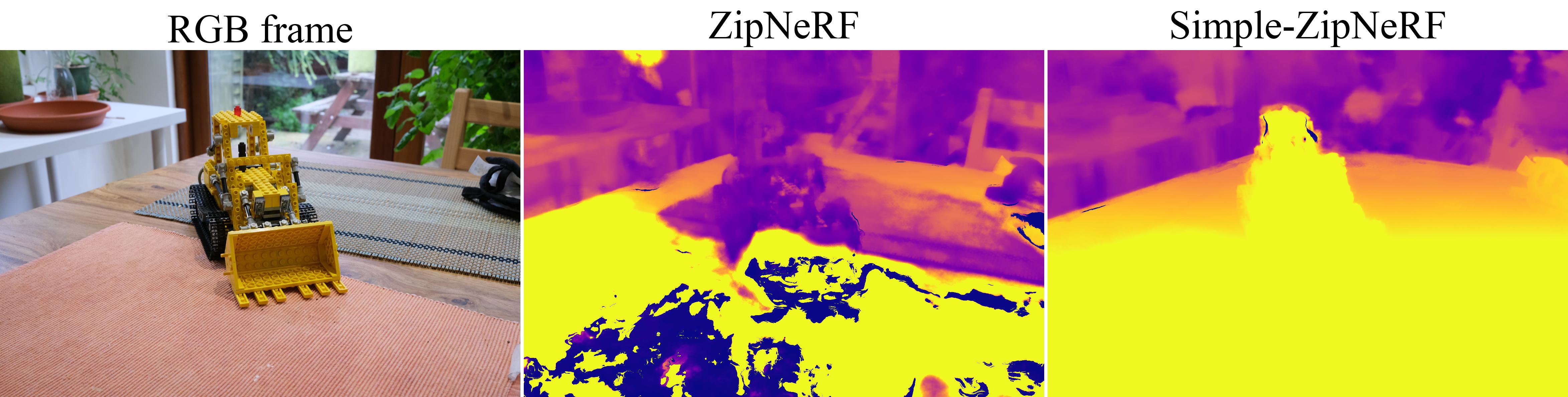}  
            \caption{\textit{Floater artifacts:}
            We visualize the depth learned by the ZipNeRF model for an input frame from the MipNeRF360 kitchen scene.
            }
            \label{fig:distortions-zipnerf-floaters}
        \end{subfigure}
        \begin{subfigure}{\linewidth}
            \centering
            \includegraphics[width=\linewidth]{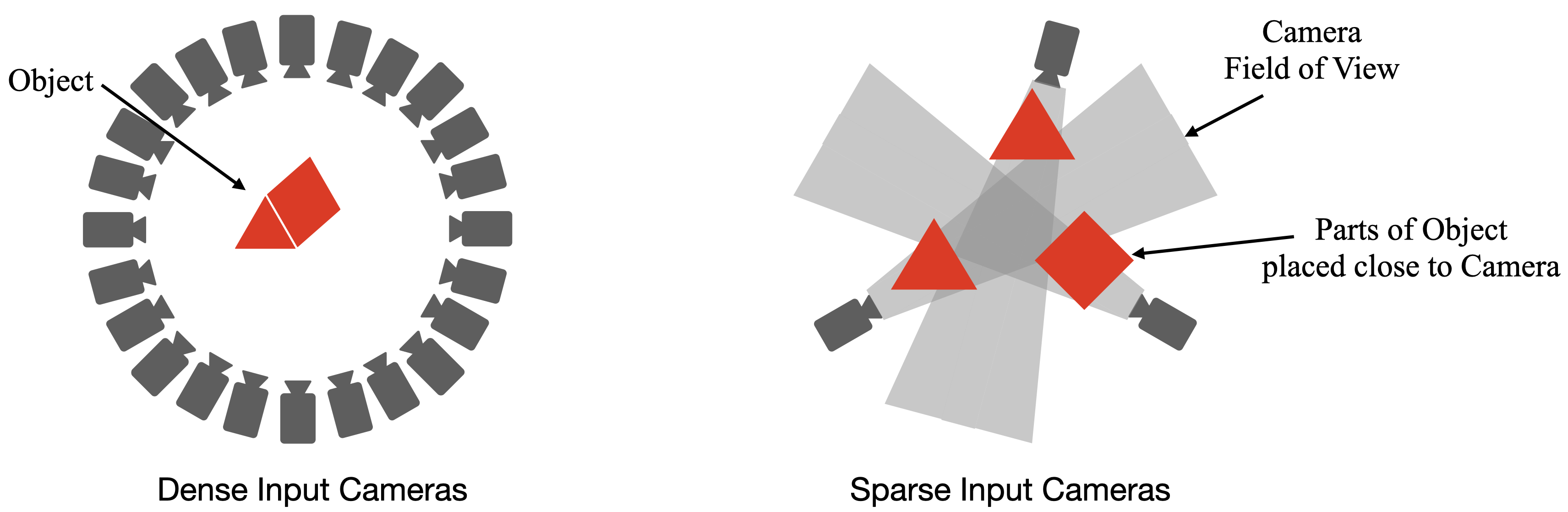}
            \caption{\textit{Objects close to camera:}
            We illustrate ZipNeRF incorrectly placing objects close to the cameras using a toy example.
            }
            \label{fig:distortions-zipnerf-replications}
        \end{subfigure}
        \caption{We show two shortcomings of ZipNeRF when trained with few input views.
        In Fig (a), while the RGB frame for an input view is reconstructed perfectly, we observe floaters in the depth image, shown by the dark-blue regions.
        For reference, we also show the depth learned by Simple-ZipNeRF, which is free from floaters and better reconstructs the scene.
        In Fig (b), the image on the left depicts the true scene, which can be accurately learned by the ZipNeRF model provided with dense input views.
        The image on the right illustrates how the sparse-input ZipNeRF model can incorrectly place parts of the object close to the cameras, yet perfectly reconstruct the input views.
        }
        \label{fig:distortions-sparse-zipnerf}
    \end{figure}
}

\newcommand{\figureReliableDepthEstimates}{
    \begin{figure}
        \centering
        \includegraphics[width=\linewidth]{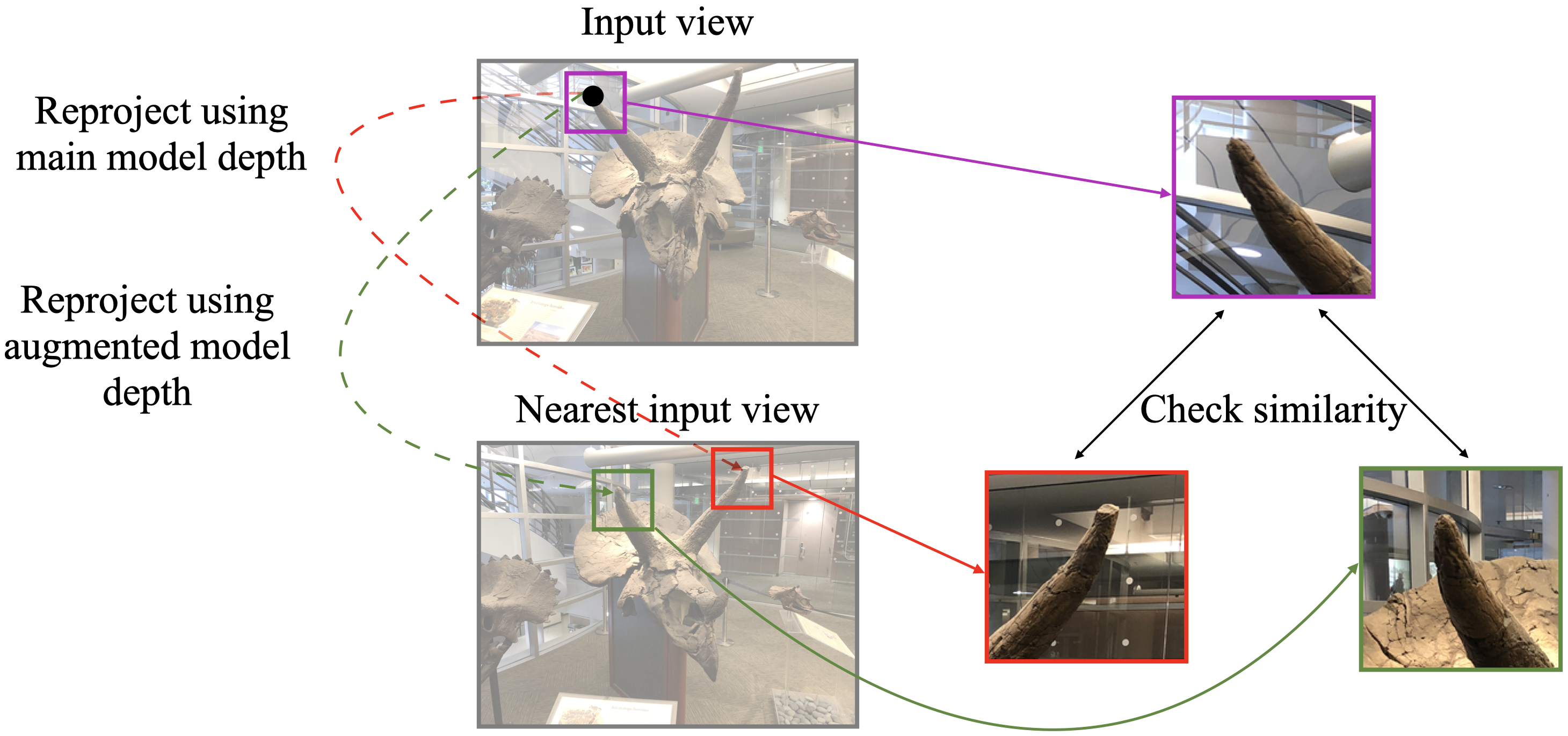}
        \caption{\textbf{Determining the reliability of depths for supervision:} We choose the depth that has higher similarity, with respect to the patches reprojected to the nearest input view, to supervise the other model (\cref{subsubsec:reliable-depth-estimates}).
        The patches are only representative and are not to scale.
        }
        \label{fig:reliable-depth-estimation}
    \end{figure}
}

\newcommand{\figureQualitativeSimpleNerfRealEstateFirst}{
    \begin{figure*}
        \centering
        \includegraphics[width=\linewidth]{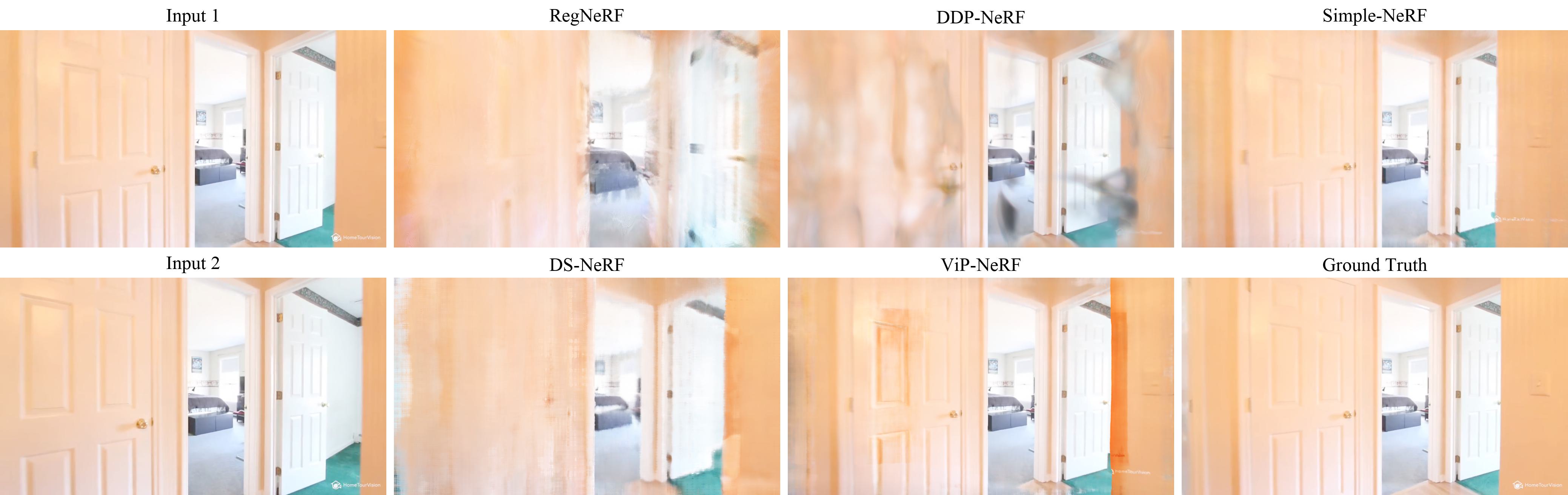}  
        \caption{\textbf{Qualitative examples of NeRF based models on the RealEstate-10K dataset with two input views.}
        While DDP-NeRF predictions contain blurred regions, ViP-NeRF predictions are color-saturated in certain regions of the door.
        Simple-NeRF does not suffer from these distortions and synthesizes a clean frame.
        For reference, we also show the input images.
        }
        \label{fig:qualitative-simple-nerf-realestate01}
    \end{figure*}
}

\newcommand{\figureQualitativeSimpleNerfRealEstateSecond}{
    \begin{figure*}
        \centering
        \includegraphics[width=0.995\linewidth]{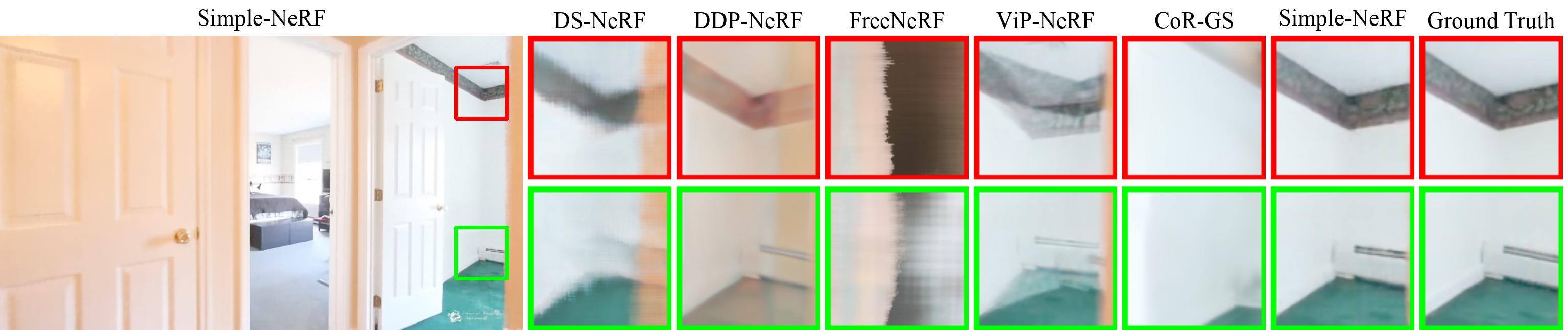}  
        \caption{\textbf{Qualitative examples of NeRF and 3DGS based models on RealEstate-10K dataset with three input views.}
        Simple-NeRF predictions are closest to the ground truth among all the models.
        In particular, DDP-NeRF predictions have a different shade of color and ViP-NeRF suffers from shape-radiance ambiguity, creating duplication artifacts.
        }
        \label{fig:qualitative-simple-nerf-realestate02}
    \end{figure*}
}

\newcommand{\figureQualitativeSimpleNerfRealEstateThird}{
    \begin{figure*}
        \centering
        \includegraphics[width=\linewidth]{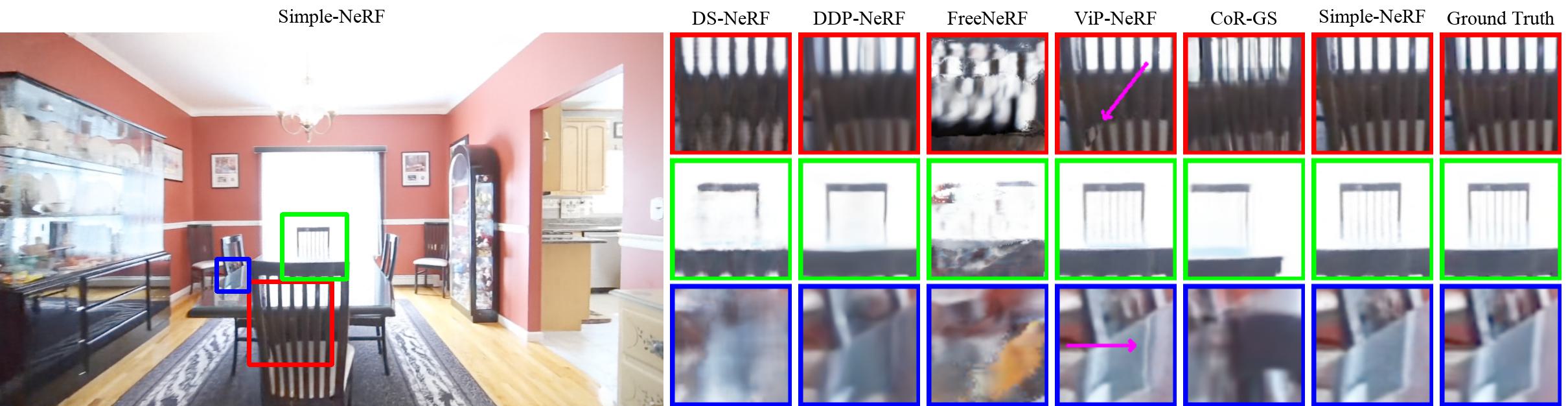}  
        \caption{\textbf{Qualitative examples of NeRF and 3DGS based models on the RealEstate-10K dataset with four input views.}
        We find that Simple-NeRF and ViP-NeRF perform the best among all the models.
        However, ViP-NeRF predictions contain minor distortions, as pointed out by the magenta arrow, which is rectified by Simple-NeRF.
        }
        \label{fig:qualitative-simple-nerf-realestate03}
    \end{figure*}
}

\newcommand{\figureQualitativeSimpleNerfLlffFirst}{
    \begin{figure*}
        \centering
        \includegraphics[width=\linewidth]{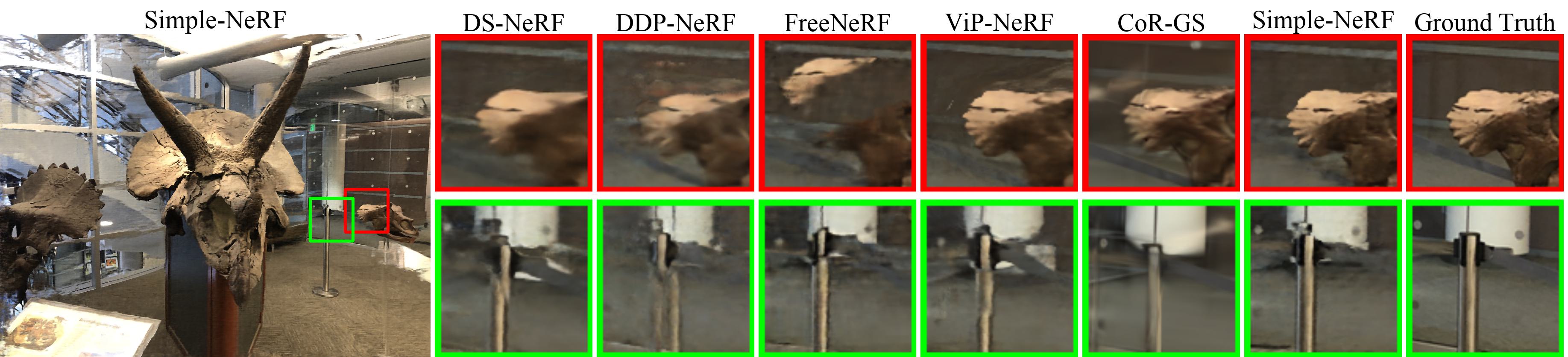}  
        \caption{\textbf{Qualitative examples of NeRF and 3DGS based models on the NeRF-LLFF dataset with two input views.}
        DDP-NeRF and ViP-NeRF synthesize frames with broken objects in the second row, and FreeNeRF breaks the object in the first row due to incorrect depth estimations.
        Simple-NeRF produces sharper frames devoid of such artifacts.
        }
        \label{fig:qualitative-simple-nerf-llff01}
    \end{figure*}
}

\newcommand{\figureQualitativeSimpleNerfLlffSecond}{
    \begin{figure*}
        \centering
        \includegraphics[width=\linewidth]{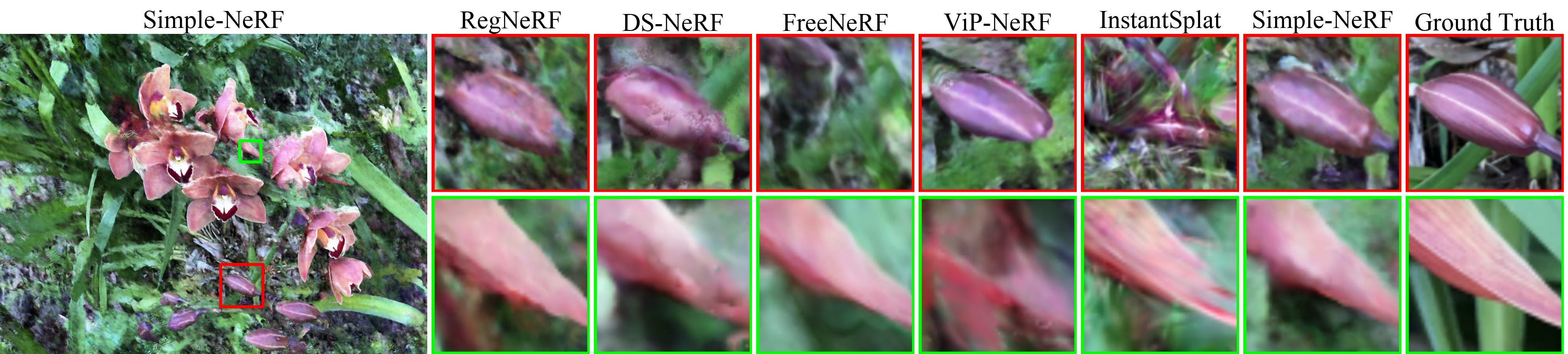}  
        \caption{\textbf{Qualitative examples of NeRF and 3DGS based models on the NeRF-LLFF dataset with three input views.}
        In the first row, the orchid is displaced out of the cropped box in the FreeNeRF prediction, due to incorrect depth estimation.
        While ViP-NeRF and RegNeRF fail to predict the complete orchid accurately causing distortions at either end, InstantSplat prediction contains severe distortions.
        In the second row, ViP-NeRF prediction contains severe distortions.
        Simple-NeRF reconstructs the best among all the models in both examples.
        }
        \label{fig:qualitative-simple-nerf-llff02}
    \end{figure*}
}

\newcommand{\figureQualitativeSimpleNerfLlffThird}{
    \begin{figure*}
        \centering
        \includegraphics[width=\linewidth]{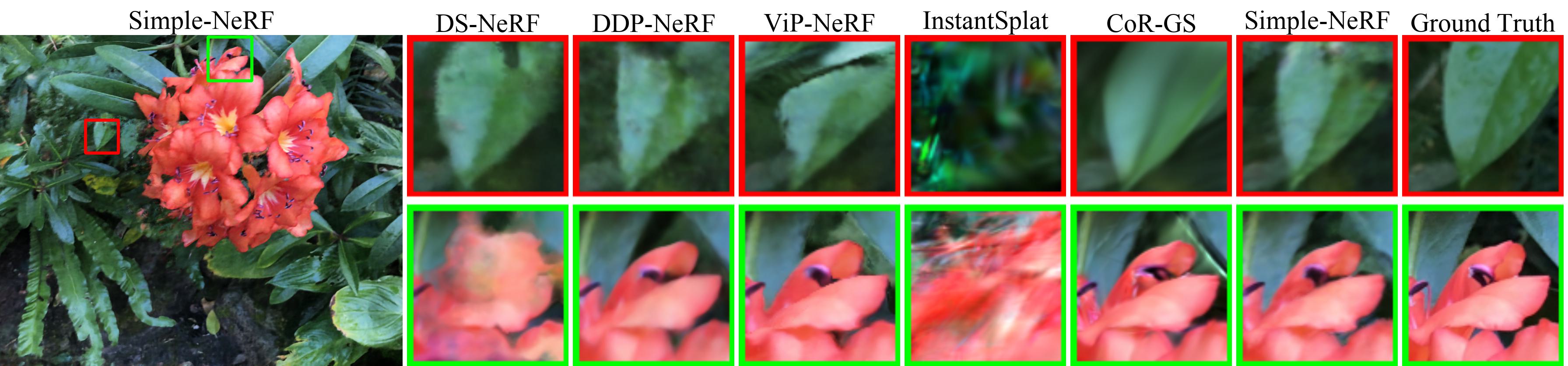}  
        \caption{\textbf{Qualitative examples of NeRF and 3DGS based models on the NeRF-LLFF dataset with four input views.}
        In the first row, we find that ViP-NeRF, FreeNeRF, and DDP-NeRF struggle to reconstruct the shape of the leaf accurately, and CoR-GS output is excessively blurred.
        In the second row, DS-NeRF introduces floaters.
        Simple-NeRF does not suffer from such artifacts and reconstructs the shapes better.
        }
        \label{fig:qualitative-simple-nerf-llff03}
    \end{figure*}
}

\newcommand{\figureQualitativeSimpleNerfLlffFourth}{
    \begin{figure*}
        \centering
        \includegraphics[width=\linewidth]{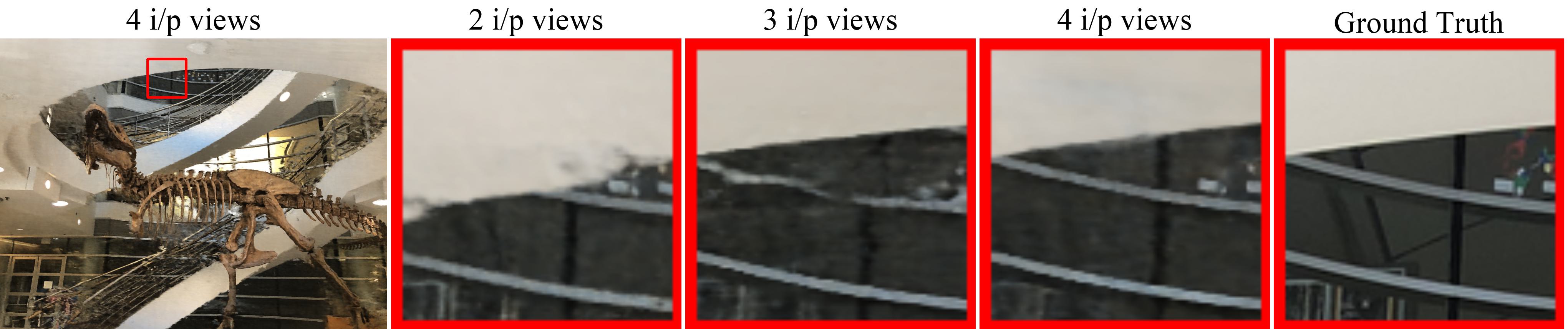}  
        \caption{\textbf{Qualitative examples of Simple-NeRF on the NeRF-LLFF dataset with two, three, and four input views.}
        We observe errors in depth estimation with two input views, causing a change in the position of the roof.
        While this is corrected with three input views, there are a few shape distortions in the metal rods.
        With four input views, even such distortions are corrected.
        }
        \label{fig:qualitative-simple-nerf-llff04}
    \end{figure*}
}

\newcommand{\figureQualitativeSimpleNerfDepthFirst}{
    \begin{figure}
        \centering
        \includegraphics[width=\linewidth]{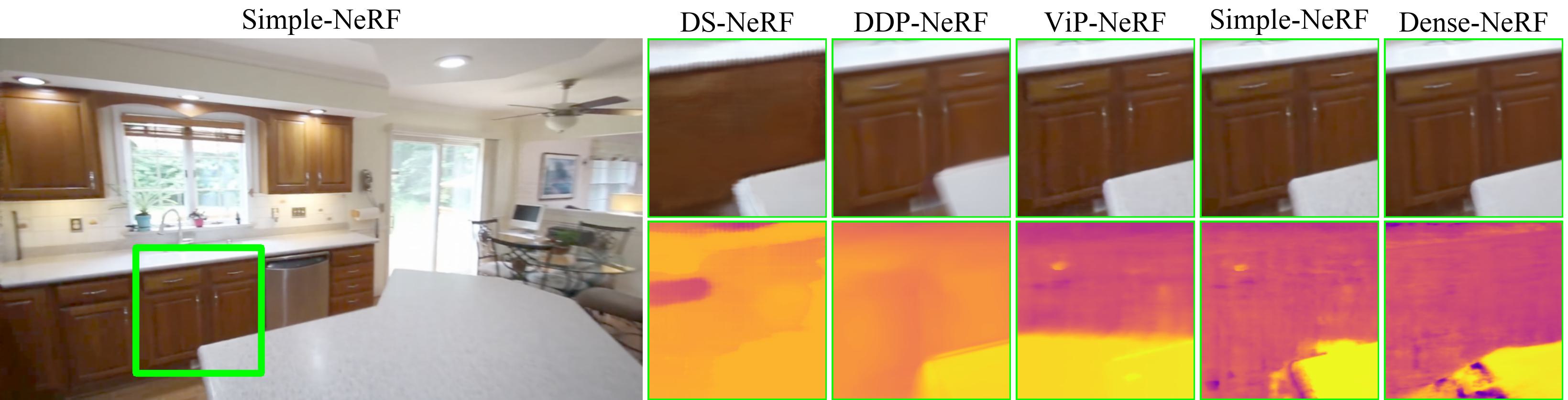}  
        \includegraphics[width=\linewidth]{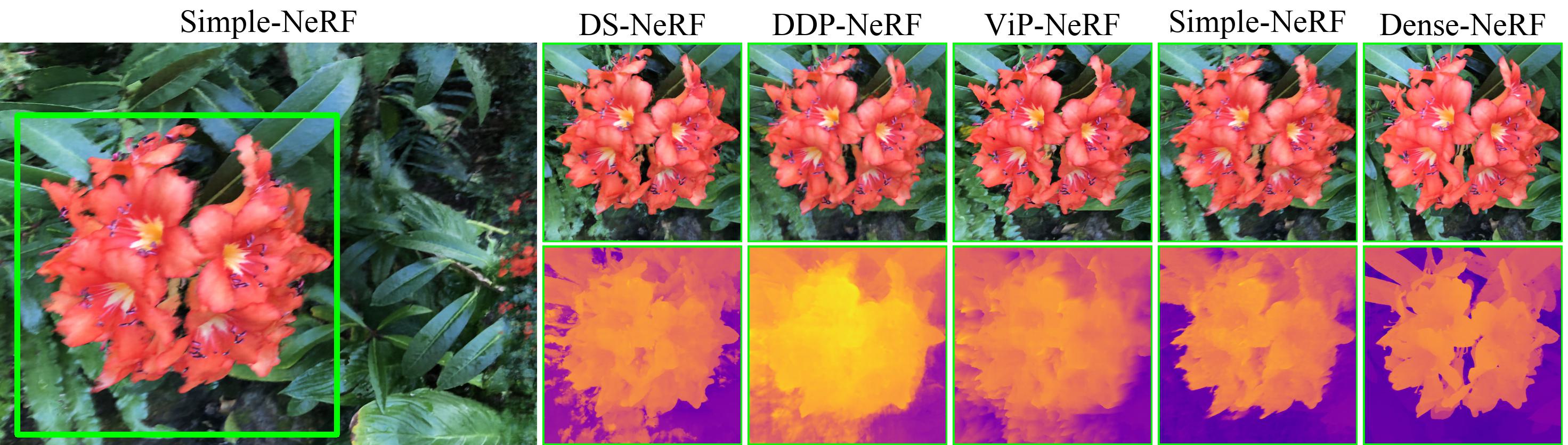}  
        \caption{\textbf{Estimated depth maps of NeRF based models} on RealEstate-10K and NeRF-LLFF datasets with two input views.
        In both examples, the two rows show the predicted images and the depths respectively.
        We find that Simple-NeRF is significantly better at estimating the scene depth.
        Also, DDP-NeRF synthesizes the left table edge at a different angle due to incorrect depth estimation.
        }
        \label{fig:qualitative-simple-nerf-depth}
    \end{figure}
}

\newcommand{\figureQualitativeSimpleNerfAblationsFirst}{
    \begin{figure*}
        \centering
        \begin{subfigure}{0.54\linewidth}
            \centering
            \includegraphics[width=\linewidth]{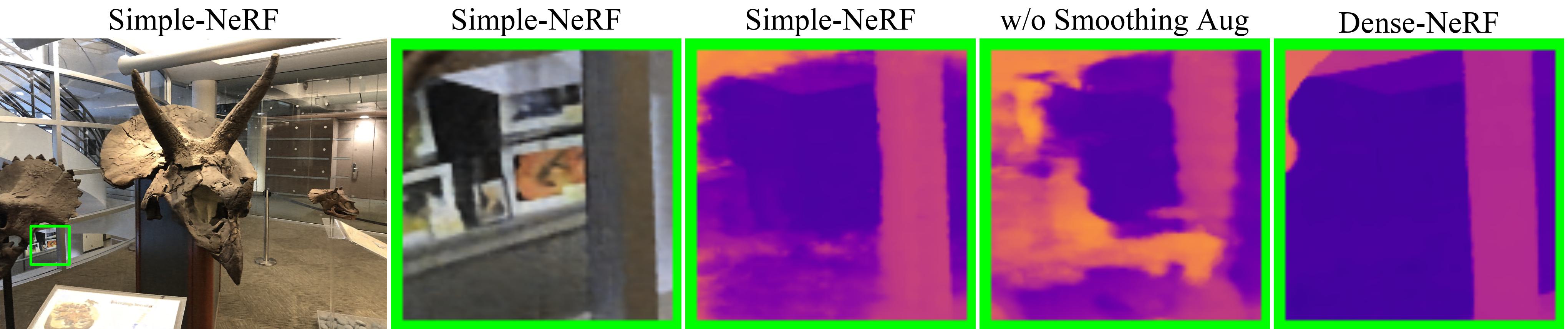}  
            \caption{\textit{Without smoothing augmentation:}
            The ablated model introduces floaters that are significantly reduced by using the smoothing augmentation model.
            }
            \label{fig:qualitative-simple-nerf-ablation01a}
        \end{subfigure}
        \hfill
        \begin{subfigure}{0.44\linewidth}
            \centering
            \includegraphics[width=\linewidth]{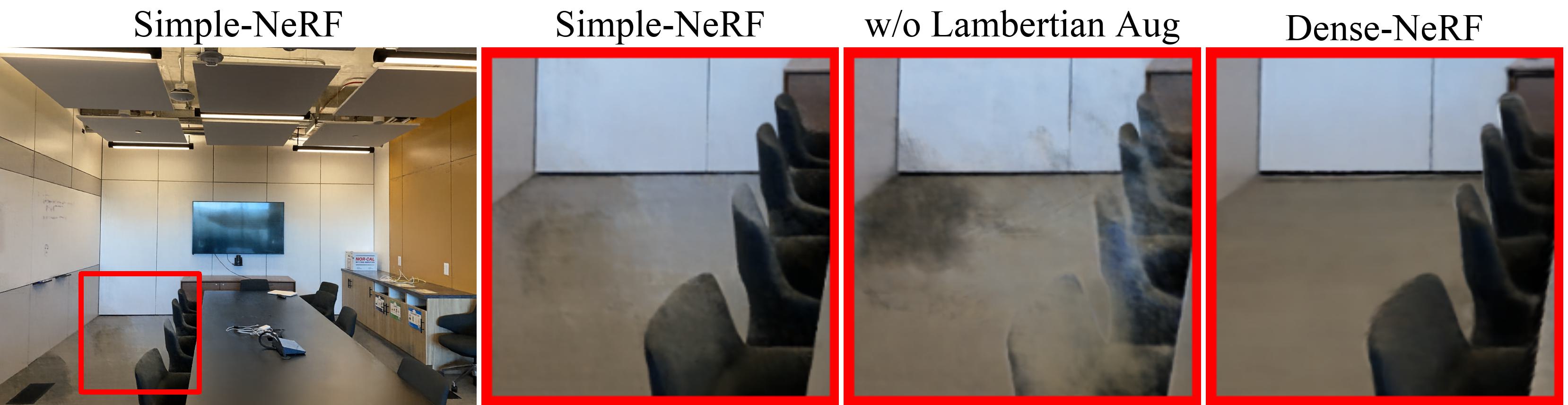}  
            \caption{\textit{Without Lambertian augmentation:}
            The ablated model suffers from shape-radiance ambiguity and produces duplication artifacts.
            }
            \label{fig:qualitative-simple-nerf-ablation01b}
        \end{subfigure}
        \begin{subfigure}{0.54\linewidth}
            \centering
            \includegraphics[width=\linewidth]{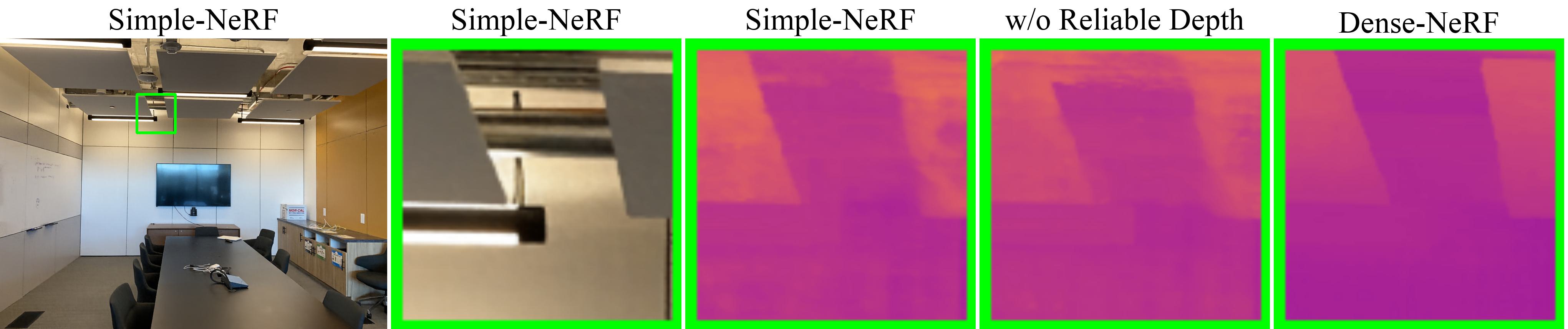}  
            \caption{\textit{Without reliability of depth supervision:}
            The smoothing augmentation model struggles to learn sharp depth discontinuities at true depth edges.
            Supervising the main model using such depths without determining their reliability causes the main model to learn incorrect depth.
            As a result, the ablated model fails to learn sharp depth discontinuities at certain regions.
            }
            \label{fig:qualitative-simple-nerf-ablation01c}
        \end{subfigure}
        \hfill
        \begin{subfigure}{0.44\linewidth}
            \centering
            \includegraphics[width=\linewidth]{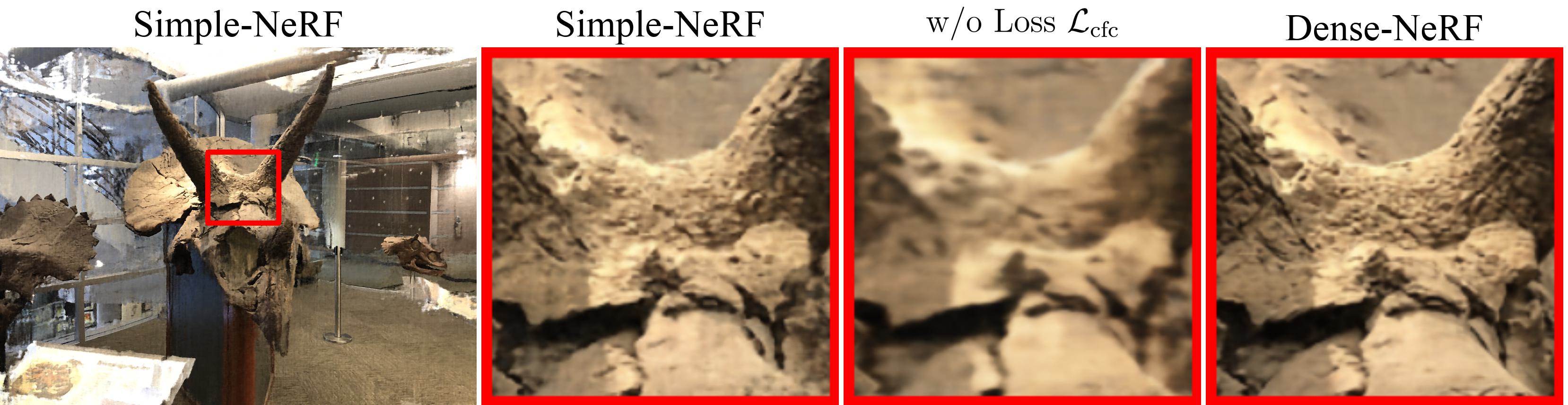}  
            \caption{\textit{Without coarse-fine consistency:}
            We observe that while Simple-NeRF predictions are sharper, the ablated model without coarse-fine consistency loss, $\lossCoarseFineConsistency$ produces blurred renders.
            This is similar to \cref{fig:qualitative-simple-nerf-llff06a}, where we observe DS-NeRF also produce blurred renders.
            }
            \label{fig:qualitative-simple-nerf-ablation01d}
        \end{subfigure}
        \caption{\textbf{Qualitative examples for Simple-NeRF ablated models on the NeRF-LLFF dataset with two input views.}
        We also show the outputs of the dense-input NeRF for reference.
        }
        \label{fig:qualitative-simple-nerf-ablations01}
    \end{figure*}
}

\newcommand{\figureQualitativeSimpleTensorfFirst}{
    \begin{figure*}
        \centering
        \begin{subfigure}{0.48\linewidth}
            \centering
            \includegraphics[width=\linewidth]{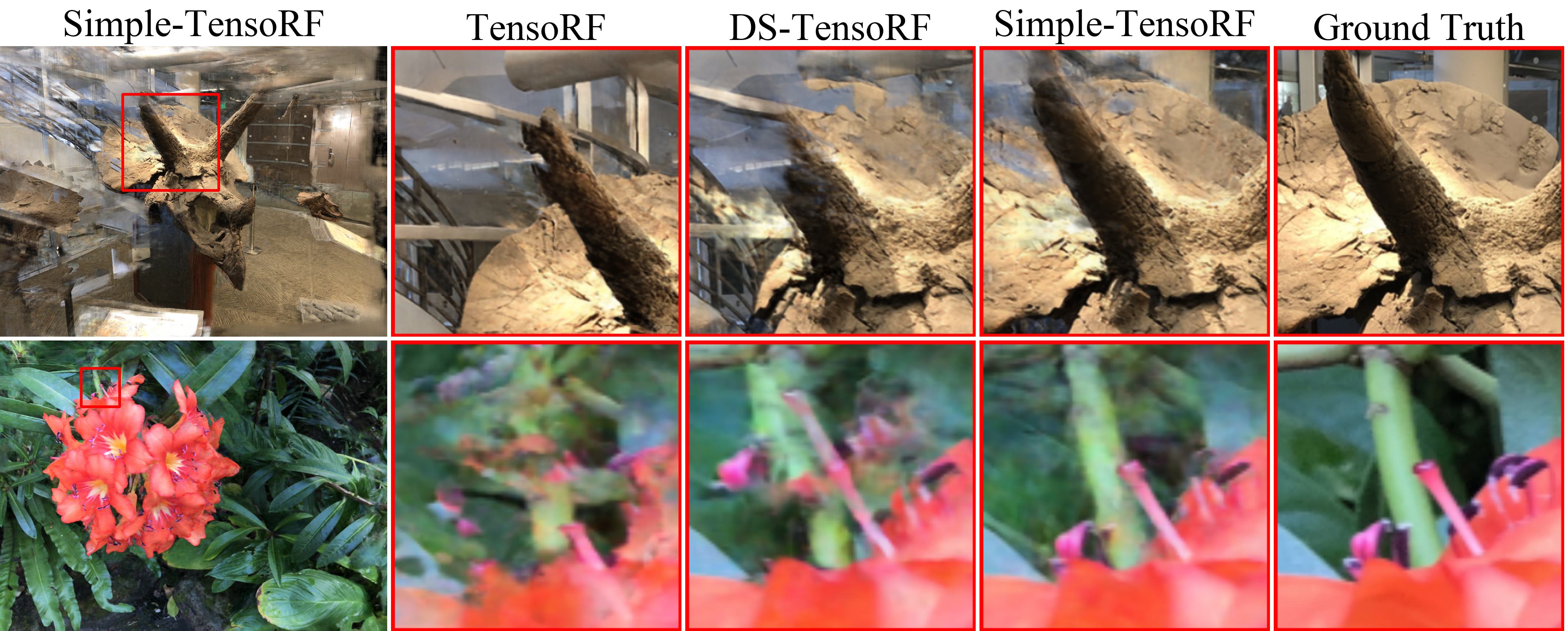}  
            \caption{\textit{NeRF-LLFF dataset:}
            In the first example, we find that the horn is broken and almost half of the bony frill is missing in the renders of TensoRF and DS-TensoRF.
            In the second example, TensoRF and DS-TensoRF extend the red stigma and break the green stem.
            Simple-TensoRF does not introduce such distortions and is closest to the ground truth.
            }
            \label{fig:qualitative-simple-tensorf-llff01}
        \end{subfigure}
        \hfill
        \begin{subfigure}{0.48\linewidth}
            \centering
            \includegraphics[width=\linewidth]{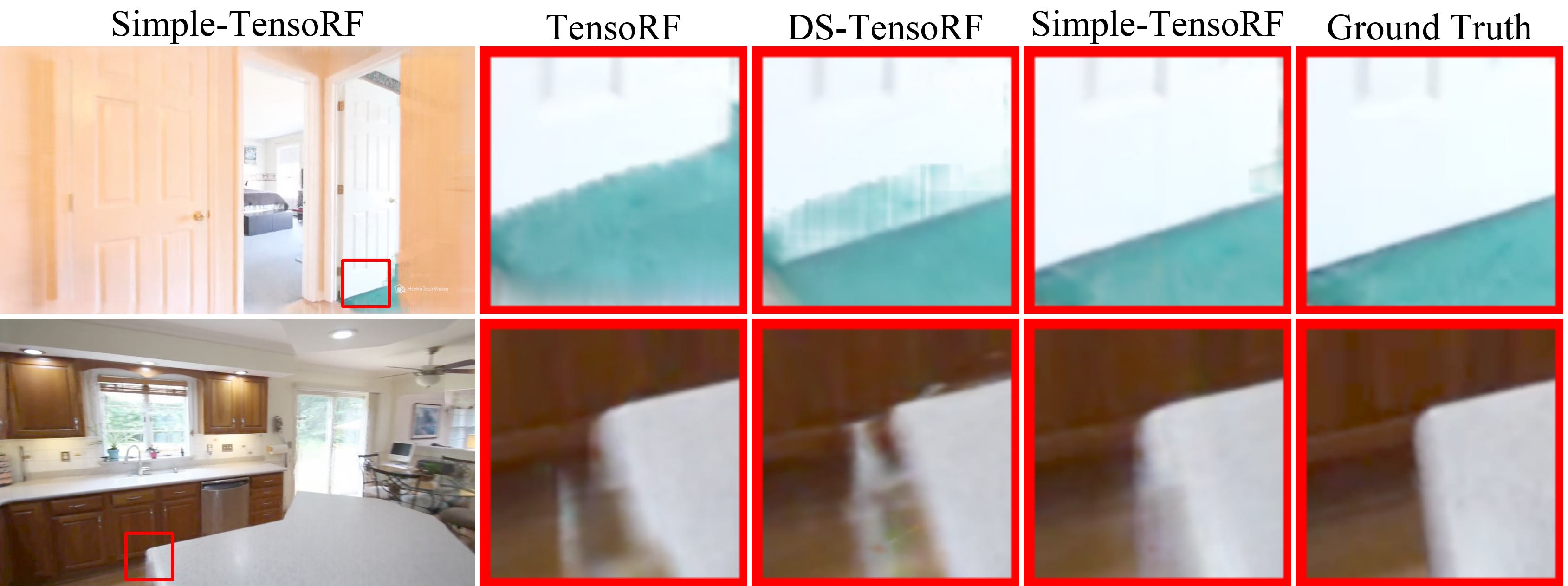}  
            \caption{\textit{RealEstate-10K dataset:}
            In the first example, we observe a shift in the position of the door due to incorrect depth estimation.
            With sparse depth supervision, DS-TensoRF moves the door to the correct position, but only partially.
            Adding our augmentations provides the best result.
            Similarly, we see distortions in the frames rendered by TensoRF and DS-TensoRF in the second example, which are reduced significantly by Simple-TensoRF.
            }
            \label{fig:qualitative-simple-tensorf-realestate01}
        \end{subfigure}
        \caption{\textbf{Qualitative examples of TensoRF based models with three input views.}
        }
        \label{fig:qualitative-simple-tensorf}
    \end{figure*}
}

\newcommand{\figureQualitativeSimpleTensorfDepthFirst}{
    \begin{figure}
        \centering
        \includegraphics[width=\linewidth]{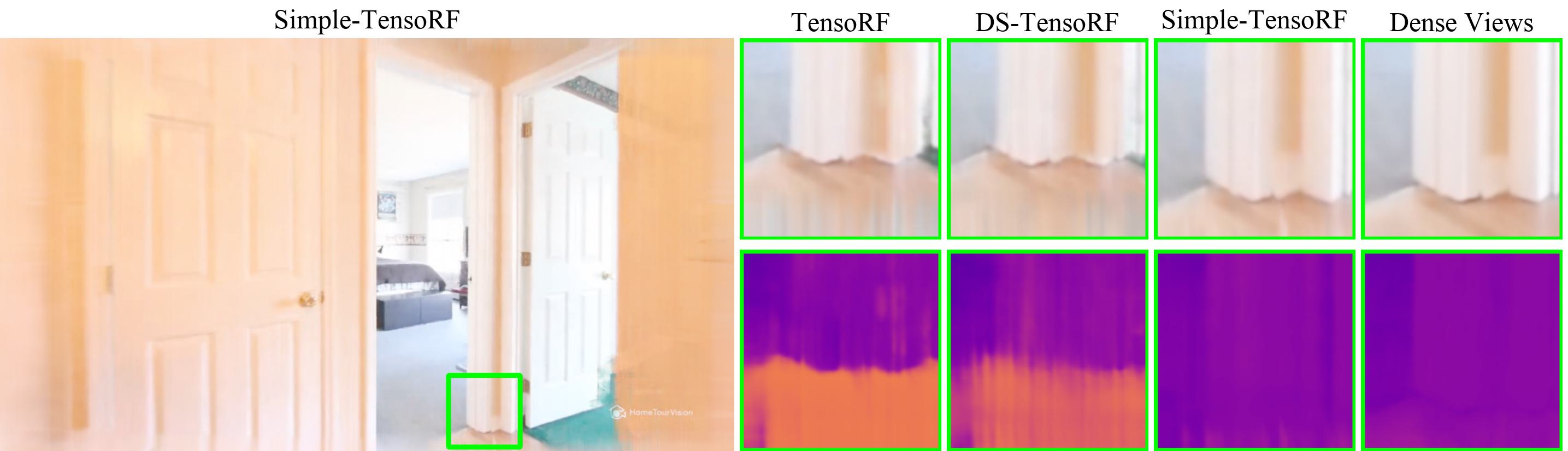}  
        \includegraphics[width=\linewidth]{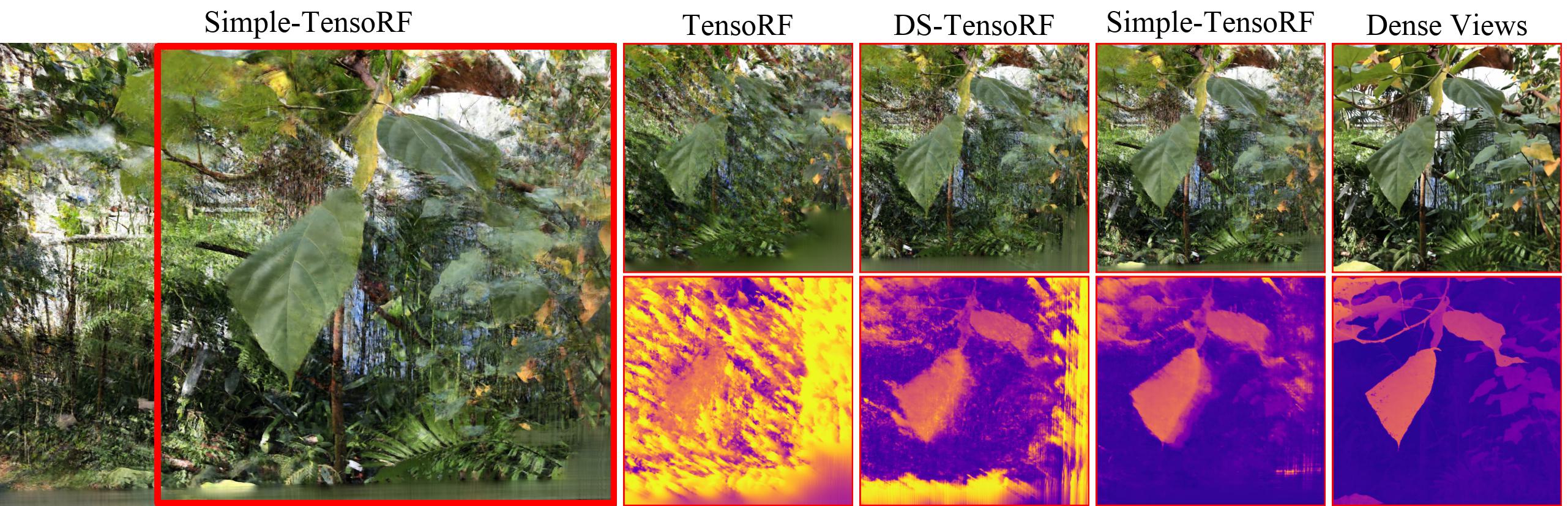}  
        \caption{\textbf{Estimated depth maps of TensoRF based models} on RealEstate-10K and NeRF-LLFF datasets with three input views.
        In both examples, the two rows show the predicted images and the depths respectively.
        In the first example, TensoRF and DS-TensoRF incorrectly estimate the depth of the floor as shown by the orange regions.
        In the second row, while TensoRF is unable to estimate the scene geometry, DS-TensoRF is unable to mitigate all the floaters in orange color.
        We find that Simple-TensoRF is significantly better at estimating the scene depth.
        }
        \label{fig:qualitative-simple-tensorf-depth}
    \end{figure}
}

\newcommand{\figureQualitativeSimpleTensorfAblationsFirst}{
    \begin{figure}
        \centering
        \includegraphics[width=\linewidth]{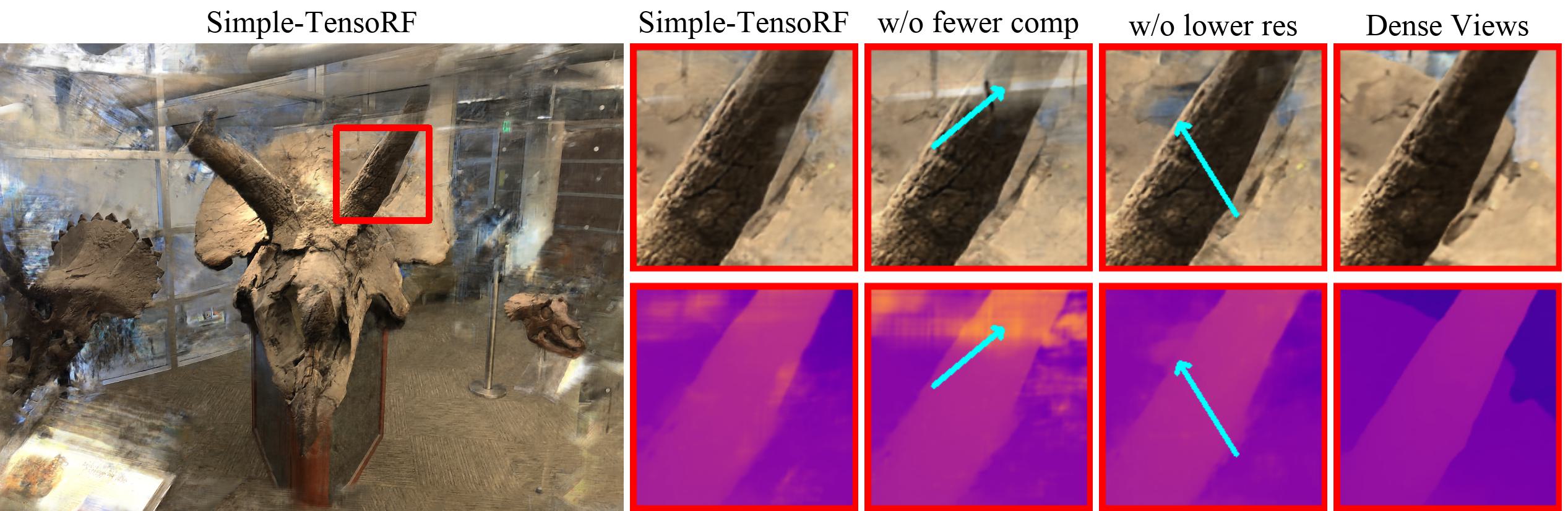}  
        \caption{\textbf{Qualitative examples of Simple-TensoRF ablations} on NeRF-LLFF dataset with three input views.
        Reducing the tensor resolution only leads to translucent floaters as shown by the arrows in the second column.
        On the other hand, only reducing the number of tensor decomposed components leads to small opaque floaters as shown by the arrows in the third column.
        }
        \label{fig:qualitative-simple-tensorf-ablations01}
    \end{figure}
}

\newcommand{\figureQualitativeSimpleZipnerfMipNerfFirstSecond}{
    \begin{figure*}
        \centering
        \begin{subfigure}{\linewidth}
            \centering
            \includegraphics[width=\linewidth]{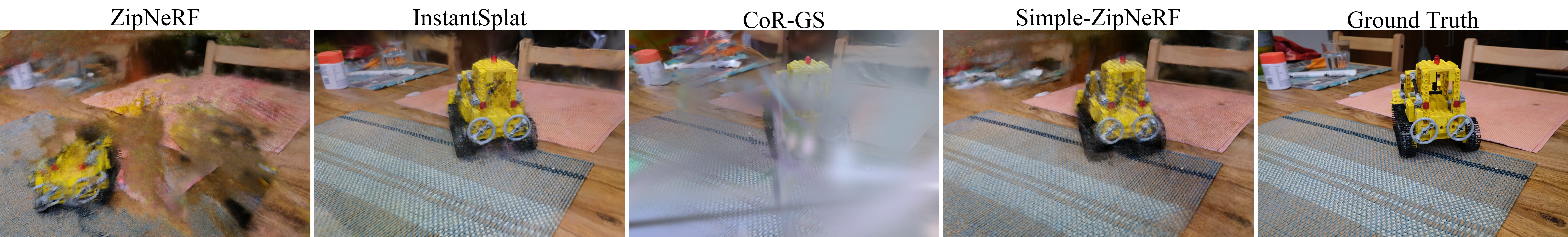}  
            \caption{12 input views.}
            \label{fig:qualitative-simple-zipnerf-mipnerf360-01}
        \end{subfigure}
        \vspace{1em}
        \begin{subfigure}{\linewidth}
            \centering
            \includegraphics[width=\linewidth]{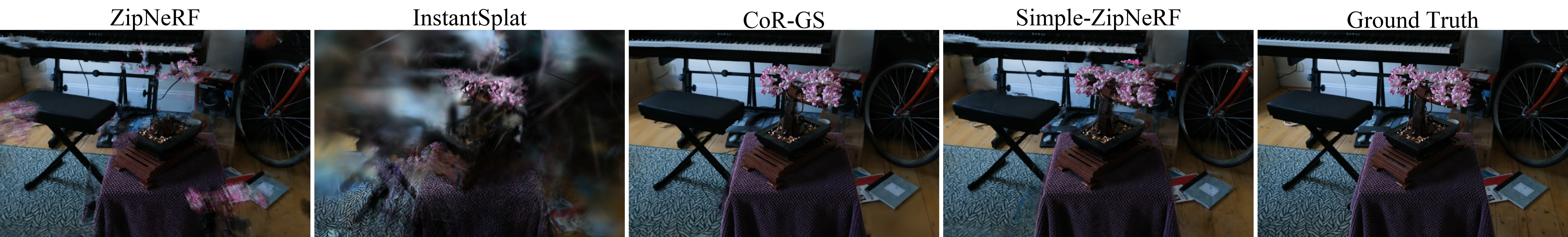}  
            \caption{20 input views.}
            \label{fig:qualitative-simple-zipnerf-mipnerf360-02}
        \end{subfigure}
        \caption{\textbf{Qualitative examples of ZipNeRF and 3DGS based models on the MipNeRF360 dataset.}
        While the ZipNeRF and CoR-GS create large floaters in the first row, InstantSplat fails to reconstruct the bonsai in the second row.
        Simple-ZipNeRF is able to faithfully reconstruct the scene without such artifacts.
        }
        \label{fig:qualitative-simple-zipnerf-mipnerf360-0102}
    \end{figure*}
}

\newcommand{\figureQualitativeSimpleZipnerfMipNerfThird}{
    \begin{figure}
        \centering
        \includegraphics[width=\linewidth]{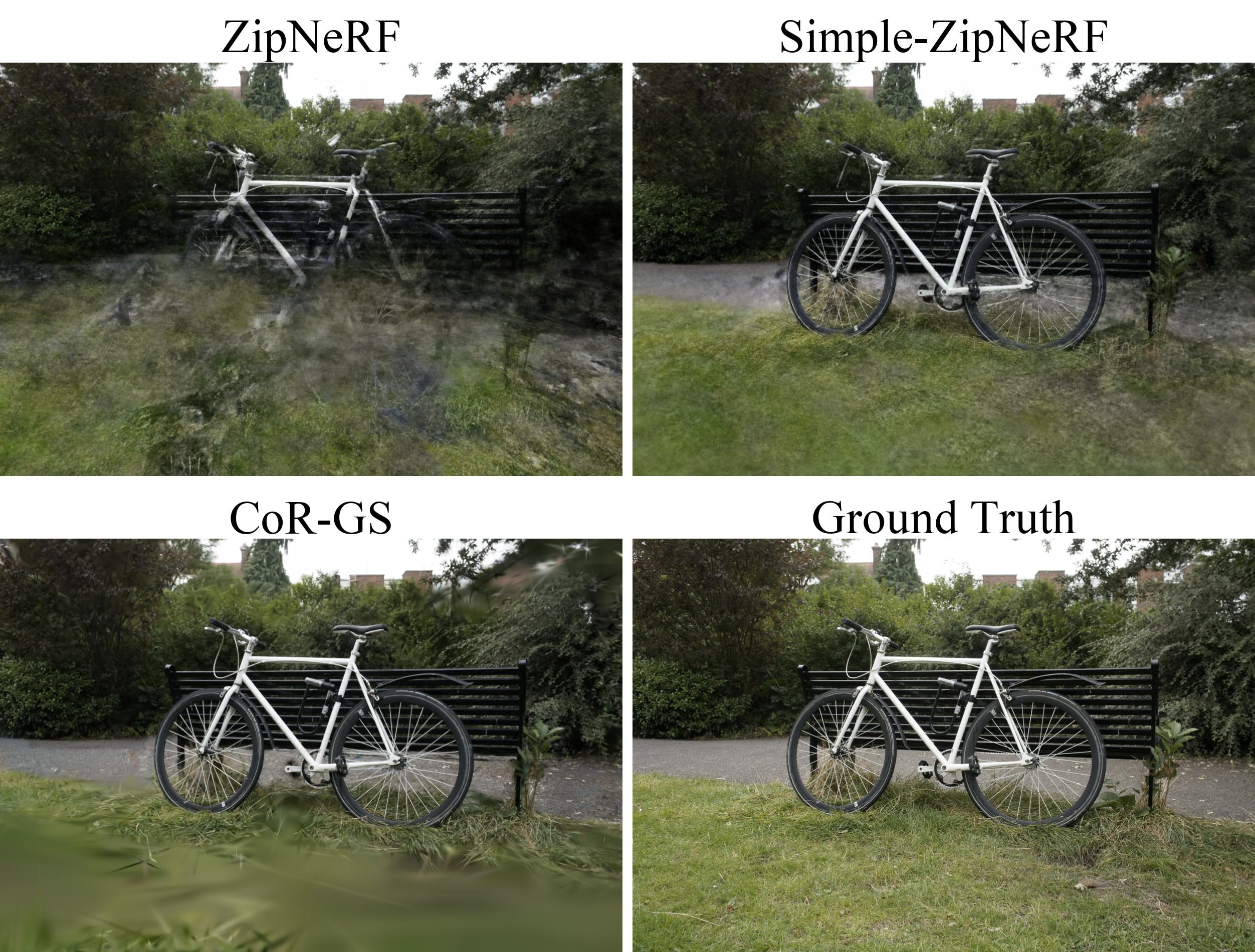}  
        \caption{\textbf{Qualitative examples of ZipNeRF and 3DGS based models on the MipNeRF360 dataset with 36 input views.}
        We observe CoR-GS excessively smoothing the grass regions and ZipNeRF distorting the geometry of the bicycle.
        Simple-ZipNeRF reconstructs both the bicycle and the bench without smoothing the grass regions.
        }
        \label{fig:qualitative-simple-zipnerf-mipnerf360-03}
    \end{figure}
}

\newcommand{\figureQualitativeSimpleZipnerfNerfSyntheticFirstSecondThird}{
    \begin{figure}
        \centering
        \begin{subfigure}{0.32\linewidth}
            \centering
            \includegraphics[width=\linewidth]{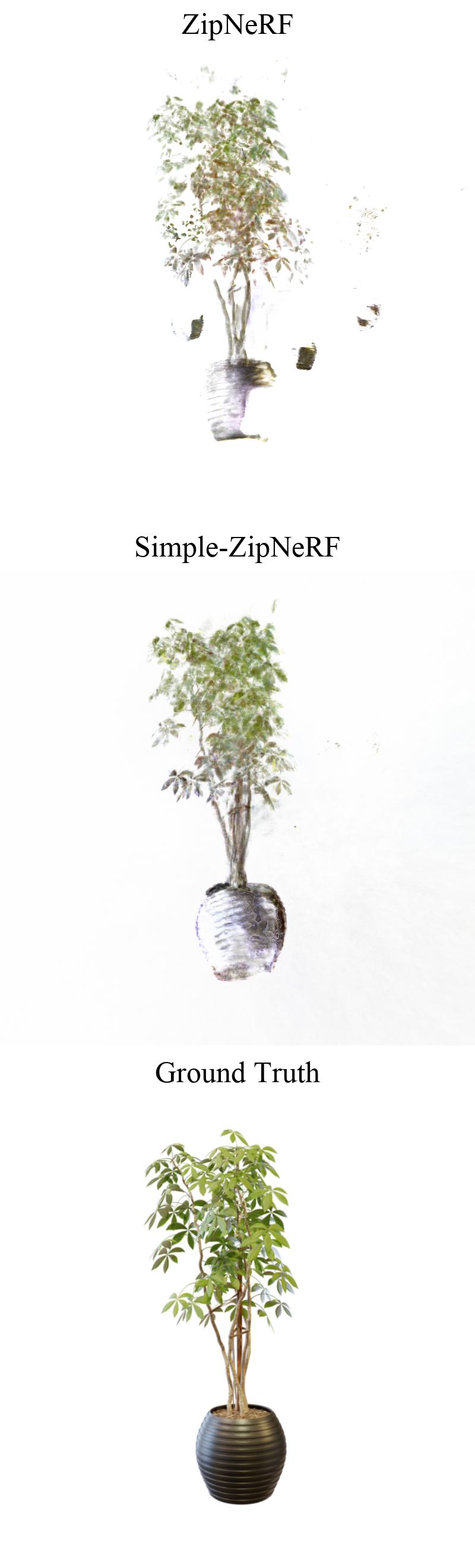}  
            \caption{4 input views.
            }
            \label{fig:qualitative-simple-zipnerf-nerfsynthetic01}
        \end{subfigure}
        \hfill
        \begin{subfigure}{0.32\linewidth}
            \centering
            \includegraphics[width=\linewidth]{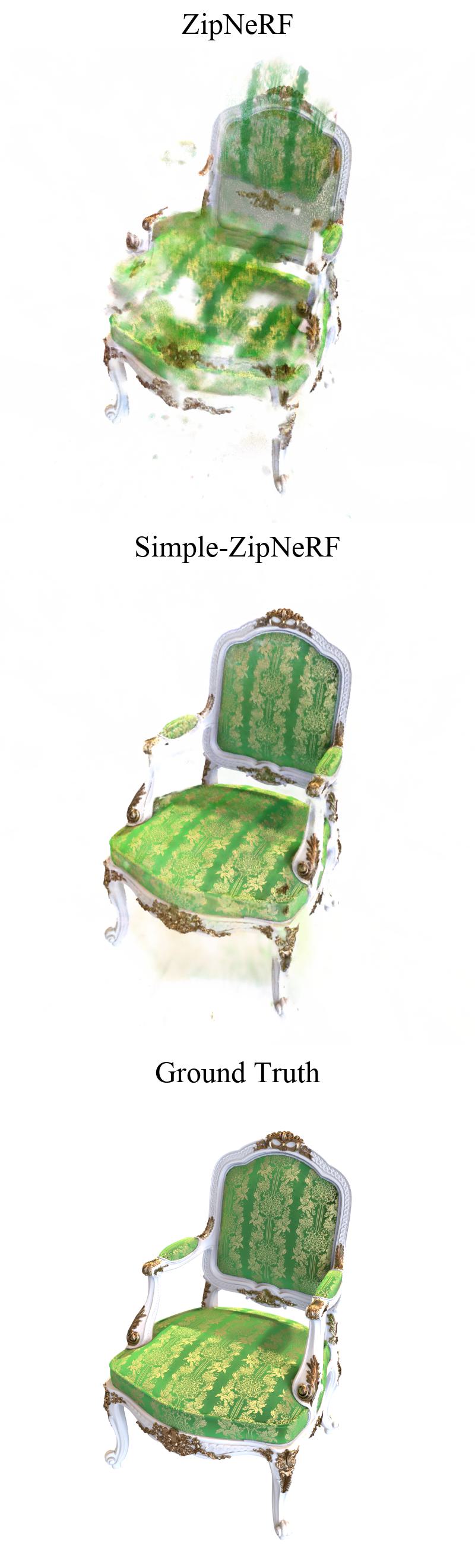}  
            \caption{8 input views.
            }
            \label{fig:qualitative-simple-zipnerf-nerfsynthetic02}
        \end{subfigure}
        \hfill
        \begin{subfigure}{0.32\linewidth}
            \centering
            \includegraphics[width=\linewidth]{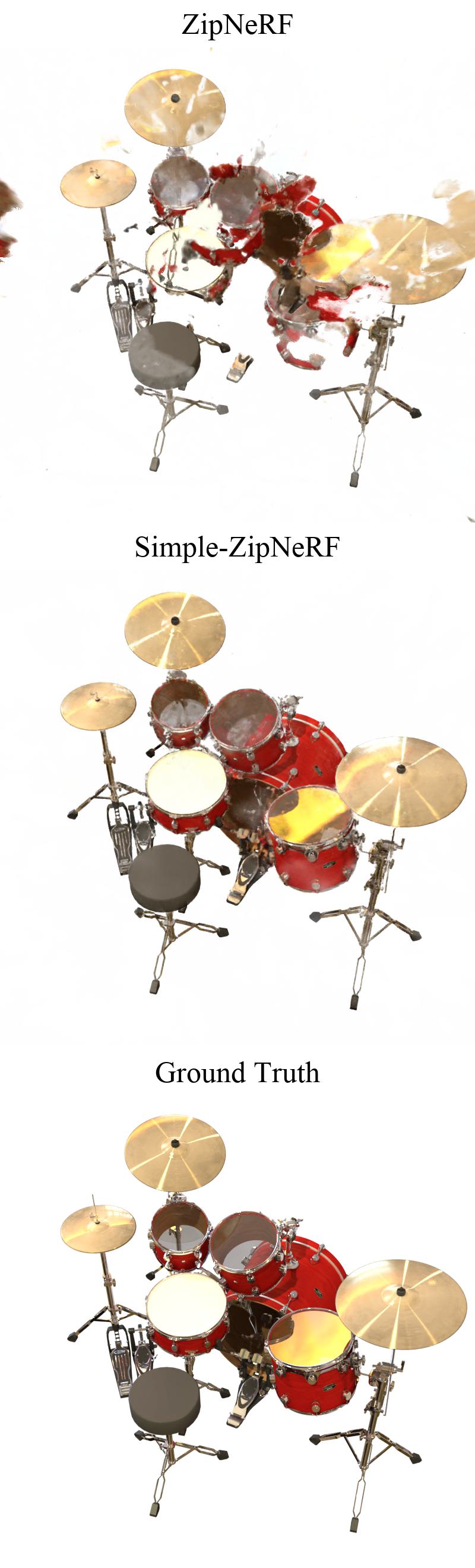}  
            \caption{12 input views.
            }
            \label{fig:qualitative-simple-zipnerf-nerfsynthetic03}
        \end{subfigure}
        \caption{\textbf{Qualitative examples of ZipNeRF and Simple-ZipNeRF on the NeRF-Synthetic dataset.}
        While the renders of ZipNeRF contain multiple floaters, Simple-ZipNeRF outputs are cleaner and free from such artifacts.
        }
        \label{fig:qualitative-simple-zipnerf-nerfsynthetic-010203}
    \end{figure}
}

\newcommand{\figureQualitativeSimpleZipnerfDepthFirst}{
    \begin{figure}
        \centering
        \includegraphics[width=\linewidth]{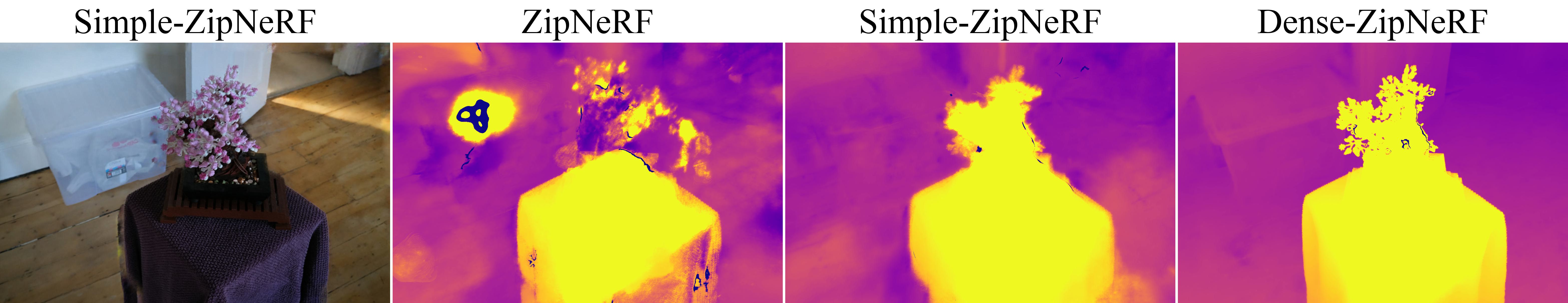}  
        \caption{\textbf{Simple-ZipNeRF estimated depth maps} on MipNeRF360 dataset with 20 input views.
        We observe that the depth map estimated by ZipNeRF contains floaters and that the depth estimates for the bonsai are incorrect.
        However, Simple-ZipNeRF does not suffer from such issues and the estimated depth is very close to that of ZipNeRF with dense input views.
        }
        \label{fig:qualitative-simple-zipnerf-depth}
    \end{figure}
}

\newcommand{\figureQualitativeSimpleZipnerfAblationsFirst}{
    \begin{figure*}
        \centering
        \includegraphics[width=\linewidth]{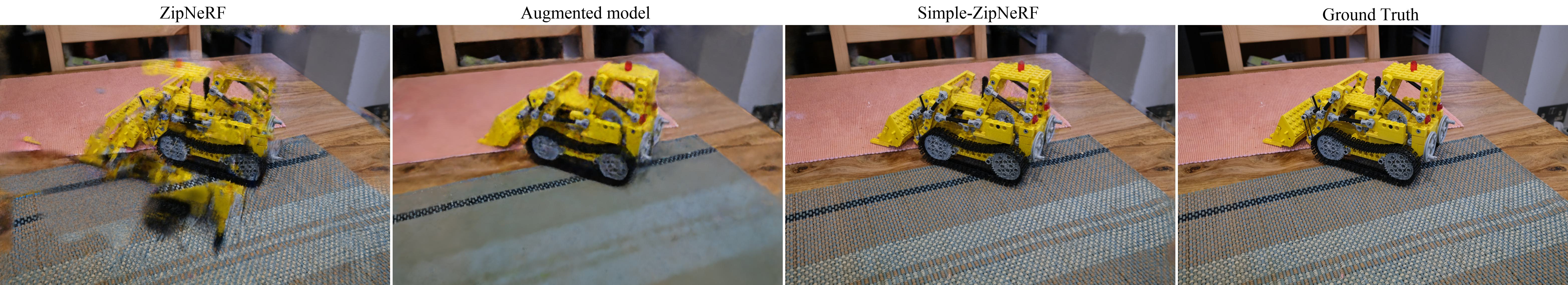}  
        \caption{\textbf{Qualitative examples to visualize the effect of our augmentation.}
        We observe that the ZipNeRF render contains severe distortions.
        The output of our augmented model is significantly better in reconstructing the scene, but the render contains severe blur on account of smoothing introduced by small hash table.
        Learning from the depth provided by the augmented model, Simple-ZipNeRF is able to reconstruct the scene better as well as retain sharpness by utilizing the larger hash table.
        }
        \label{fig:qualitative-simple-zipnerf-ablations01}
    \end{figure*}
}

\newcommand{\figureQuantitativeSimpleZipnerfAblationsSecond}{
    \begin{figure}
        \centering
        \begin{subfigure}{0.48\linewidth}
            \centering
            \includegraphics[width=\linewidth]{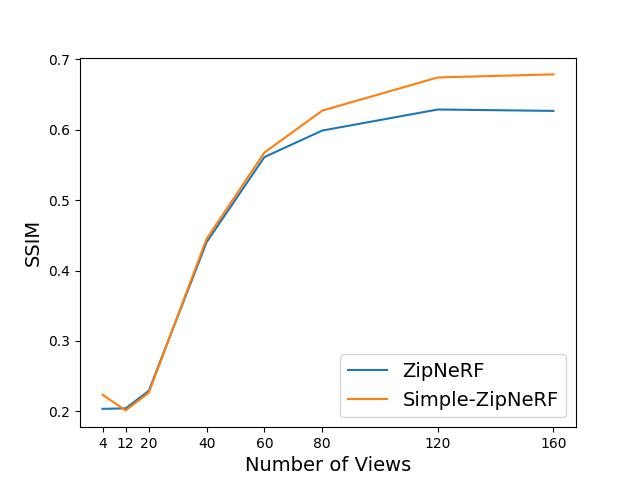}  
            \caption{MipNeRF360 Bicycle scene.
            }
            \label{fig:quantitative-simple-zipnerf-increasing-views01}
        \end{subfigure}
        \hfill
        \begin{subfigure}{0.48\linewidth}
            \centering
            \includegraphics[width=\linewidth]{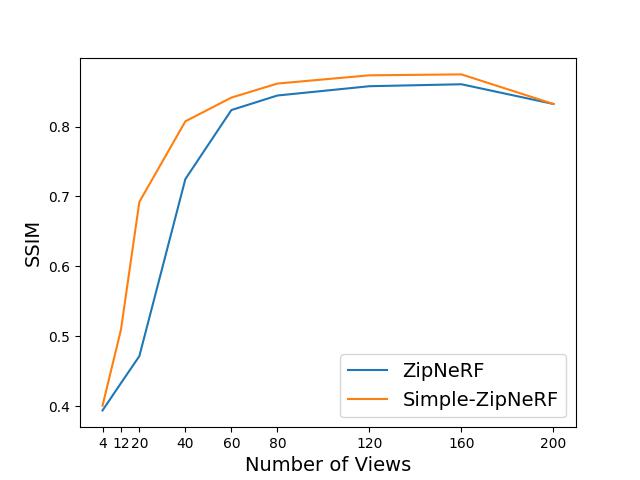}  
            \caption{MipNeRF360 Counter scene.
            }
            \label{fig:quantitative-simple-zipnerf-increasing-views02}
        \end{subfigure}
        \caption{\textbf{Performance of ZipNeRF and Simple-ZipNeRF with increasing number of input views.}
        We observe that our augmentation improves performance significantly over ZipNeRF, when the performance of the base model is moderate.
        When the performance of the base model is extremely poor or extremely good, the augmentation does not have a significant impact.
        However, our augmentation does not lead to significant degradation in performance in either case.
        }
        \label{fig:quantitative-simple-zipnerf-increasing-views}
    \end{figure}
}

\newcommand{\figureAnalysisSceneComplexityFirst}{
    \begin{figure*}
        \begin{subfigure}{0.48\textwidth}
            \centering
            \includegraphics[width=\textwidth]{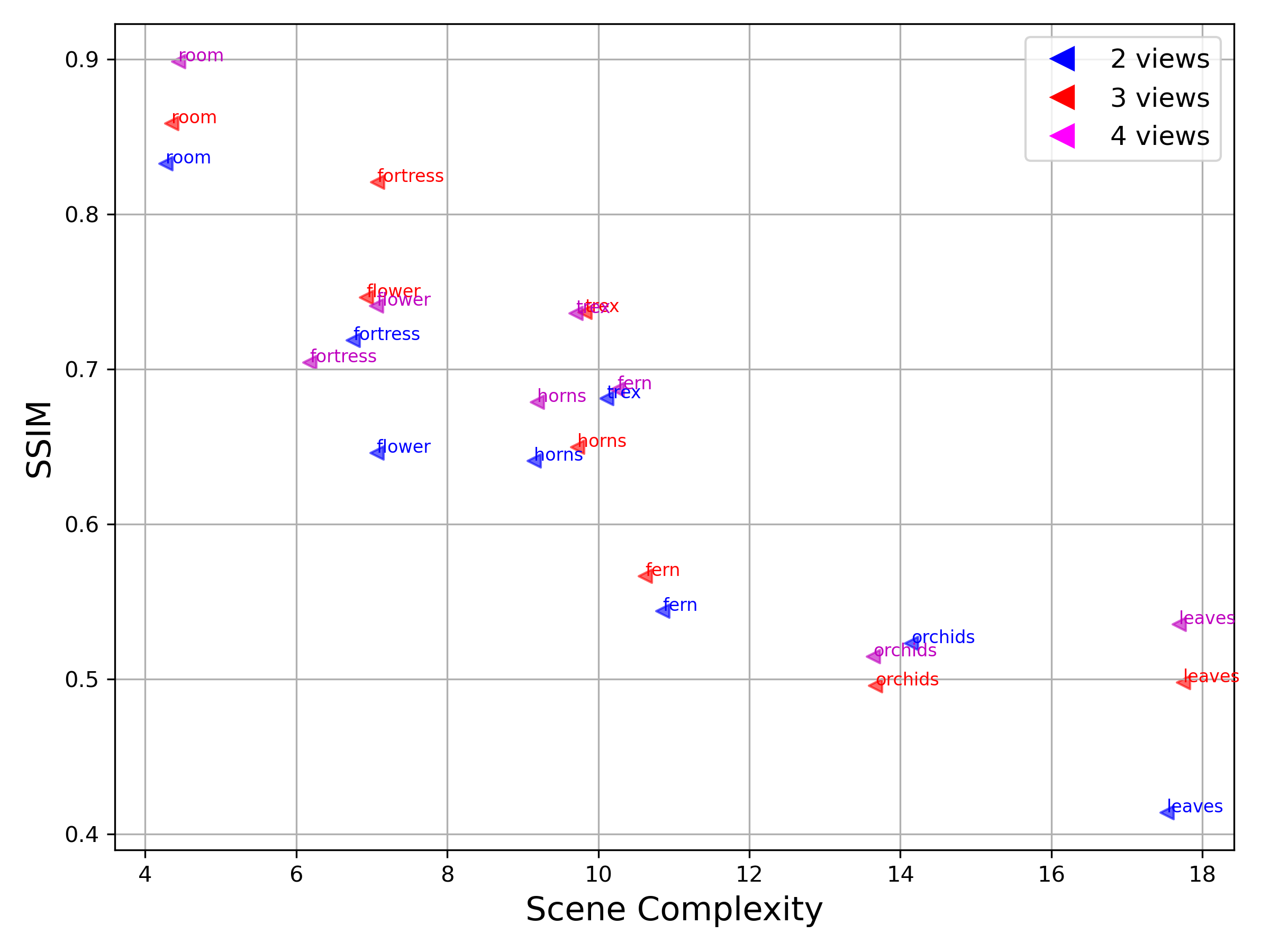}
            \caption{NeRF-LLFF Dataset}
            \label{fig:spatial-complexity-vs-performance-llff}
        \end{subfigure}
        \hfill
        \begin{subfigure}{0.48\textwidth}
            \centering
            \includegraphics[width=\textwidth]{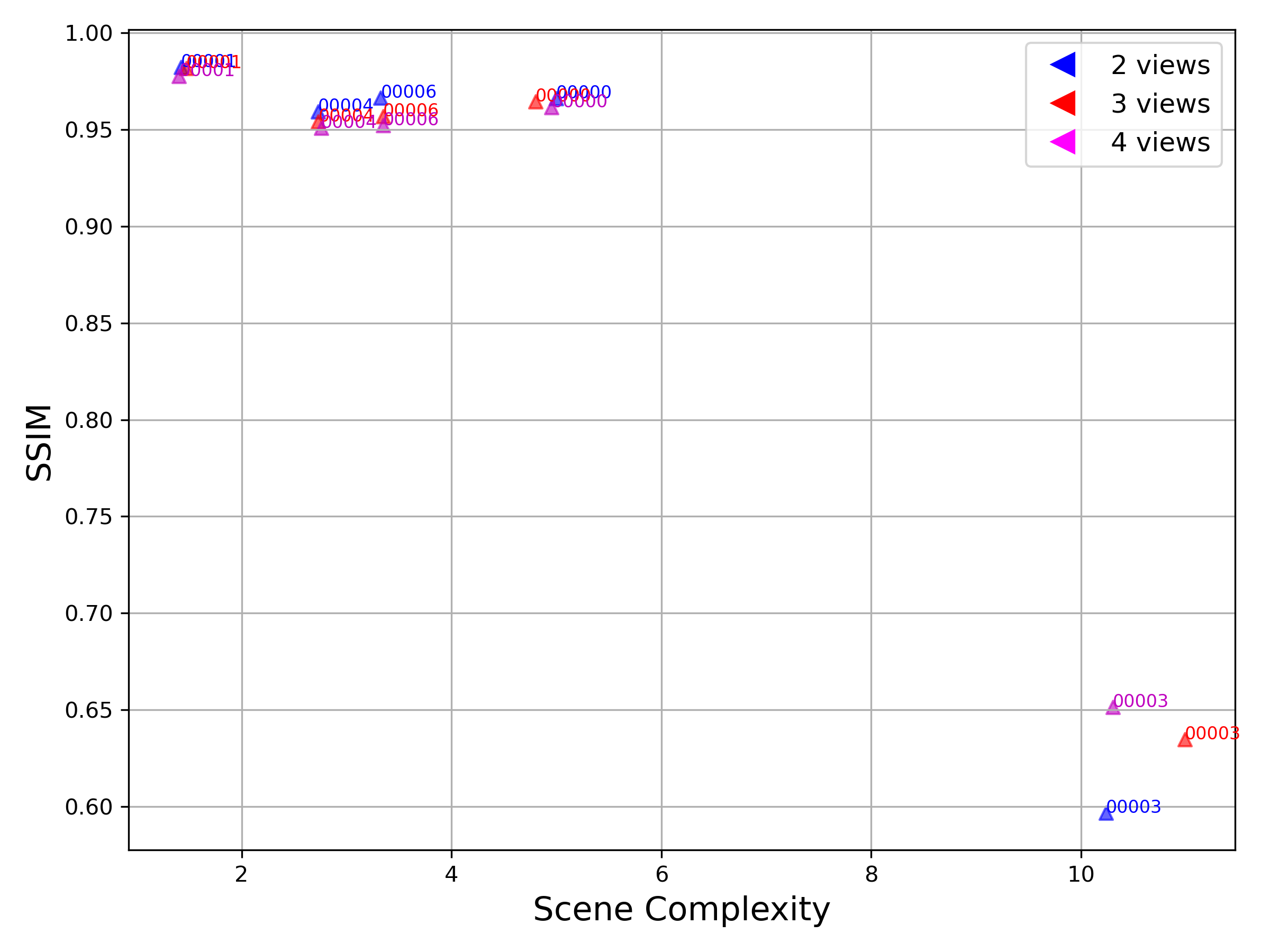}
            \caption{RealEstate10K Dataset}
            \label{fig:spatial-complexity-vs-performance-re10k}
        \end{subfigure}
        \caption{
        \textbf{Correlation between spatial complexity and the SSIM scores of Simple-NeRF across multiple scenes and number of input views.}
        We observe a strong negative correlation i.e., the performance of Simple-NeRF is very good for scenes with low spatial complexity and the performance reduces with increasing spatial complexity.
        }
        \label{fig:analysis-scene-complexity}
    \end{figure*}
}

\newcommand{\figureAnalysisPerformanceImprovementFirst}{
    \begin{figure}
        \centering
        \includegraphics[width=\linewidth]{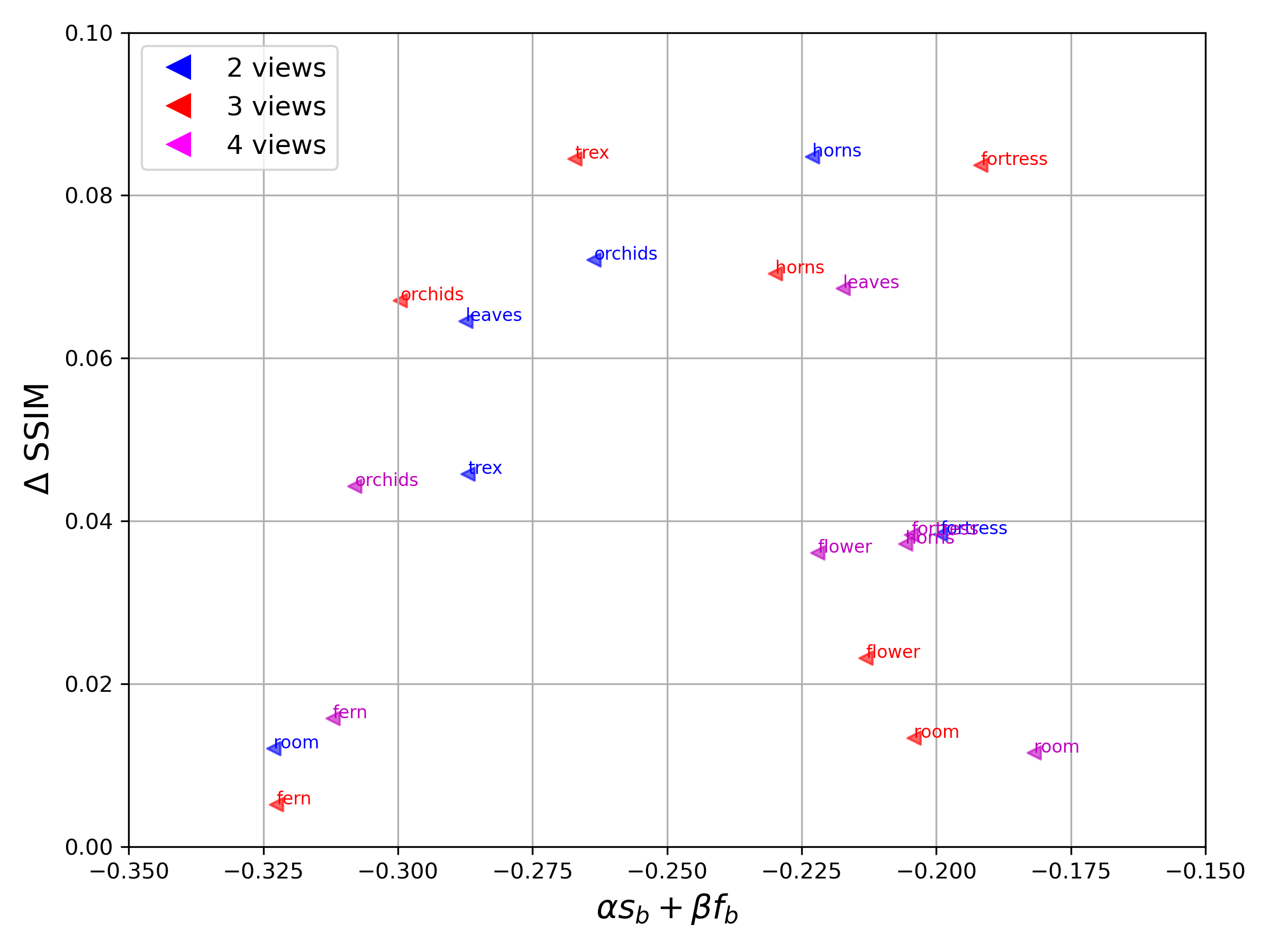}
        \caption{\textbf{Performance improvement of Simple-NeRF over DS-NeRF baseline with respect to the baseline performance.}
        We observe that Simple-NeRF provides significant improvement when the baseline performance of ZipNeRF is moderate.
        When the baseline performance is either very poor or very good, the improvement is marginal.
        }
        \label{fig:analysis-performance-improvement01}
    \end{figure}
}

\newcommand{\figureQualitativeSimpleZipnerfLimitationsFirst}{
    \begin{figure}
        \centering
        \begin{subfigure}{\linewidth}
            \centering
            \includegraphics[width=\linewidth]{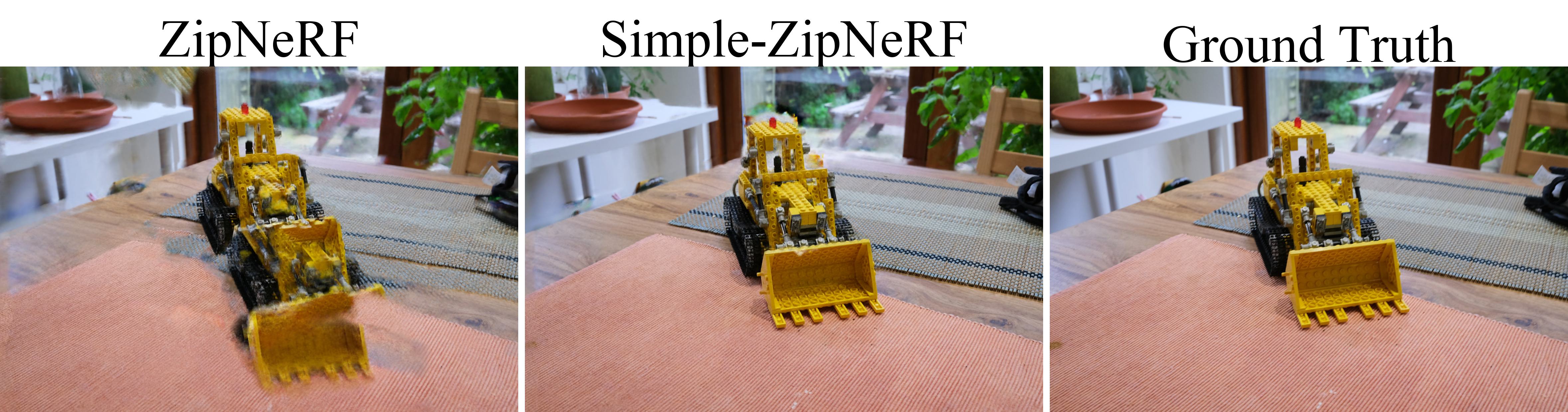}  
            \caption{MipNeRF360 Kitchen scene with 12 input views.
            }
            \label{fig:qualitative-simple-zipnerf-limitations01a}
        \end{subfigure}
        \begin{subfigure}{\linewidth}
            \centering
            \includegraphics[width=\linewidth]{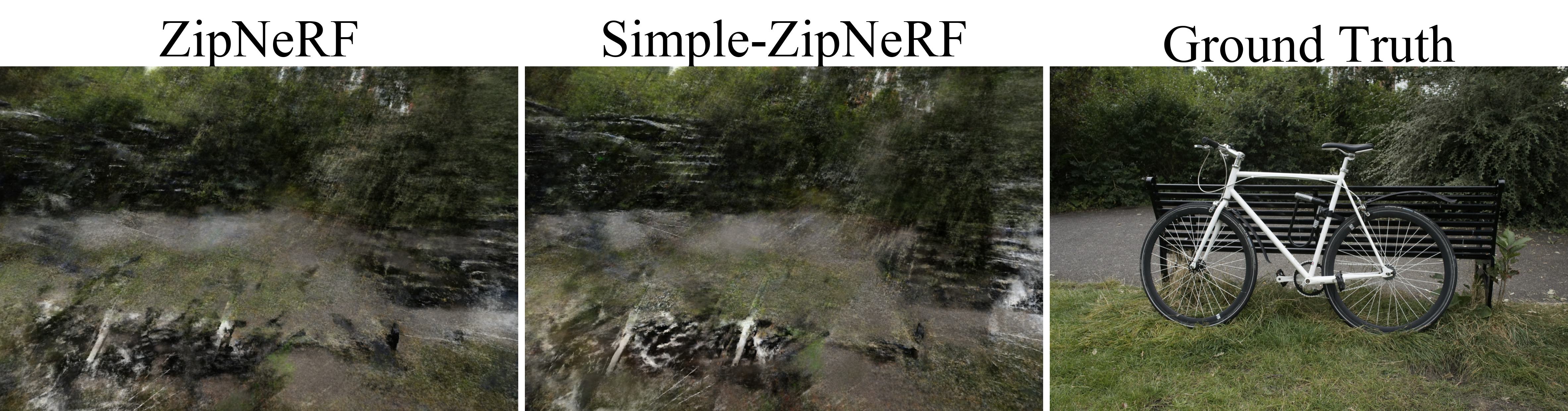}  
            \caption{MipNeRF360 Bicycle scene with 12 input views.
            }
            \label{fig:qualitative-simple-zipnerf-limitations01b}
        \end{subfigure}
        \caption{\textbf{Limitations of our augmentation.}
        With 12 input views, we observe that ZipNeRF is able to reconstruct the bulldozer in the kitchen scene, but has severe floater artifacts.
        In this case, our augmentation is successful in removing the floaters.
        However, ZipNeRF is unable to reconstruct the bicycle and the bench in the bicycle scene, where our augmentation is unable to help.
        }
        \label{fig:qualitative-simple-zipnerf-limitations01}
    \end{figure}
}

\newcommand{\tableRealEstateTrainTest}{
    \begin{table}
        \centering
        \caption{Train and test frame numbers of RealEstate-10K dataset used in the three different settings.}
        \begin{tabular}{|c|c|c|}
            \hline
            No. of i/p frames & Train frame nos.\ & Test frame nos.\ \\
            \hline
            2 & 10, 20        & 5--9, 11--19, 21--25 \\
            3 & 10, 20, 30    & 5--9, 11--19, 21--29, 31--35 \\
            4 & 0, 10, 20, 30 & 1--9, 11--19, 21--29, 31--35 \\
            \hline
        \end{tabular}
        \label{tab:realestate-train-test}
    \end{table}
}

\newcommand{\tableQuantitativeSimpleNerfLlff}{
    \begin{table*}
        \centering
        \caption{Quantitative results of NeRF and 3DGS based models on the NeRF-LLFF dataset.}
        \begin{tabular}{l|ccc|ccc|ccc}
            \hline
            & \multicolumn{3}{c|}{2 views} & \multicolumn{3}{c|}{3 views} & \multicolumn{3}{c}{4 views} \\
            Model &
            LPIPS \textdownarrow & SSIM \textuparrow & PSNR \textuparrow &
            LPIPS \textdownarrow & SSIM \textuparrow & PSNR \textuparrow &
            LPIPS \textdownarrow & SSIM \textuparrow & PSNR \textuparrow \\
            \hline
            InfoNeRF          & 0.6024 & 0.2219 & 9.16 & 0.6732 & 0.1953 & 8.37 & 0.6985 & 0.2270 & 9.18 \\
            DietNeRF          & 0.5465 & 0.3283 & 11.94 & 0.6120 & 0.3405 & 11.76 & 0.6506 & 0.3496 & 11.86 \\
            RegNeRF           & 0.3056 & 0.5712 & 18.52 & 0.2908 & 0.6334 & 20.22 & 0.2794 & 0.6645 & 21.32 \\
            FreeNeRF          & \textbf{0.2638} & 0.6322 & 19.52 & 0.2754 & 0.6583 & 20.93 & 0.2848 & 0.6764 & 21.91 \\
            DS-NeRF           & 0.3106 & 0.5862 & 18.24 & 0.3031 & 0.6321 & 20.20 & 0.2979 & 0.6582 & 21.23 \\
            DDP-NeRF          & 0.2851 & 0.6218 & 18.73 & 0.3250 & 0.6152 & 18.73 & 0.3042 & 0.6558 & 20.17 \\
            ViP-NeRF          & 0.2768 & 0.6225 & 18.61 & 0.2798 & 0.6548 & 20.54 & 0.2854 & 0.6675 & 20.75 \\  
            InstantSplat      & 0.2832 & 0.4627 & 16.93 & 0.2974 & 0.4603 & 17.49 & 0.3058 & 0.4818 & 17.81 \\
            CoR-GS            & 0.2644 & 0.6050 & 18.44 & \textbf{0.1860} & \textbf{0.7332} & 21.25 & \textbf{0.1693} & \textbf{0.7560} & \textbf{22.44} \\
            Simple-NeRF       & 0.2688 & \textbf{0.6501} & \textbf{19.57} & 0.2559 & 0.6940 & \textbf{21.37} & 0.2633 & 0.7016 & 21.99 \\  
            \hline
        \end{tabular}
        \label{tab:quantitative-simple-nerf-llff}
    \end{table*}
}

\newcommand{\tableQuantitativeSimpleNerfRealEstate}{
    \begin{table*}
        \centering
        \caption{Quantitative results of NeRF and 3DGS based models on the RealEstate-10K dataset.}
        \begin{tabular}{l|ccc|ccc|ccc}
            \hline
            & \multicolumn{3}{c|}{2 views} & \multicolumn{3}{c|}{3 views} & \multicolumn{3}{c}{4 views} \\
            Model &
            LPIPS \textdownarrow & SSIM \textuparrow & PSNR \textuparrow &
            LPIPS \textdownarrow & SSIM \textuparrow & PSNR \textuparrow &
            LPIPS \textdownarrow & SSIM \textuparrow & PSNR \textuparrow \\
            \hline
            InfoNeRF     & 0.5924 & 0.4342 & 12.27 & 0.6561 & 0.3792 & 10.57 & 0.6651 & 0.3843 & 10.62 \\
            DietNeRF     & 0.4381 & 0.6534 & 18.06 & 0.4636 & 0.6456 & 18.01 & 0.4853 & 0.6503 & 18.01 \\
            RegNeRF      & 0.4129 & 0.5916 & 17.14 & 0.4171 & 0.6132 & 17.86 & 0.4316 & 0.6257 & 18.34 \\
            DS-NeRF      & 0.2709 & 0.7983 & 26.26 & 0.2893 & 0.8004 & 26.50 & 0.3103 & 0.7999 & 26.65 \\
            DDP-NeRF     & 0.1290 & 0.8640 & 27.79 & 0.1518 & 0.8587 & 26.67 & 0.1563 & 0.8617 & 27.07 \\
            FreeNeRF     & 0.5036 & 0.5354 & 14.70 & 0.4635 & 0.5708 & 15.26 & 0.5226 & 0.6027 & 16.31 \\
            ViP-NeRF     & 0.0687 & 0.8889 & 32.32 & 0.0758 & 0.8967 & 31.93 & 0.0892 & 0.8968 & 31.95 \\  
            InstantSplat & 0.1338 & 0.8445 & 28.97 & 0.0925 & 0.8761 & 30.01 & 0.1618 & 0.8643 & 29.60 \\
            CoR-GS       & 0.2776 & 0.6840 & 20.22 & 0.2660 & 0.7131 & 20.58 & 0.2953 & 0.7119 & 20.33 \\
            Simple-NeRF  & \textbf{0.0635} & \textbf{0.8942} & \textbf{33.10} & \textbf{0.0726} & \textbf{0.8984} & \textbf{33.21} & \textbf{0.0847} & \textbf{0.8987} & \textbf{32.88} \\  
            \hline
        \end{tabular}
        \label{tab:quantitative-simple-nerf-realestate}
    \end{table*}
}

\newcommand{\tableQuantitativeSimpleNerfDepth}{
    \begin{table}
        \centering
        \caption{Evaluation of depth estimated by different NeRF based models with two input views.
        The reference depth is obtained using NeRF with dense input views.
        The depth MAE on the two datasets is of different orders on account of different depth ranges.
        }
        \begin{tabular}{l|cc|cc}
            \hline
            & \multicolumn{2}{c|}{NeRF-LLFF} & \multicolumn{2}{c}{RealEstate-10K} \\
            model &
            MAE \textdownarrow & SROCC \textuparrow &
            MAE \textdownarrow & SROCC \textuparrow \\
            \hline
            DS-NeRF    & 0.2074 & 0.7230 & 0.7164 & 0.6660 \\
            DDP-NeRF   & 0.2048 & 0.7480 & 0.4831 & 0.7921 \\
            ViP-NeRF   & 0.1999 & 0.7344 & 0.3856 & 0.8446 \\
            Simple-NeRF & \textbf{0.1420} & \textbf{0.8480} & \textbf{0.3269} & \textbf{0.9215} \\
            \hline
        \end{tabular}
        \label{tab:quantitative-simple-nerf-depth}
    \end{table}
}

\newcommand{\tableQuantitativeSimpleNerfAblations}{
    \begin{table}
        \centering
        \setlength\tabcolsep{0pt}
        \caption{SimpleNeRF ablation experiments on RealEstate-10K and NeRF-LLFF datasets with two input views.}
        \begin{tabular}{l|cc|cc}
            \hline
            & \multicolumn{2}{c|}{RealEstate-10K} & \multicolumn{2}{c}{NeRF-LLFF} \\
            model                & \textbf{LPIPS \textdownarrow} & \textbf{MAE \textdownarrow} & \textbf{LPIPS \textdownarrow} & \textbf{MAE \textdownarrow} \\
            \hline
            Simple-NeRF                     & \textbf{0.0635} & \textbf{0.33} & \textbf{0.2688} & \textbf{0.14} \\
            w/o smoothing augmentation      & 0.0752 & 0.38 & 0.2832 & 0.15 \\  
            w/o Lambertian augmentation     & 0.0790 & 0.39 & 0.2834 & 0.15 \\  
            w/o coarse-fine consistency     & 0.0740 & 0.42 & 0.3002 & 0.19 \\  
            w/o reliable depth              & 0.0687 & 0.45 & 0.3020 & 0.22 \\  
            w/o residual pos enc            & 0.0790 & 0.40 & 0.2837 & 0.16 \\  
            \hline
            w/ identical augmentations      & 0.0777 & 0.40 & 0.2849 & 0.15 \\  
            w/ smaller n/w as smoothing aug & 0.0740 & 0.38 & 0.2849 & 0.15 \\  
            \hline
        \end{tabular}
        \label{tab:quantitative-simple-nerf-ablations}
    \end{table}
}

\newcommand{\tableQuantitativeSimpleTensorf}{
    \begin{table*}
        \centering
        \setlength\tabcolsep{6pt}
        \caption{Quantitative results of TensoRF based models with three input views.}
        \begin{tabular}{l|ccccc|ccccc}
            \hline
            & \multicolumn{5}{c|}{NeRF-LLFF} & \multicolumn{5}{c}{RealEstate-10K} \\
            \textbf{Model} &
            LPIPS \textdownarrow & SSIM \textuparrow & PSNR \textuparrow &
            \makecell{Depth \\ MAE \textdownarrow} & \makecell{Depth \\ SROCC \textuparrow} &
            LPIPS \textdownarrow & SSIM \textuparrow & PSNR \textuparrow &
            \makecell{Depth \\ MAE \textdownarrow} & \makecell{Depth \\ SROCC \textuparrow} \\
            \hline
            TensoRF                     & 0.5474 & 0.3163 & 12.29 & 0.67 & 0.03 & 0.0986 & 0.8532 & 29.62 & 0.44 & 0.63 \\  
            DS-TensoRF                  & 0.2897 & 0.6291 & 18.58 & 0.23 & 0.73 & 0.0739 & 0.8872 & 32.50 & 0.27 & 0.75 \\  
            Simple-TensoRF              & \textbf{0.2461} & \textbf{0.6749} & \textbf{20.22} & \textbf{0.17} & \textbf{0.83} & \textbf{0.0706} & \textbf{0.8920} & \textbf{32.70} & \textbf{0.22} & 0.80 \\  
            \hline
            $R_\sigma^s = R_\sigma$  & 0.2536 & 0.6677 & 19.85 & 0.18 & 0.81 & 0.085 & 0.8821 & 30.94 & 0.27 & 0.77 \\  
            $\numTensorfVoxels[s] = \numTensorfVoxels$ & 0.2568 & 0.6579 & 19.95 & 0.19 & 0.79 & 0.0735 & 0.8896 & 32.22 & \textbf{0.22} & \textbf{0.82} \\  
            $R_\sigma^s = R_\sigma; \numTensorfVoxels[s] = \numTensorfVoxels$ & 0.2728 & 0.6424 & 19.50 & 0.22 & 0.74 & 0.0787 & 0.8871 & 31.73 & 0.23 & 0.79 \\  
            \hline
        \end{tabular}
        \label{tab:quantitative-simple-tensorf}
    \end{table*}
}

\newcommand{\tableQuantitativeSimpleZipnerfMipNerf}{
    \begin{table*}
        \centering
        \setlength\tabcolsep{3pt}
        \caption{Quantitative results of ZipNeRF and 3DGS based models on the MipNeRF360 dataset.}
        \begin{tabular}{l|ccccc|ccc|ccc}
            \hline
            & \multicolumn{5}{c|}{12 input views} & \multicolumn{3}{c|}{20 input views} & \multicolumn{3}{c}{36 input views} \\
            \textbf{Model} &
            LPIPS \textdownarrow & SSIM \textuparrow & PSNR \textuparrow & \makecell{Depth \\ MAE \textdownarrow} & \makecell{Depth \\ SROCC \textuparrow} &
            LPIPS \textdownarrow & SSIM \textuparrow & PSNR \textuparrow &
            LPIPS \textdownarrow & SSIM \textuparrow & PSNR \textuparrow \\
            \hline
            ZipNeRF                              & 0.5614 & 0.4616 & 15.86 & 7.43 & 0.28 & 0.435 & 0.5911 & 18.89 & 0.3316 & 0.6737 & 21.78 \\ 
            InstantSplat                         & \textbf{0.4503} & 0.5204 & 17.57 &  --   &  --  & 0.4816 & 0.5545 & 18.02 & 0.6268 & 0.5461 & 16.37 \\
            CoR-GS                               & 0.5039 & \textbf{0.5834} & \textbf{18.32} &  --   &  --  & 0.3500 & \textbf{0.7008} & \textbf{21.98} & \textbf{0.2239} & \textbf{0.7907} & \textbf{24.98} \\
            Augmented ZipNeRF                    & 0.6825 & 0.4462 & 16.27 & 96.42 & 0.49 & 0.6190 & 0.5244 & 19.31 & 0.5917 & 0.5646 & 21.21 \\ 
            Simple-ZipNeRF                       & 0.4876 & 0.5245 & 17.60 & \textbf{3.54} & \textbf{0.51} & \textbf{0.3421} & 0.6456 & 21.03 & 0.2390 & 0.7458 & 24.19 \\ 
            \hline
        \end{tabular}
        \label{tab:quantitative-simple-zipnerf-mipnerf360}
    \end{table*}
}

\newcommand{\tableQuantitativeSimpleZipnerfNerfSynthetic}{
    \begin{table*}
        \centering
        \setlength\tabcolsep{3pt}
        \caption{Quantitative results of ZipNeRF based models on the NeRF-Synthetic dataset.}
        \begin{tabular}{l|ccc|ccc|ccc}
            \hline
            & \multicolumn{3}{c|}{4 input views} & \multicolumn{3}{c|}{8 input views} & \multicolumn{3}{c}{12 input views} \\
            \textbf{Model} &
            LPIPS \textdownarrow & SSIM \textuparrow & PSNR \textuparrow &
            LPIPS \textdownarrow & SSIM \textuparrow & PSNR \textuparrow &
            LPIPS \textdownarrow & SSIM \textuparrow & PSNR \textuparrow \\
            \hline
            ZipNeRF                              & 0.4263 & 0.7548 & 11.04 & 0.2877 & 0.7973 & 15.01 & 0.1625 & 0.8528 & 20.12 \\ 
            Simple-ZipNeRF                       & \textbf{0.3878} & \textbf{0.7715} & \textbf{11.50} & \textbf{0.2461} & \textbf{0.8063} & \textbf{15.88} & \textbf{0.1532} & \textbf{0.8531} & \textbf{20.51} \\ 
            \hline
        \end{tabular}
        \label{tab:quantitative-simple-zipnerf-nerf-synthetic}
    \end{table*}
}

\newcommand{\tableComparisonSpaceTime}{
    \begin{table}
        \centering
        \setlength\tabcolsep{3pt}
        \caption{Training and inference (per frame) time and memory comparison of various models.}
        \begin{tabular}{l|cc|cc}
            \hline
            & \multicolumn{2}{c|}{Training} & \multicolumn{2}{c}{Inference} \\
            Model & Time (hrs) & Mem (GB) & Time (sec) & Mem (GB) \\
            \hline
            NeRF            & 14 & 6.1 & 54 & 0.8 \\
            Simple-NeRF     & 21 & 8.8 & 54 & 0.8 \\
            \hline
            TensoRF         & 2.1 & 6.8 & 21 & 4.0 \\
            Simple-TensoRF  & 3.7 & 7.2 & 21 & 4.0 \\
            \hline
            ZipNeRF         & 2.0 & 6.7 & 13 & 4.8 \\
            Simple-ZipNeRF  & 4.2 & 8.6 & 13 & 4.8 \\
            \hline
        \end{tabular}
        \label{tab:comparison-space-time}
    \end{table}
}

\begin{document}
    \title{Simple-RF: Regularizing Sparse Input Radiance Fields with Simpler Solutions}

\author{Nagabhushan Somraj}
\orcid{0000-0002-2266-759X}
\affiliation{%
  \institution{Indian Institute of Science}
  \city{Bengaluru}
  \state{Karnataka}
  \postcode{560012}
  \country{India}
}
\email{nagabhushans@iisc.ac.in}
\author{Sai Harsha Mupparaju}
\affiliation{%
  \institution{Indian Institute of Science}
  \city{Bengaluru}
  \state{Karnataka}
  \country{India}
}
\email{saiharsham@iisc.ac.in}
\author{Adithyan Karanayil}
\affiliation{%
  \institution{Indian Institute of Science}
  \city{Bengaluru}
  \state{Karnataka}
  \country{India}
}
\email{adithyanv@iisc.ac.in}
\author{Rajiv Soundararajan}
\affiliation{%
 \institution{Indian Institute of Science}
 \city{Bengaluru}
 \state{Karnataka}
 \country{India}
}
\email{rajivs@iisc.ac.in}

    \begin{abstract}
        Neural Radiance Fields (NeRF) show impressive performances in photo-realistic free-view rendering of scenes.
        Recent improvements such as TensoRF and ZipNeRF employ explicit models for faster optimization and rendering.
        However, all these radiance fields require a dense sampling of images in the given scene for effective training.
        Their performances degrade significantly when only a sparse set of views is available.
        Existing depth priors used to supervise the radiance fields are either sparse or suffer from generalization issues.
        We seek to learn scene-specific dense depth priors to regularize the radiance fields.
        Further, we desire a framework of regularizations that can work across different radiance field models.
        We observe that certain features of the radiance fields, such as positional encoding, number of decomposed tensor components or size of the hash table, cause overfitting in the sparse-input scenario.
        We design augmented models by reducing the capacity of these features and train them along with the main radiance field.
        These augmented models learn simpler solutions, which estimate better depth in certain regions.
        By supervising the main radiance field with such depths, we significantly improve the performance of the radiance fields on popular forward-facing and 360$^\circ$ datasets by employing the above regularization.
    \end{abstract}

%
%


%
%


    \figureTeaser

    \maketitle

    \section{Introduction}\label{sec:introduction}
    Neural Radiance Fields (NeRFs)~\cite{mildenhall2020nerf} show unprecedented levels of performance in synthesizing novel views of a scene by learning a volumetric representation implicitly within the weights of multi-layer perceptrons (MLP).
    Although NeRFs are very promising for view synthesis, there is a need to improve their design in a wide array of scenarios.
    For example, NeRFs have been enhanced to optimize and render quickly~\cite{muller2022instant}, reduce aliasing artifacts~\cite{barron2021mipnerf}, and learn on unbounded scenes~\cite{barron2022mipnerf360}.
    Yet all these models require tens to hundreds of images per scene to learn the scene geometry accurately, and their quality deteriorates significantly when only a few training images are available~\cite{jain2021dietnerf}.
    In this work, we focus on training both implicit radiance fields such as NeRF and explicit radiance fields such as TensoRF~\cite{chen2022tensorf} and ZipNeRF~\cite{barron2023zipnerf} with a sparse set of input images and aim to design novel regularizations for effective training.

    Researchers have extensively studied the training of NeRFs with sparse input views.
    One approach to training NeRFs with sparse input views is to use generalized NeRFs, where the NeRF is additionally conditioned on a latent scene representation obtained using a convolutional neural network.
    However, these models require a large multi-view dataset for pre-training and may suffer from generalization issues when used to render a novel scene~\cite{niemeyer2022regnerf}.
    The other thread of work on sparse input NeRFs follows the original NeRF paradigm of training scene-specific NeRFs, and designs novel regularizations to assist NeRFs in converging to a better scene geometry~\cite{zhang2021ners,ni2024colnerf,guo2024depth}.
    One popular approach among such models is to supervise the depth estimated by the NeRF\@.
    RegNeRF~\cite{niemeyer2022regnerf}, DS-NeRF~\cite{deng2022dsnerf} and ViP-NeRF~\cite{somraj2023vipnerf} use simple priors such as depth smoothness, sparse depth or relative depth respectively obtained through classical methods.
    On the other hand, DDP-NeRF~\cite{roessle2022ddpnerf} and SCADE~\cite{uy2023scade} pre-train convolutional neural networks (CNN) on a large dataset of scenes to learn a dense depth prior.
    These approaches may also suffer from issues similar to those of the generalized models.
    This raises the question of whether we can instead learn the dense depth supervision in-situ without employing any pre-training.

    Recently, there is also a growing interest in sparse input explicit radiance fields owing to their fast optimization and rendering times.
    However, the regularizations designed for these models are limited to a specific explicit radiance field and do not generalize to more recent models.
    For example, while the regularizations designed for ZeroRF~\cite{shi2024zerorf} can be applied to TensoRF based models only, other regularizations are applicable to implicit models only~\cite{yang2023freenerf,zhu2024vanilla}.
    It is desirable to design regularizations that are relevant to different radiance field models through a single framework.
    While there exists a plethora of implicit and explicit radiance field models, we consider the NeRF as the representative model for implicit radiance fields and consider two explicit radiance fields, namely, TensoRF~\cite{chen2022tensorf} and ZipNeRF~\cite{barron2023zipnerf}.
    We note that although the NeRF based models are slow in optimization and rendering, NeRFs are versatile in learning different properties of the scenes ~\cite{verbin2022refnerf,bi2020neural,srinivasan2021nerv,zhang2021nerfactor} and may be easier to optimize as compared to the explicit models.
    Hence, we believe that designing regularizations for sparse-input NeRF is also of considerable interest.

    We first observe that the radiance field models often exploit their high capability to learn unnecessary complex solutions when training with sparse input views.
    While these solutions perfectly explain the observed images, they can cause severe distortions in novel views.
    For example, some of the key components of the radiance fields, such as positional encoding in the NeRF or vector-matrix decomposition employed in TensoRF, provide powerful capabilities to the radiance field and are designed for training the model with dense input views.
    Existing implementations of these components may be sub-optimal with fewer input views due to the highly under-constrained system, causing several distortions.
    \cref{fig:distortions-sparse-nerf,fig:distortions-sparse-tensorf,fig:distortions-sparse-zipnerf} show common distortions observed with NeRF, TensoRF and ZipNeRF in the few-shot setting respectively.
    We follow the popular Occam's razor principle and regularize the radiance fields to choose simpler solutions over complex ones, wherever possible.
    In particular, we design augmented models by reducing the capabilities of the radiance fields and use the depth estimated by these models to supervise the main radiance field.

    We design different augmentations for NeRF, TensoRF and ZipNeRF based on different shortcomings and architectures of these models.
    The high positional encoding degree used in the NeRF leads to undesired depth discontinuities, creating floaters.
    Further, the view-dependent radiance feature leads to shape-radiance ambiguity, creating duplication artifacts.
    We design augmentations for the NeRF by reducing the positional encoding degree and disabling the view-dependent radiance feature.
    On the other hand, the large number of high-resolution factorized components in TensoRF and the large hash table in ZipNeRF cause floaters in these models in the few-shot setting.
    The unifying theme across the three models is that we design augmentations to constrain the model with respect to such components to learn simpler solutions.

    We use the simplified models as augmentations for depth supervision and not as the main NeRF model since na\"{\i}vely reducing the capacity of the radiance fields may lead to sub-optimal solutions in certain regions~\cite{jain2021dietnerf}.
    For example, the model that can learn only smooth depth transitions may fail to learn sharp depth discontinuities at object boundaries.
    Further, the augmented models need to be used for supervision only if they explain the observed images accurately.
    We gauge the reliability of the depths by reprojecting pixels using the estimated depths onto a different nearest train view and comparing them with the corresponding images.

    We refer to our family of regularized models as Simple Radiance Fields (Simple-RF) since we regularize the models to choose simple solutions over complex ones, wherever feasible.
    We refer to the individual models as Simple-NeRF, Simple-TensoRF and Simple-ZipNeRF respectively.
    We evaluate our models on four popular datasets that include forward-facing scenes (NeRF-LLFF), unbounded forward-facing scenes (RealEstate-10K), unbounded 360$\degree$ scenes (MipNeRF360) and bounded 360$\degree$ scenes (NeRF-Synthetic) and show that our models achieve significant improvement in performance on all the datasets.
    \cref{fig:teaser} qualitatively shows the improvements achieved through our regularizations on NeRF, TensoRF and ZipNeRF on three different datasets.
    Further, we show that our model learns geometry significantly better than prior art.

    We list the main contributions of our work in the following.
    \begin{itemize}
        \item We find that the high capability of various radiance field models lead to learning on unnecessary complex geometries when training with sparse input views.
        We design specific augmented models that learn simpler solutions to regularize the different radiance field models.
        In particular,
        \begin{itemize}
            \item We find that the high positional encoding degree and view-dependent radiance of the NeRF cause floater and duplication artifacts when training with sparse inputs.
            We design augmented models on both these fronts to supervise the main NeRF and mitigate both artifacts.
            \item We observe that the large number of high-resolution decomposed components in TensoRF leads to floater artifacts with sparse inputs.
            Thus, the augmented model is obtained by reducing the number and resolutions of the decomposed components.
            \item We find that the large hash table in ZipNeRF causes floaters when training with sparse inputs.
            The augmented model is designed by reducing the size of the hash table.
        \end{itemize}
        \item We design a mechanism to determine whether the depths estimated by the augmented models are accurate and utilize only the accurate estimates to supervise the main radiance field.
    \end{itemize}

    \section{Related Work}\label{sec:related-work}
    \citet{chen1993view} introduce the problem of novel view synthesis and propose an image-based rendering (IBR) approach to synthesize novel views.
    The follow-up approaches introduce the geometry of the scene for synthesizing novel views through approximate representations such as light fields~\cite{levoy1996light}, lumigraphs~\cite{gortler1996lumigraph}, plenoptic functions~\cite{mcmillan1995plenoptic} and layered depth images~\cite{shade1998ldi}.
    \citet{chai2000plenoptic} study the minimum sampling needed for light field rendering and also show that depth information enables better view synthesis with sparse viewpoints.
    \citet{mcmillan1997image} and \citet{mark1999postrendering} introduce depth image based rendering (DIBR) to synthesize new views.
    Multiple variants of DIBR~\cite{chaurasia2013depth,sun2010overview,kanchana2022ivp} find use in various applications such as 3D-TV~\cite{fehn2004depth} and free-viewpoint video~\cite{carranza2003free,smolic20063d,collet2015high}.
    \citet{ramamoorthi2023nerfs} conducts a detailed survey on classical work for novel view synthesis.

    With the advent of deep learning, volumetric models utilize the power of learning by training the model on a large dataset of multi-view images.
    While the early approaches predict volumetric representations in each of the target views~\cite{flynn2016deepstereo,kalantari2016learning}, latter approaches predict a single volumetric representation and warp the representation to the target view while rendering~\cite{penner2017soft3d,zhou2018stereomag,srinivasan2019pushing,mildenhall2019llff,shih20203dp}.
    However, these approaches employ discrete depth planes and hence suffer from discretization artifacts.
    The seminal work by \citet{mildenhall2020nerf} employ a continuous representation using multi-layer perceptrons (MLP).
    This started a new pathway in neural view synthesis.
    However, these models suffer from two major limitations, namely, the need for the dense sampling of input views and the large time required to render novel views from the given input views.
    The prior work that address these limitations can be broadly classified into three categories.
    In \cref{subsec:implicit-radiance-fields}, we review various approaches in the literature to regularize the NeRF when training with sparse input views.
    We review the explicit radiance fields that aim at fast optimization and rendering in \cref{subsec:explicit-radiance-fields}, and also review the recent work on regularizing explicit models for the few-shot setting.
    Finally, in \cref{subsec:generalized-sparse-input-nerf}, we review the generalized NeRFs that address both issues jointly.

    \subsection{Implicit Radiance Fields}\label{subsec:implicit-radiance-fields}
    There exists extensive literature on regularizing scene-specific NeRFs when training with sparse inputs.
    Hence, we further group these models based on their approaches.

    \emph{Hand-Crafted Depth Priors:}
    The prior work on sparse input NeRFs explore a plethora of hand-crafted priors on the NeRF rendered depth.
    RegNeRF~\cite{niemeyer2022regnerf} imposes a smoothness constraint on the rendered depth maps.
    DS-NeRF~\cite{deng2022dsnerf} uses sparse depth provided by a Structure from Motion (SfM) module to supervise the NeRF estimated depth at sparse keypoints.
    ViP-NeRF~\cite{somraj2023vipnerf} augments the sparse-depth regularization of DS-NeRF with a regularization on the relative depth of objects by obtaining a prior on the visibility of objects.
    HG3-NeRF~\cite{gao2024hg3nerf} uses sparse depth given by colmap to guide the sampling 3D points instead of supervising the NeRF rendered depth.
    While these priors are more robust across different scenes, they do not exploit the power of learning.

    \emph{Deep Learning Based Depth Priors:}
    There exist multiple models that utilize the advances in dense depth estimation using deep neural networks.
    DDP-NeRF~\cite{roessle2022ddpnerf} extends DS-NeRF by employing a CNN to complete the sparse depth into dense depth for more supervision.
    SCADE~\cite{uy2023scade} and SparseNeRF~\cite{wang2023sparsenerf} use the depth map output by single image depth models to constrain the absolute and the relative order of pixel depths, respectively.
    DiffusioNeRF~\cite{wynn2023diffusionerf} learns the joint distribution of RGBD patches using denoising diffusion models (DDM) and utilizes the gradient of the distribution provided by the DDM to regularize NeRF rendered RGBD patches.
    However, the deep-learning based priors require pre-training on a large dataset and may suffer from generalization issues when obtaining the prior on unseen test scenes.
    Our work obtains depth supervision by harnessing the power of learning without suffering from generalization issues by employing and training augmented models on the given scene alone.
    In addition, our approach imposes depth supervision selectively i.e. only in the regions where the depth supervision is determined to be reliable.

    \emph{View Hallucination based Methods:}
    Another line of regularization based approaches simulate dense sampling by hallucinating new viewpoints and regularizing the NeRF on different aspects such as semantic consistency~\cite{jain2021dietnerf}, depth smoothness~\cite{niemeyer2022regnerf}, sparsity of mass~\cite{kim2022infonerf} and depth based reprojection consistency~\cite{kwak2023geconerf,xu2022sinnerf,bortolon2022dvm,chen2022geoaug}.
    Instead of sampling new viewpoints randomly, FlipNeRF~\cite{seo2023flipnerf} utilizes ray reflections to determine new viewpoints.
    Deceptive-NeRF~\cite{liu2023deceptive} and ReconFusion~\cite{wu2024reconfusion} employ a diffusion model to generate images in hallucinated views and use the generated views in addition to the input views to train the NeRF\@.
    However, supervision with generative models could lead to content hallucinations, leading to poor fidelity~\cite{lee2024posediff}.

    \emph{Other regularizations:}
    A few models also explore regularizations other than depth supervision and view hallucinations.
    FreeNeRF \cite{yang2023freenerf} and MI-MLP-NeRF~\cite{zhu2024vanilla} regularize the NeRF by modifying the inputs.
    Specifically, FreeNeRF anneals the frequency range of positional encoded NeRF inputs as the training progresses, and MI-MLP-NeRF adds the 5D inputs to every layer of the NeRF MLP\@.
    MixNeRF~\cite{seo2023mixnerf} models the volume density along a ray as a mixture of Laplacian distributions.
    \citet{ritschel2023gradientscaling} scale the gradients corresponding to 3D points close to the camera when sampling the 3D points in inverse depth to reduce floaters close to the camera.
    VDN-NeRF~\cite{zhu2023vdnnerf} on the other hand, aims to resolve shape-radiance ambiguity in the case of dense input views.
    However, these approaches are designed for specific cases and are either sub-optimal or do not extend to more recent radiance field models.

    \subsection{Explicit Radiance Fields}\label{subsec:explicit-radiance-fields}
    The NeRF takes a long time to optimize and render novel views due to the need to query the NeRF MLP hundreds of times to render a single pixel.
    Hence, a common approach to fast optimization and rendering is to reduce the time taken per query.
    Early works such as PlenOctress~\cite{yu2021plenoctrees} and KiloNeRF~\cite{reiser2021kilonerf} focus on improving only the rendering time by baking the trained NeRF into an explicit structure such as Octrees or thousands of tiny MLPs.
    PlenOxels~\cite{fridovich2022plenoxels} and DVGO~\cite{sun2022dvgo} reduce the optimization time by directly optimizing voxel grids, but at the cost of large memory requirements to store the voxel grids.
    TensoRF~\cite{chen2022tensorf} and K-Planes~\cite{fridovich2023kplanes} reduce the memory consumption using factorized tensors that exploit the spatial correlation of the radiance field.
    Alternately, iNGP~\cite{muller2022instant} and ZipNeRF~\cite{barron2023zipnerf} employ multi-resolution hash-grids to reduce the memory consumption.
    Recently, 3DGS~\cite{kerbl20233dgs} propose an alternative volumetric model for real-time rendering of novel views.
    Specifically, 3DGS employs 3D Gaussians to represent the scene and renders a view by splatting the Gaussians onto the corresponding image plane.
    While the above methods enable fast optimization and rendering, their performance still reduces significantly with fewer input views.

    Recently, there is increasing interest in regularizing explicit models to learn with sparse inputs~\cite{li2024dngaussian,yang2024gaussianobject}.
    However, the regularizations designed in these models are limited to a specific explicit radiance field and do not generalize to other explicit models.
    For example, ZeroRF~\cite{shi2024zerorf} imposes a deep image prior~\cite{ulyanov2018deepimageprior} on the components of the TensoRF~\cite{chen2022tensorf} model.
    FSGS~\cite{zhu2023fsgs} and SparseGS~\cite{xiong2023sparsegs} improve the performance of 3DGS~\cite{kerbl20233dgs} in the sparse input case by improving the initialization of the 3D Gaussian point cloud and pruning Gaussians responsible for floaters respectively.
    On the other hand, our approach of regularizing with simpler solutions is applicable to various explicit models, such as TensoRF, iNGP and ZipNeRF as well as to implicit models such as NeRF\@.

    Despite the recent work on sparse input 3DGS models, we do not explore designing augmentations for 3DGS\@.
    As noted in the recent literature, 3DGS mainly suffers from poor initialization with few input views~\cite{chen2024mvsplat}.
    We believe 3DGS requires a combination of good initialization and supervision from augmentations to learn from few input views.
    This necessitates a separate study on designing better initializations for 3DGS, which is beyond the scope of this work.

    \subsection{Generalized Sparse Input NeRF}\label{subsec:generalized-sparse-input-nerf}
    Obtaining a volumetric model of a scene by optimizing the NeRF is a time-consuming process.
    In order to reduce the time required to obtain a volumetric model of a scene and learn with fewer input views, generalized NeRF models train a neural network on a large dataset of multi-view scenes that can be directly applied to a test scene without any optimization~\cite{chen2021mvsnerf,tancik2021metanerf,lee2023extremenerf}.
    Early pieces of work such as PixelNeRF~\cite{yu2021pixelnerf}, GRF~\cite{trevithick2021grf}, and IBRNet~\cite{wang2021ibrnet} obtain convolutional features of the input images and additionally condition the NeRF by projecting the 3D points onto the feature grids.
    MVSNeRF~\cite{chen2021mvsnerf} incorporates cross-view knowledge into the features by constructing a 3D cost volume.
    However, the resolution of the 3D cost volume is limited by the available memory size, which limits the performance of MVS-NeRF~\cite{lin2023vision}.
    On the other hand, SRF~\cite{chibane2021srf} processes individual frame features in a pair-wise manner, and GNT~\cite{wang2022gnt} employs a transformer to efficiently incorporate cross-view knowledge.

    NeuRay~\cite{liu2022neuray} and GeoNeRF~\cite{johari2022geonerf} further improve the performance by employing visibility priors and a transformer respectively to effectively reason about the occlusions in the scene.
    More recent work such as GARF~\cite{shi2022garf}, DINER~\cite{prinzler2023diner} and MatchNeRF~\cite{chen2023explicit} try to provide explicit knowledge about the scene geometry through depth maps and similarity of the projected features.
    This approach of conditioning the NeRF on learned features is also popular among single image NeRF models~\cite{lin2023vision}, which can be considered as an extreme case of the sparse input NeRF\@.
    However, the need for pre-training on a large dataset of scenes with multi-view images and generalization issues due to domain shift have motivated researchers to continue to be interested in regularizing scene-specific radiance fields.

    \figureModelArchitectureSimpleRF
    \section{Radiance Fields and Volume Rendering Preliminaries}\label{sec:preliminaries-radiance-fields}
    We first provide a brief recap of the radiance fields and volume rendering.
    We also describe the notation required for further sections.
    To render a pixel $\pixelq$, we shoot the corresponding ray into the scene and sample $N$ 3D points $\pointp_1, \pointp_2, \ldots, \pointp_N$, where $\pointp_1$ and $\pointp_N$ are the closest to and farthest from the camera, respectively.
    At every 3D point $\pointp_i$, the radiance field $\fieldf = \fieldf_1 \circ \fieldf_2$ is queried to obtain a view-independent volume density $\sigma_i$ and a view-dependent color $\colorc_i$ as
    \begin{align}
        \sigma_i, \featureh_i = \fieldf_1(\pointp_i), \quad \colorc_i = \fieldf_2(\featureh_i, \viewv), \label{eq:radiance-field}
    \end{align}
    where $\viewv$ is the viewing direction and $\featureh_i$ is a latent feature of $\pointp_i$.
    Volume rendering is then applied along every ray to obtain the color for each pixel as
    $\colorc = \sum_{i=1}^{N} w_i \colorc_i,$
    where the weights $w_i$ are computed as
    \begin{align}
        w_i = \exp \left( -\sum_{j=1}^{i-1} \delta_j \sigma_j \right) \cdot \left( 1 - \exp \left( -\delta_i \sigma_i \right) \right), \label{eq:volumetric-rendering-weights}
    \end{align}
    and $\delta_i$ is the distance between $\pointp_i$ and $\pointp_{i+1}$.
    The expected ray termination length is computed as $z = \sum_{i=1}^{N} w_i z_i$, where $z_i$ is the depth of $\pointp_i$.
    $z$ is typically also used as the depth of the pixel $\pixelq$~\cite{deng2022dsnerf}.
    $\fieldf_1$ and $\fieldf_2$ are modelled differently for NeRF, TensoRF and ZipNeRF, and are trained using the photometric loss $\lossColor = \Vert \colorc - \hat{\colorc} \Vert^2,$ where $\hat{\colorc}$ is the true color of $\pixelq$.

    \section{Method}\label{sec:method}
    Learning a radiance field with sparse input views leads to overfitting on the input views with severe distortions in novel views.
    Our key observation is that most of the distortions are due to the sub-optimal use of the high capabilities of the radiance field model.
    Further, we find that reducing the capability of the radiance field helps constrain the model to learn only simpler solutions, which can provide better depth supervision in certain regions of the scene.
    However, the lower capability models are not optimal either, since they cannot learn complex solutions where necessary.
    Our solution here is to use the higher capability model as the main model and employ the lower capability models as augmentations to provide guidance on where to use simpler solutions.
    The challenge is that, it is not known apriori where one needs to employ supervision from the augmented model.
    We determine the more accurate model among the main and augmented models in terms of the estimated depth and use the more reliable depth to supervise the other.
    We note that the augmented models are employed only during the learning phase and not during inference.
    Thus, there is no additional overhead during inference.
    The augmented models are similar to the main model, but we modify their parameters to reduce their capability, and train them in tandem with the main model.

    To design the augmented models, we first analyze the shortcomings of the radiance field with sparse input views.
    Specifically, we determine the components of the model that cause overfitting with fewer input views, causing distortions in novel views.
    We then design the augmented models by reducing the model capability with respect to such components.
    Thus, designing the augmented models is non-trivial, and the design may need to be different for different radiance fields based on the architecture of the radiance fields and the distortions observed.
    Nonetheless, the core idea of designing augmentations by reducing their capability to learn simpler solutions is common across all radiance fields.

    We discuss the design of augmentations for NeRF, TensoRF and ZipNeRF in \cref{subsec:simple-nerf,subsec:simple-tensorf,subsec:simple-zip-nerf} respectively.
    We describe our approach to determining the reliability of the depth estimates in \cref{subsubsec:reliable-depth-estimates}.
    Finally, \cref{subsec:overall-loss} summarizes all the loss functions used to train our full model.
    \cref{fig:model-architecture-simple-rf} shows the architecture of our family of simple radiance fields.

    \subsection{Simple-NeRF}\label{subsec:simple-nerf}
    We start by discussing the specific details of the NeRF that are relevant for the design of augmentations in \cref{subsubsec:preliminaries-nerf}, then analyze the shortcomings of the NeRF with sparse input views in \cref{subsubsec:distortions-sparse-input-nerf} and finally discuss the design of augmentations in \cref{subsubsec:simple-nerf-design-of-augmentations}.
    \cref{fig:model-architecture-simple-nerf} shows the detailed architecture of Simple-NeRF\@.

    \figureModelArchitectureSimpleNeRF

    \subsubsection{NeRF Preliminaries}\label{subsubsec:preliminaries-nerf}
    The NeRF learns the radiance field $\fieldf$ using two neural networks $\mlpn_1, \mlpn_2$ and predicts the view-independent volume density $\sigma_i$ and view-dependent color $\mathbf{c}_i$ as
    \begin{align}
        \sigma_i, \featureh_i = \mlpn_1 \left( \gamma(\pointp_i, 0, l_p) \right); \qquad \colorc_i = \mlpn_2 \left( \featureh_i, \gamma(\viewv, 0, l_v) \right),  \label{eq:nerf-mlp-n1-n2}
    \end{align}
    where $\viewv$ is the viewing direction, $\featureh_i$ is a latent feature of $\pointp_i$ and
    \begin{align}
        \gamma(x, d_1, d_2) = [\sin(2^{d_1}x), \cos(2^{d_1}x), \ldots, \sin(2^{d_2-1}x), \cos(2^{d_2-1}x)] \label{eq:positional-encoding}
    \end{align}
    is the positional encoding from frequency $d_1$ to $d_2$.
    $l_p$ and $l_v$ are the highest positional encoding frequencies for $\pointp_i$ and $\viewv$ respectively.
    When $d_1=0$, $x$ is concatenated to the positional encoding features in \cref{eq:positional-encoding}.
    NeRF circumvents the need for the dense sampling of 3D points along a ray by employing two sets of MLPs, a coarse NeRF and a fine NeRF, both trained using $\lossColor$.
    The coarse NeRF is trained with a coarse stratified sampling, and the fine NeRF with dense sampling around object surfaces, where object surfaces are coarsely localized based on the predictions of the coarse NeRF\@.
    Since the scene geometry is mainly learned by the coarse NeRF, we add the augmentations only to the coarse NeRF\@.

    \subsubsection{Analysing Sparse Input NeRF}\label{subsubsec:distortions-sparse-input-nerf}
    \figureDistortionsSparseInputNeRF

    With sparse input views, we find that two components of the NeRF, namely positional-encoding and view-dependent radiance, can cause overfitting, leading to distortions in novel views.
    Both positional encoding and view-dependent radiance are elements designed to increase the capability of the NeRF to explain different complex phenomena.
    For example, the former helps in learning thin objects against a farther background, and the latter helps in learning specular objects.
    However, when training with sparse views, the fewer constraints coupled with the higher capacity of the NeRF lead to solutions that overfit the observed images and learn implausible scene geometries.
    Specifically, the high positional encoding degree leads to undesired depth discontinuities in smooth-depth regions resulting in floater artifacts, where a part of an object is broken away from it and floats freely in space~\cite{barron2022mipnerf360}, as shown in \cref{fig:distortions-nerf-floaters}.
    The view-dependent radiance causes shape-radiance ambiguity, leading to duplication artifacts in the novel views as shown in \cref{fig:distortions-nerf-duplications}.
    With sparse input views, the NeRF explains different objects by varying the color of the same 3D points based on the viewing direction, thereby giving us an illusion of the object without learning the correct geometry of the object.
    This is, in a way, similar to the illusion created by lenticular images and can be observed better in the supplementary video.
    Our augmentations consist of reducing the capability of the NeRF model with respect to the positional encoding and view-dependent radiance to obtain better depth supervision.

    \subsubsection{Design of Augmentations}\label{subsubsec:simple-nerf-design-of-augmentations}
    We employ two augmentations, one each for regularizing positional encoding and view-dependent radiance, which we describe in the following.
    We refer to the two augmentations as smoothing and Lambertian augmentations, respectively.

    \emph{Smoothing augmentation:}
    The positional encoding maps two nearby points in $\mathbb{R}^3$ to two farther away points in $\mathbb{R}^{3(2l_p+1)}$ allowing the NeRF to learn sharp discontinuities in volume density between the two points in $\mathbb{R}^3$ as a smooth function in $\mathbb{R}^{3(2l_p+1)}$.
    We reduce the depth discontinuities, which are caused by discontinuities in the volume density, by reducing the highest positional encoding frequency for $\pointp_i$ to $l_p^s < l_p$ as
    \begin{align}
        \sigma_i, \mathbf{h}_i = \mlpn^s_1 (\gamma(\pointp_i, 0, l_p^s)), \label{eq:nerf-augmentation-smoothing-mlp-n1}
    \end{align}
    where $\mlpn^s_1$ is the MLP of the augmented model.
    The main model is more accurate where depth discontinuities are required, and the augmented model is more accurate where discontinuities are not required.
    We determine the respective locations as binary masks and use only the reliable depth estimates from one model to supervise the other model, as we explain in \cref{subsubsec:reliable-depth-estimates}.

    Since color tends to have more discontinuities than depth in regions such as textures, we include the remaining high-frequency positional encoding components of $\pointp_i$ in the input for $\mlpn_2$ as
    \begin{align}
        \mathbf{c}_i = \mlpn^s_2 (\mathbf{h}_i, \gamma(\pointp_i, l_p^s, l_p), \gamma(\mathbf{v}_i, 0, l_v)). \label{eq:nerf-augmentation-smoothing-mlp-n2}
    \end{align}
    Note that $\featureh_i$ already includes the low-frequency positional encoding components of $\pointp_i$.

    \emph{Lambertian Augmentation:}
    The ability of the NeRF to predict view-dependent radiance helps it learn non-Lambertian surfaces.
    With fewer images, the NeRF can simply learn any random geometry and change the color of 3D points in accordance with the input viewpoint to explain away the observed images~\cite{zhang2020nerfpp}.
    To guard the NeRF against this, we disable the view-dependent radiance in the second augmented NeRF model to output color based on $\pointp_i$ alone as
    \begin{align}
        \sigma_i, \featureh_i = \mlpn^l_1 (\gamma(\pointp_i, 0, l_p)); \qquad \colorc_i = \mlpn^l_2 (\featureh_i), \label{eq:nerf-augmentation-lambertian-mlp-n1-n2}
    \end{align}
    We note that while the augmented model is more accurate in Lambertian regions, the main model is better equipped to handle specular objects.
    We determine the respective locations as we explain in the following and use only the reliable depth estimates for supervision.

    \subsubsection{Determining Reliable Depth Estimates}\label{subsubsec:reliable-depth-estimates}
    \figureReliableDepthEstimates

    Let the depths estimated by the main and augmented models for pixel $\pixelq$ be $z_m$ and $z_a$ respectively.
    We now seek to determine the more accurate depth among the two.
    \cref{fig:reliable-depth-estimation} shows our approach to determining the reliability of the estimated depth.
    Specifically, we reproject a $k \times k$ patch around $\pixelq$ to the nearest training view using both $z_m$ and $z_a$.
    We then compute the similarity of the reprojected patch with the corresponding patch in the first image using the mean squared error (MSE) in intensities.
    We choose the depth corresponding to lower MSE as the reliable depth.
    To filter out the cases where both the main and augmented models predict incorrect depth, we define a threshold $e_\tau$ and mark the depth to be reliable only if its MSE is also less than $e_\tau$.
    If $e_m$ and $e_a$ are the reprojection MSE corresponding to $z_m$ and $z_a$ respectively, we compute a mask $m_a$ that indicates where the augmented model is more reliable as
    \begin{align}
        m_a =
        \begin{cases}
            1 \qquad\ &\text{ if } (e_a \le e_m) \ \text{and} \ (e_a \le e_\tau)\\
            0 &\text{ otherwise }.
        \end{cases}
        \label{eq:mask-stable-sample}
    \end{align}
    We compute the reliability mask $m_m$ for the main model similarly.
    We now impose the depth supervision as
    \begin{align}
        \lossAugDepth = \indicatorfunc_{\{m_a = 1\}}  \odot \Vert z_m - \stopgradient(z_a) \Vert^2 + \indicatorfunc_{\{m_m = 1\}} \odot \Vert \stopgradient(z_m) - z_a \Vert^2,
        \label{eq:loss-augmentation}
    \end{align}
    where $\odot$ denotes element-wise product, $\indicatorfunc$ is the indicator function and $\stopgradient$ is the stop-gradient operator.
    We impose two losses, $\lossAugSmoothing$ for the smoothing augmentation and $\lossAugLambertian$ for the Lambertian augmentation.
    The final depth supervision loss is the sum of the two losses.

    For specular regions, the intensities of the reprojected patches may not match, leading to the masks being zero.
    This only implies supervision for fewer pixels and not supervision with incorrect depth estimates.

    \subsubsection{Hierarchical Sampling}\label{subsubsec:hierarchical-sampling}
    \figureSimpleNerfHierarchicalSampling
    Since multiple solutions can explain the observed images in the few-shot setting, the coarse and fine MLP of the NeRF may converge to different depth estimates for a given pixel as shown in \cref{fig:qualitative-simple-nerf-llff06b}.
    Thus, dense sampling may not be employed around the region where the fine NeRF predicts the object surface, which is equivalent to using only the coarse sampling for the fine NeRF\@.
    This can lead to blur in rendered images as seen in \cref{fig:qualitative-simple-nerf-llff06a}.
    To prevent such inconsistencies, we drive the two NeRFs to be consistent in their solutions by imposing an MSE loss between the depths predicted by the two NeRFs.
    If $z_c$ and $z_f$ are the depths estimated by the coarse and fine NeRFs respectively, we define the coarse-fine consistency loss as
    \begin{align}
        \lossCoarseFineConsistency = \indicatorfunc_{\{m_f = 1\}}  \odot \Vert z_c - \stopgradient(z_f) \Vert^2 + \indicatorfunc_{\{m_c = 1\}} \odot \Vert \stopgradient(z_c) - z_f \Vert^2,
        \label{eq:loss-coarse-fine-consistency}
    \end{align}
    where the masks $m_c$ and $m_f$ are determined as we describe in \cref{subsubsec:reliable-depth-estimates}.

    Apart from enforcing consistency between the coarse and fine NeRF models, $\lossCoarseFineConsistency$ provides two additional benefits.
    Without $\lossCoarseFineConsistency$, the augmentations need to be imposed on the fine NeRF as well, leading to an increase in the training time and memory requirements.
    Secondly, if one of the coarse or fine NeRFs converges to the correct solution, $\lossCoarseFineConsistency$ helps quickly convey the knowledge to the other NeRF, thereby facilitating faster convergence.

    \subsection{Simple-TensoRF}\label{subsec:simple-tensorf}
    We first provide a brief overview of TensoRF~\cite{chen2022tensorf} in \cref{subsubsec:preliminaries-tensorf} and also describe the notation required to explain the design of our augmentations.
    We discuss the distortions observed with sparse input TensoRF in \cref{subsubsec:distortions-sparse-input-tensorf} and then discuss the design of augmentations in \cref{subsubsec:simple-tensorf-design-of-augmentations}.

    \subsubsection{TensoRF Preliminaries}\label{subsubsec:preliminaries-tensorf}
    TensoRF models the fields $\fieldf_1$ and $\fieldf_2$ with a tensor $\gridg_1$ and a tiny MLP $\mlpn_2$, respectively.
    The 3D tensor $\gridg_1$ is factorized as the sum of outer products of 1D vectors $\vectorv$ and 2D matrices $\matrixm$.
    Specifically, $\gridg_1$ consists of two 3D tensors, $\gridg_\sigma$ to learn the volume density and $\gridg_c$ to learn the latent features of the color as
    \begin{align}
        \gridg_{\sigma} =& \sum_{r=1}^{R_\sigma} \vectorv_{\sigma, r}^{X} \circ \matrixm_{\sigma, r}^{YZ} +
                          \sum_{r=1}^{R_\sigma} \vectorv_{\sigma, r}^{Y} \circ \matrixm_{\sigma, r}^{XZ} +
                          \sum_{r=1}^{R_\sigma} \vectorv_{\sigma, r}^{Z} \circ \matrixm_{\sigma, r}^{XY}, \label{eq:tensorf-grid-sigma} \\
        \gridg_{c} =& \sum_{r=1}^{R_c} \vectorv_{c, r}^{X} \circ \matrixm_{c, r}^{YZ} \circ \vectora_{3r-2} +
                      \sum_{r=1}^{R_c} \vectorv_{c, r}^{Y} \circ \matrixm_{c, r}^{XZ} \circ \vectora_{3r-1}   \\ \nonumber
                    & + \sum_{r=1}^{R_c} \vectorv_{c, r}^{Z} \circ \matrixm_{c, r}^{XY} \circ \vectora_{3r}, \label{eq:tensorf-grid-color} \\
        \sigma_i &= \text{sigmoid}(\gridg_{\sigma}(\pointp_i)); \qquad \qquad \featureh_i = \gridg_{c}(\pointp_i),
    \end{align}
    where $\circ$ represents the outer product, and $R_\sigma$ and $R_c$ represent the number of components in the factorization of sigma and color grids, respectively.
    $\gridg (\pointp_i)$ is obtained by trilinearly interpolating $\gridg$ at $\pointp_i$.
    $\vectorv^X \in \realR^{I}$ and $\matrixm^{YZ} \in \realR^{J \times K}$ represent the vector along the x-axis and the matrix in the yz-plane respectively and so on, where $I, J $ and $K$ represent the resolution of the tensor in the $x, y$ and $z$ dimensions respectively.
    Thus, the total number of voxels in the tensor is $\numTensorfVoxels = I \times J \times K$.
    Note that $\gridg_c$ uses an additional vector $\vectora_r \in \realR^{D}$ to learn appearance as a latent feature of dimension $D$.

    TensoRF assumes that the entire scene is contained within a 3D bounding box $\vectorb$ as shown in \cref{fig:distortions-tensorf-replications}, whose vertices are given by $\{ (b_{x_1}, b_{x_2}), (b_{y_1}, b_{y_2}), (b_{z_1}, b_{z_2}) \}$.
    TensoRF handles unbounded forward-facing scenes by transforming the space into normalized device coordinates (ndc) similar to the NeRF\@.
    The coarse to fine training is implemented by using lower resolution tensors $\gridg_\sigma$ and $\gridg_c$ during the initial stages of the optimization and gradually increasing the resolution as the training progresses.
    Finally, the color at $\pointp_i$ is obtained using the tiny MLP $\mlpn_2$ as
    \begin{align}
        \colorc_i = \mlpn_2(\featureh_i, \gamma(\viewv, 0, l_v)), \label{eq:tensorf-mlp-color}
    \end{align}
    where $\gamma$ is the positional encoding described by \cref{eq:positional-encoding}.
    Thus, we note that $\fieldf_2 = \gamma \circ \mlpn_2$.
    The color of the pixel is then obtained through volume rendering using $\sigma_i$ and $\colorc_i$ as in \cref{sec:preliminaries-radiance-fields}.
    For further details, we refer the readers to TensoRF~\cite{chen2022tensorf}.

    \subsubsection{Analysing Sparse Input TensoRF}\label{subsubsec:distortions-sparse-input-tensorf}
    \figureDistortionsSparseInputTensoRF
    When training a TensoRF model with sparse input views, we find that three of its components cause overfitting, leading to distortions in novel views.
    Employing a higher resolution tensor $\gridg_\sigma$ with a large number of components $R_\sigma$ allows the TensoRF to learn sharp depth edges, but results in undesired depth discontinuities in smooth regions causing floaters as shown in \cref{fig:distortions-tensorf-floaters}.
    Further, the large bounding box $\vectorb$ allows the TensoRF to handle objects that are truly very close to the camera.
    On account of large distances between cameras when only a few input views are available, it may be possible to place objects close to one camera such that they are out of the field of view of the other cameras, even for objects visible in multiple input views.
    Specifically, TensoRF learns multiple copies of the same object, each visible in only one input view, thereby explaining the observations without learning the geometry of the objects as shown in \cref{fig:distortions-tensorf-replications}.
    We design the augmentations to reduce the capability of the TensoRF model with respect to these three components.

    \subsubsection{Design of Augmentations}\label{subsubsec:simple-tensorf-design-of-augmentations}
    Employing a high-resolution and high-rank tensor $\gridg_\sigma$ enables TensoRF to learn significantly different $\sigma$ values for two nearby points in $\mathbb{R}^3$ leading to undesired depth discontinuities in smooth regions.
    We constrain the augmented TensoRF to learn only smooth and continuous depth surfaces by reducing the number of components to $R_\sigma^s < R_\sigma$ and also reducing the number of voxels of $\gridg_\sigma$ from to $\numTensorfVoxels[s] < \numTensorfVoxels$.
    We note that modifying only one of these components is insufficient to achieve the desired smoothing.
    For example, only reducing the resolution of the grid allows TensoRF to learn sharp changes in $\sigma$ at the voxel edges, leading to block artifacts.
    On the other hand, only reducing the number of components allows TensoRF to learn sharp changes in $\sigma$ on account of the high-resolution grid.

    Further, we find that reducing $R_\sigma$ to $R_\sigma^s$ leads to the augmented TensoRF learning cloudy volumes instead of hard object surfaces.
    We encourage the augmented TensoRF to learn hard surfaces by employing a mass concentration loss that minimizes the entropy of mass, grouped into $\numTensorfAugMassConcentrationBins$ intervals as
    \begin{align}
        \lossMassConcentration = H \commonbrackets{\flowerbrackets{\sum_{i = (j-1)(N / \numTensorfAugMassConcentrationBins)+1}^{j(N / \numTensorfAugMassConcentrationBins)} w_i}_{j=1}^{\numTensorfAugMassConcentrationBins}}, \label{eq:loss-tensorf-mass-concentration}
    \end{align}
    where $H(w_1, w_2, \ldots, w_n) = \sum_{i=1}^{n}(-w_i \log w_i)$ is the entropy operator, $N$ is the number of 3D points $\pointp_i$ along a ray and $w_i$ is the weight corresponding to $\pointp_i$ as described in \cref{eq:volumetric-rendering-weights}.
    The effect of $\lossMassConcentration$ can be observed in the supplementary videos.

    Objects that are incorrectly placed close to the camera due to a large bounding box are typically smooth in depth and hence are not mitigated by the above augmentation.
    We design a second augmentation to mitigate such distortions by reducing the size of the bounding box $\vectorb$ along the z-axis by increasing $b_{z_1}$ to $b_{z_1}^s$.
    We note that replicating the same in the main TensoRF model could lead to distortions in objects that are truly close to the camera.
    In practice, we find that including both the augmentations in a single augmented TensoRF model works reasonably well, and hence we employ a single augmented model.
    We then use the reliable depth estimates from the augmented model to supervise the main model as in \cref{eq:loss-augmentation}.

    \subsection{Simple-ZipNeRF}\label{subsec:simple-zip-nerf}
    \figureDistortionsSparseInputZipNeRF
    ZipNeRF~\cite{barron2023zipnerf} integrates the iNGP model~\cite{muller2022instant}, which achieves significant improvements in optimization and rendering times, with the anti-aliasing ability of MipNeRF~\cite{barron2021mipnerf} and the ability to handle unbounded 360$\degree$ scenes of MipNeRF360~\cite{barron2022mipnerf360}.
    Our contributions to enable the training of ZipNeRF with sparse input views are mainly with respect to the components of the iNGP model, and hence, we believe that the augmentations designed for ZipNeRF are relevant to iNGP as well.
    We discuss the specific components of ZipNeRF that are relevant in our augmentations in \cref{subsubsec:preliminaries-zipnerf}, analyze the limitations of ZipNeRF with sparse input views in \cref{subsubsec:distortions-sparse-input-zipnerf} and then discuss the design of our augmentations in \cref{subsubsec:simple-zipnerf-design-of-augmentations}.

    \subsubsection{ZipNeRF Preliminaries}\label{subsubsec:preliminaries-zipnerf}
    ZipNeRF employs a multi-resolution grid and a hash function that maps every vertex of the grid to an entry in a hash table.
    The hash table contains the latent features representing the volume density and the radiance.
    Concretely, given a point $\pointp_i \in \realR^3$, the vertices of the voxel enclosing $\pointp_i$ are mapped to an entry in a hash table of length $T$ through the use of hash function $\hashgridh_1$ as,
    \begin{align}
        \hashgridh_1(\pointp) = \commonbrackets{\bigoplus_{j=1}^{3} p_j \pi_j} \text{ mod } T,
    \end{align}
    where $\oplus$ denotes the bit-wise XOR operation, $\pi_j$ is a prime number, and $p_j$ is the $j$-th coordinate of $\pointp$.
    The feature vectors corresponding to the eight vertices of the voxel are trilinearly interpolated.
    The same procedure is repeated for every level of the multi-resolution grid, and the corresponding interpolated features are concatenated to form the latent feature $\hashgridh_1(\pointp_i)$.
    Two tiny MLPs are employed to decode $\hashgridh_1(\pointp_i)$ into the volume density and the radiance as
    \begin{align}
        \sigma_i, \featureh_i = \mlpn_1(\hashgridh_1(\pointp_i)); \qquad \colorc_i = \mlpn_2(\featureh_i, \gamma(\viewv, 0, l_v)),
    \end{align}
    where $\gamma$ is the positional encoding as defined in \cref{eq:positional-encoding}.
    We note that $\fieldf_1$ and $\fieldf_2$ in \cref{eq:radiance-field} are thus represented as $\fieldf_1 = \hashgridh_1 \circ \mlpn_1$ and $\fieldf_2 = \gamma \circ \mlpn_2$.
    The color of the pixel is then obtained through volume rendering using $\sigma_i$ and $\colorc_i$ as in \cref{sec:preliminaries-radiance-fields}.
    Note that multiple vertices in the grid could map to the same entry in the hash table at every level.
    iNGP and ZipNeRF rely on the MLP $\mlpn_1$ to resolve such collisions based on multi-resolution features.
    Unbounded scenes are handled by employing a contraction function that maps the distance along the ray from $z \in [z_{\text{near}}, z_{\text{far}}]$ to a normalized distance $s \in [0, 1]$.
    The 3D points $\pointp_i$ are then sampled in $s$-domain.
    For further details, we refer the readers to iNGP~\cite{muller2022instant} and ZipNeRF~\cite{barron2023zipnerf}.

    \subsubsection{Analysing Sparse Input ZipNeRF}\label{subsubsec:distortions-sparse-input-zipnerf}
    We find that two components of ZipNeRF tend to cause overfitting when trained with sparse input views, leading to distortions in novel views.
    Firstly, employing a hash table with a large size $T$ enables ZipNeRF to learn sharp depth edges, but introduces undesired depth discontinuities in smooth regions, causing floaters as shown in \cref{fig:distortions-zipnerf-floaters}.
    Secondly, since ZipNeRF handles unbounded 360$\degree$ scenes, it learns the radiance field over the entire 3D space.
    Similar to TensoRF, ZipNeRF tends to incorrectly place multiple copies of objects very close to the camera without learning the correct geometry as shown in \cref{fig:distortions-zipnerf-replications}.
    Thus, we design the augmentations to reduce the capability of the ZipNeRF model with respect to these two components.

    \subsubsection{Design of Augmentations}\label{subsubsec:simple-zipnerf-design-of-augmentations}
    Employing a hash table of larger size $T$ allows ZipNeRF to avoid collisions and not share features across multiple 3D points.
    This enables ZipNeRF to map two nearby points in $\realR^3$ to two independent entries in the hash table, thus mapping them to two farther away points in the latent feature space.
    This allows the MLP $\mlpn_1$ to learn discontinuities in the volume density, resulting in sharp depth edges.
    We encourage the augmented model to share features across more 3D points by reducing the size of the hash table to $T^s < T$.

    To mitigate the objects being placed close to the camera incorrectly, we cannot reduce the size of the bounding box as in TensoRF, since ZipNeRF handles unbounded scenes.
    We achieve a similar effect by sampling the 3D points $\pointp_i$ along the ray in $s$-domain in the range $s \in [s_{\text{near}}, 1]$ instead of $s \in [0, 1]$.
    This ensures that the objects are placed at least at a certain distance away from the camera in the augmented model.
    However, we note that employing the above modification in the main model is detrimental to learning or rendering any objects that are truly close to the camera.
    In practice, we find that including both the augmentations in a single augmented ZipNeRF model works reasonably well, and hence, we employ a single augmented model.
    We then use these depth estimates as in \cref{eq:loss-augmentation}.

    \subsection{Overall Loss}\label{subsec:overall-loss}
    Let $\lossMainModel$ denote the combination of the losses employed by the corresponding radiance fields.
    For example, while the NeRF employs only the photometric loss $\lossColor$, TensoRF employs a total variation regularization in addition to $\lossColor$.
    We refer the readers to the corresponding papers for the details of all the losses imposed.
    We impose all such losses on the augmented models as well and denote them by $\lossAugModel$.
    In addition, we also include the sparse depth loss on both the main and augmented models as,
    \begin{align}
        \lossSparseDepth = \Vert z_m - \hat{z} \Vert^2 + \Vert z_{a} - \hat{z} \Vert^2, \label{eq:loss-sparse-depth}
    \end{align}
    where $z_m$ and $z_a$ are the depths obtained from the main and augmented models respectively, and $\hat{z}$ is the sparse depth given by the SfM model~\cite{deng2022dsnerf}.
    Our final loss is a combination of all the losses as,
    \begin{align}
        \mathcal{L} = \lossWeightMainModel \lossMainModel + \lossWeightAugModel \lossAugModel + \lossWeightSparseDepth \lossSparseDepth + \lossWeightAugDepth \lossAugDepth + \\ \nonumber
        \lossWeightCoarseFineConsistency \lossCoarseFineConsistency + \lossWeightMassConcentration \lossMassConcentration,
    \end{align}
    where $\lossCoarseFineConsistency$ and $\lossMassConcentration$ are respectively imposed for the main NeRF and augmented TensoRF models only, and $\lossWeightMainModel, \lossWeightAugModel, \lossWeightSparseDepth, \lossWeightAugDepth, \lossWeightCoarseFineConsistency$ and $\lossWeightMassConcentration$ are hyper-parameters.

    \subsection{Discussion}\label{subsec:discussion}
    We note that, although we design different augmented models for Simple-NeRF, Simple-TensoRF and Simple-ZipNeRF, the core idea behind the design of the augmentations is the same.
    In particular, all three radiance field models learn undesired depth discontinuities leading to floaters.
    We mitigate this by designing augmentations that learn smoother geometry through lower positional encoding degree in the NeRF, smaller grid in TensoRF and smaller hash-table in ZipNeRF.
    While the specifics of the smoothing augmentations differ, the mechanism of regularizing the main model using the learnt depth prior from the augmentations remains the same across the three Simple-RF models.

    Further, our augmentations can be extended to regularize other radiance field models when training with limited views.
    For example, most of the NeRF family of models such as MipNeRF360~\cite{barron2022mipnerf360} and Ref-NeRF~\cite{verbin2022refnerf} suffer from floater artifacts when trained with sparse input views.
    Our smoothing augmentation can be employed to mitigate the floaters.
    Further, Ref-NeRF~\cite{verbin2022refnerf} extends the NeRF to handle scenes with shiny objects.
    When training with limited views, Ref-NeRF suffers from duplication artifacts similar to \cref{fig:distortions-nerf-duplications}.
    Such artifacts can be mitigated using our Lambertian augmentation discussed in \cref{subsubsec:simple-nerf-design-of-augmentations}.
    Similarly, K-Planes~\cite{fridovich2023kplanes} extends TensoRF to handle varying illumination in the input images, by learning a global appearance code per input image, similar to NeRF-W~\cite{martin2021nerfw}.
    With sparse input views, the global appearance code might end up learning geometric information as well, leading to a lack of geometric details in novel views.
    An augmented model with a lower dimensional appearance code can help mitigate such artifacts.

    \section{Experimental Setup}\label{sec:experimental-setup}

    \subsection{Datasets}\label{subsec:datasets}
    \tableRealEstateTrainTest
    We evaluate the performance of our models on four popular datasets, namely NeRF-LLFF~\cite{mildenhall2019llff}, RealEstate-10K~\cite{zhou2018stereomag}, MipNeRF360~\cite{barron2022mipnerf360} and NeRF-Synthetic \cite{mildenhall2020nerf}.
    We assume the camera parameters are known for the input images, since in applications such as robotics or extended reality, external sensors or a pre-calibrated set of cameras may provide the camera poses.

    \emph{NeRF-LLFF} dataset contains eight real-world forward-facing scenes typically consisting of an object at the centre against a complex background.
    Each scene contains a varying number of images ranging from 20 to 60, each with a spatial resolution of $1008 \times 756$.
    Following prior work~\cite{niemeyer2022regnerf}, we use every 8th view as the test view and uniformly sample 2, 3 or 4 input views from the remaining.

    \emph{RealEstate-10K} dataset contains a large number of real-world forward-facing scenes, from which we select 5 test scenes for our experiments.
    We include both indoor and unbounded outdoor scenes and select 50 temporally continuous frames from each scene.
    The frames have a spatial resolution of $1024 \times 576$.
    Following prior work~\cite{somraj2023vipnerf}, we reserve every 10th frame for training and choose 2, 3 or 4 input views among them.
    In the remaining 45 frames, we use those frames that are not very far from the input frames for testing.
    Specifically, we choose all the frames between the training views that correspond to interpolation and five frames on either side that correspond to extrapolation.
    \cref{tab:realestate-train-test} shows the train and test frame numbers we use for the three different settings.

    \emph{MipNeRF360} dataset contains seven publicly available unbounded 360$\degree$ real-world scenes including both indoor and outdoor scenes.
    Each scene contains 100 to 300 images.
    The four indoor scenes have a spatial resolution of approximately $1560 \times 1040$, and the three outdoor scenes have an approximate spatial resolution of $1250 \times 830$.
    Following prior work~\cite{barron2023zipnerf}, we reserve every 8th view for testing and uniformly sample 12, 20 and 36 input views from the remaining.
    We use more input views on this dataset as compared to the other datasets owing to the significantly larger fields of view.

    \emph{NeRF-Synthetic} dataset contains eight bounded 360$\degree$ synthetic scenes, each containing 100 train images and 200 test images.
    All the scenes have a spatial resolution of $800 \times 800$.
    For training, we uniformly sample 4, 8 and 12 input views from the training set and test on all 200 test images.

    \tableQuantitativeSimpleNerfLlff
    \tableQuantitativeSimpleNerfRealEstate
    \tableQuantitativeSimpleNerfDepth
    
    \subsection{Evaluation measures}\label{subsec:evaluation-measures}
    We quantitatively evaluate the predicted frames from various models using the peak signal-to-noise ratio (PSNR), structural similarity (SSIM)~\cite{wang2004image} and LPIPS~\cite{zhang2018unreasonable} measures.
    For LPIPS, we use the v0.1 release with the AlexNet~\cite{krizhevsky2012alexnet} backbone, as suggested by the authors.
    We also employ depth mean absolute error (MAE) to evaluate the models on their ability to predict absolute depth in novel views.
    In addition, we also evaluate the models with regard to their ability to predict better relative depth using the spearman rank order correlation coefficient (SROCC).
    Obtaining better relative depth might be more crucial in downstream applications such as 3D scene editing.
    Since the ground truth depth is not provided in the datasets, we train NeRF and ZipNeRF models with dense input views on forward-facing and 360$\degree$ datasets respectively and use their depth predictions as pseudo ground truth.
    On the NeRF-LLFF and MipNeRF360 datasets, we normalize the predicted depths by the median ground truth depth, since the scenes have different depth ranges.
    With very few input views on forward-facing datasets, the test views could contain regions that are not visible in the input views, and hence, we also evaluate both the view synthesis and depth performance in visible regions only.
    To determine such regions, we use the depth estimated by a NeRF trained with dense input views and compute the visible region mask through reprojection error in depth.
    We provide more details on the mask computation in the supplementary.
    On the other hand, the input views cover most of the scene in the 360$\degree$ datasets, and hence we evaluate the performance on full frames.
    We do not evaluate the rendered depth on the NeRF-Synthetic dataset since the depth estimated with the dense input ZipNeRF is unreliable, especially in the white background regions, and the ground truth depth is not provided in the dataset either.
    We also do not evaluate the rendered depth of 3DGS based models since we find that their depth is inconsistent with NeRF based models.

    \section{Experimental Results}\label{sec:experimental-results}
    We present the main results of our work with Simple-NeRF in \cref{subsec:results-simple-nerf} and then show the extension of our ideas to explicit models in \cref{subsec:results-simple-tensorf,subsec:results-simple-zipnerf}.

    \subsection{Simple-NeRF}\label{subsec:results-simple-nerf}
    \figureQualitativeSimpleNerfRealEstateFirst
    \figureQualitativeSimpleNerfRealEstateSecond
    \figureQualitativeSimpleNerfRealEstateThird
    \figureQualitativeSimpleNerfLlffFirst
    \figureQualitativeSimpleNerfLlffSecond
    \figureQualitativeSimpleNerfLlffThird
    \figureQualitativeSimpleNerfLlffFourth
    \figureQualitativeSimpleNerfDepthFirst

    \subsubsection{Comparisons}\label{subsubsec:comparisons-simple-nerf}
    We evaluate the performance of our Simple-NeRF on the two forward-facing datasets only since the NeRF does not natively support unbounded 360 scenes.
    We evaluate the performance of our model against various sparse input NeRF models.
    We compare with DS-NeRF~\cite{deng2022dsnerf}, DDP-NeRF~\cite{roessle2022ddpnerf} and RegNeRF~\cite{niemeyer2022regnerf} which regularize the depth estimated by the NeRF\@.
    We also evaluate DietNeRF~\cite{jain2021dietnerf} and InfoNeRF~\cite{kim2022infonerf} that regularize the NeRF in hallucinated viewpoints.
    We also include two recent models, FreeNeRF~\cite{yang2023freenerf} and ViP-NeRF~\cite{somraj2023vipnerf} and two 3DGS based models, InstantSplat~\cite{fan2024instantsplat} and CoR-GS~\cite{zhang2024corgs}, among the comparisons.
    We find that InstantSplat performs better with known camera extrinsics and hence report the results with known camera extrinsics.
    We train the models on both datasets using the codes provided by the respective authors.

    \subsubsection{Implementation details}\label{subsubsec:implementation-details-simple-nerf}
    We develop our code in PyTorch and on top of DS-NeRF~\cite{deng2022dsnerf}.
    We employ the Adam Optimizer with an initial learning rate of 5e-4 and exponentially decay it to 5e-6.
    We adjust the weights for the different losses such that their magnitudes after scaling are of similar orders.
    For the first 10k iterations of the training, we only impose $\lossMainModel, \lossAugModel$ and $\lossSparseDepth$.
    $\lossAugDepth$ and $\lossCoarseFineConsistency$ are imposed after 10k iterations.
    We set the hyper-parameters as follows:
    $l_p = 10$, $l_v = 4$, $l_p^s = 3$, $k=5$, $e_{\tau} = 0.1$, $\lossWeightMainModel = \lossWeightAugModel = 1$, $\lossWeightSparseDepth = \lossWeightAugDepth = \lossWeightCoarseFineConsistency = 0.1$ and $\lossWeightMassConcentration = 0$.
    The network architecture is exactly the same as DS-NeRF\@.
    For the augmented models, we only change the input dimension of the MLPs $\mlpn_1$ and $\mlpn_2$ appropriately.
    The augmented models are employed only during training, and the network is exactly the same as Vanilla NeRF for inference.
    We train the models on a single NVIDIA RTX 2080 Ti GPU for 100k iterations.

    \subsubsection{Quantitative and Qualitative Results}\label{subsubsec:quantitative-and-qualitative-simple-nerf}
    \cref{tab:quantitative-simple-nerf-llff,tab:quantitative-simple-nerf-realestate} show the view-synthesis performance of Simple-NeRF and other prior art on NeRF-LLFF and RealEstate-10K datasets respectively.
    We find that Simple-NeRF achieves state-of-the-art performance on both datasets in most cases.
    The higher performance of all the models on the RealEstate-10K dataset is perhaps due to the scenes being simpler.
    Hence, the performance improvement is also smaller as compared to the NeRF-LLFF dataset.
    \cref{fig:qualitative-simple-nerf-realestate01} shows predictions of various models on an example scene from the RealEstate-10K dataset, where we observe that Simple-NeRF is the best in reconstructing the novel view.
    In particular, we see that DDP-NeRF~\cite{roessle2022ddpnerf}, which uses depth estimated by deep neural network to constrain the NeRF, provides incorrect depth supervision leading to blurred floaters.
    On the other hand, RegNeRF~\cite{niemeyer2022regnerf}, which uses depth smoothness prior, fuses objects at different depths.
    In contrast to both these models, our approach of learning in-situ depth priors enables Simple-NeRF to mitigate both the issues.
    \cref{fig:qualitative-simple-nerf-realestate02,fig:qualitative-simple-nerf-realestate03,fig:qualitative-simple-nerf-llff01,fig:qualitative-simple-nerf-llff02,fig:qualitative-simple-nerf-llff03,fig:qualitative-simple-nerf-llff04} show more comparisons on both datasets with 2, 3, and 4 input views.
    Further, Simple-NeRF improves significantly in estimating the depth of the scene as seen in ~\cref{tab:quantitative-simple-nerf-depth} and ~\cref{fig:qualitative-simple-nerf-depth}.
    We provide video comparisons in the supplementary.

    We note that the quantitative results in \cref{tab:quantitative-simple-nerf-llff,tab:quantitative-simple-nerf-realestate} differ from the values reported in ViP-NeRF~\cite{somraj2023vipnerf} on account of the following two differences.
    Firstly, the quality evaluation metrics are computed on full frames in ViP-NeRF.
    However, we exclude the regions not seen in the input views as explained in \cref{subsec:evaluation-measures}.
    Secondly, while we use the same train set as that of ViP-NeRF on the RealEstate-10K dataset, we modify the test set as shown in \cref{tab:realestate-train-test}.
    We change the test set since the test views that are very far away from the train views may contain large unobserved regions.
    We provide more reasoning and details in the supplementary.

    \subsubsection{Ablations}\label{subsubsec:ablations-simple-nerf}
    \tableQuantitativeSimpleNerfAblations
    \figureQualitativeSimpleNerfAblationsFirst
    \tableQuantitativeSimpleTensorf

    We test the importance of each of the components of our model by disabling them one at a time.
    We disable the smoothing and Lambertian augmentations and coarse-fine consistency loss individually.
    When disabling $\lossCoarseFineConsistency$, we additionally add augmentations to the fine NeRF since the knowledge learned by coarse NeRF may not efficiently propagate to the fine NeRF\@.
    We also analyze the need to supervise with only the reliable depth estimates by disabling the mask and stop-gradients in $\lossAugDepth$ and $\lossCoarseFineConsistency$.
    In addition, we also analyze the effect of including residual positional encodings $\gamma(\pointp_i, l_p^s, l_p)$ while predicting the color in the smoothing augmentation model.
    \cref{tab:quantitative-simple-nerf-ablations} shows a quantitative comparison between the ablated models.
    We observe that each of the components is crucial, and disabling any of them leads to a drop in performance.
    Further, using all the depths for supervision instead of only the reliable depths leads to a significant drop in performance.
    Finally, disabling $\lossCoarseFineConsistency$ also leads to a drop in performance in addition to increasing the training time by almost $2\times$ due to the inclusion of augmentations for the fine NeRF\@.

    Since we design our regularizations on top of DS-NeRF~\cite{deng2022dsnerf} baseline, our framework can be seen as a semi-supervised learning model by considering the sparse depth from a Structure from Motion (SfM) module as providing limited depth labels and the remaining pixels as the unlabeled data.
    Our approach of using augmented models in tandem with the main radiance field model is perhaps closest to the Dual-Student architecture~\cite{ke2019dualstudent} that trains another identical model in tandem with the main model and imposes consistency regularization between the predictions of the two models.
    However, our augmented models have complementary abilities as compared to the main radiance field model.
    We now analyze if there is a need to design augmentations that learn ``simpler'' solutions by replacing our novel augmentations with identical replicas of the NeRF as augmentations.
    The seventh row of \cref{tab:quantitative-simple-nerf-ablations} shows a performance drop when using identical augmentations.

    Finally, we analyze the need for an augmentation that explicitly achieves depth smoothing.
    In other words, we ask if naively reducing the model capacity in the augmented model achieves a similar effect to that of our smoothing augmentation.
    We test this by replacing the smoothing augmentation with an augmented model that has a smaller MLP $\mlpn_1$.
    Specifically, we reduce the number of layers from eight to four in the augmented model.
    From the results in the last row of \cref{tab:quantitative-simple-nerf-ablations}, we conclude that reducing the positional encoding degree is more effective, perhaps because the MLP with fewer layers may still be capable of learning floaters on account of using all the positional encoding frequencies.

    \subsection{Simple-TensoRF}\label{subsec:results-simple-tensorf}
    \figureQualitativeSimpleTensorfFirst
    \figureQualitativeSimpleTensorfDepthFirst
    \figureQualitativeSimpleTensorfAblationsFirst

    \subsubsection{Implementation Details}\label{subsubsec:implementation-details-simple-tensorf}
    Building on the original TensoRF code base, we employ Adam Optimizer with an initial learning rate of $2e-2$ and $1e-3$ for the tensor and MLP parameters respectively, which decay to $2e-3$ and $1e-4$.
    We employ the same hyper-parameters as the original implementation for the main model as follows:
    $R_\sigma = 24$, $R_c = 72$, $\vectorb = \flowerbrackets{(-1.5, 1.5), (-1.67, 1.67), (-1.0, 1.0)}$, $\numTensorfVoxels = 640^3$, $D=27$, and $l_v = 0$.
    We set $R_\sigma^s = 12$, $b_{z_1}^s = -0.5$ and $\numTensorfVoxels[s] = 160^3$, $\numTensorfAugMassConcentrationBins=5$, $k=5$, $e_{\tau} = 0.1$ for the augmented model and the remaining hyper-parameters are the same as the main model.
    We weigh the losses as $\lossWeightMainModel = \lossWeightAugModel = 1, \lossWeightSparseDepth = \lossWeightAugDepth = 0.1, \lossWeightMassConcentration = 0.01 \text{ and } \lossWeightCoarseFineConsistency = 0$.
    We train the models on a single NVIDIA RTX 2080 Ti 11GB GPU for 25k iterations and enable $\lossAugDepth$ after 5k iterations.

    \subsubsection{Quantitative and Qualitative Results}\label{subsubsec:quantitative-and-qualitative-simple-tensorf}
    \cref{tab:quantitative-simple-tensorf} shows the view-synthesis performance of Simple-TensoRF on the NeRF-LLFF and RealEstate-10K datasets.
    We compare the performance of our model against the vanilla TensoRF and a baseline we create by adding sparse depth loss on TensoRF, which we refer to as DS-TensoRF\@.
    We find that Simple-TensoRF significantly improves performance over TensoRF and DS-TensoRF on both datasets.
    \cref{fig:qualitative-simple-tensorf} compares the three models visually, where we observe that Simple-TensoRF mitigates multiple distortions observed in the renders of TensoRF and DS-TensoRF\@.
    From \cref{tab:quantitative-simple-tensorf} and \cref{fig:qualitative-simple-tensorf-depth}, we observe that Simple-TensoRF is significantly better at estimating the scene depth than both TensoRF and DS-TensoRF\@.
    While we observe that Simple-NeRF performs marginally better than Simple-TensoRF in most cases, Simple-TensoRF achieves a lower depth MAE on the RealEstate-10K dataset.


    We test the need for the different components of our augmentation by disabling them one at a time and show the quantitative results in the second half of \cref{tab:quantitative-simple-tensorf}.
    Specifically, we disable the reduction in the number of tensor decomposition components and the number of voxels in the first two rows respectively.
    In the third row, we disable both the components, where the augmented model consists of the reduction in the bounding box size only.
    We find that disabling either or both of the components leads to a drop in performance.
    In particular, \cref{fig:qualitative-simple-tensorf-ablations01} shows that reducing only the tensor resolution and not reducing the number of tensor decomposition components leads to translucent blocky floaters.
    On the other hand, reducing only the number of components causes small and completely opaque floaters.
    These effects can be better observed in the supplementary videos.
    Further, we find that reducing the number of components $R_\sigma$ is more crucial in obtaining simpler solutions on the RealEstate-10K dataset.

    \subsection{Simple-ZipNeRF}\label{subsec:results-simple-zipnerf}
    \tableQuantitativeSimpleZipnerfMipNerf
    \tableQuantitativeSimpleZipnerfNerfSynthetic
    \figureQualitativeSimpleZipnerfMipNerfFirstSecond
    \figureQualitativeSimpleZipnerfMipNerfThird
    \figureQualitativeSimpleZipnerfDepthFirst
    \figureQualitativeSimpleZipnerfNerfSyntheticFirstSecondThird
    \figureQualitativeSimpleZipnerfAblationsFirst
    \figureQuantitativeSimpleZipnerfAblationsSecond
    \figureQualitativeSimpleZipnerfLimitationsFirst

    \subsubsection{Implementation Details}\label{subsubsec:implementation-details-simple-zipnerf}
    We build our code in PyTorch on top of an unofficial ZipNeRF implementation\footnote{ZipNeRF implementation: https://github.com/SuLvXiangXin/zipnerf-pytorch}.
    For the main model, we retain the hyper-parameters of the original ZipNeRF\@.
    For the augmented model, we reduce the size of the hash table from $T=2^{21}$ to $T^s=2^{11}$ and set $s_{\text{near}} = 0.3$.
    We impose $\lossAugDepth$ after 5k iterations and use $k=5$, $e_{\tau} = 0.2$.
    The rest of the hyper-parameters for the augmented model are the same as the main model.
    We weigh the losses as $\lossWeightMainModel = \lossWeightAugModel = 1$, $\lossWeightAugDepth = 10$ and $\lossWeightSparseDepth = \lossWeightCoarseFineConsistency = \lossWeightMassConcentration = 0$.
    We do not impose the sparse depth loss $\lossSparseDepth$ since we find that Colmap either fails in sparse reconstruction or provides noisy sparse depth for 360$\degree$ scenes.
    We train the models on a single NVIDIA RTX 2080 Ti 11GB GPU for 25k iterations.

    \subsubsection{Quantitative and Qualitative Results}\label{subsubsec:quantitative-and-qualitative-simple-zipnerf}
    We compare the performance of ZipNeRF with and without our augmentations on the MipNeRF360 and NeRF-Synthetic datasets in \cref{tab:quantitative-simple-zipnerf-mipnerf360,tab:quantitative-simple-zipnerf-nerf-synthetic} respectively.
    We observe that including our augmentations improves performance significantly on both datasets in terms of all the evaluation measures.
    This observation is further supported by the qualitative examples in \cref{fig:qualitative-simple-zipnerf-mipnerf360-0102,fig:qualitative-simple-zipnerf-mipnerf360-03,fig:qualitative-simple-zipnerf-nerfsynthetic-010203,fig:qualitative-simple-zipnerf-depth}, where we observe a clear improvement in the quality of the rendered novel views and depth when employing our augmentations.
    From \cref{tab:quantitative-simple-zipnerf-mipnerf360}, we observe that the performace of InstantSplat decreases with more number of input views.
    This is perhaps because of applying DUSt3R~\cite{wang2024dust3r} independently on each pair of input views, leading to accumulation of errors.
    While CoR-GS achieves very good performance, we find that it does not reconstruct high frequency details in the scene as seen in \cref{fig:qualitative-simple-zipnerf-mipnerf360-03}.
    As a result, it achieves better PSNR and SSIM scores, but poorer LPIPS scores.
    Further, we observe that CoR-GS also suffers from large floater artifacts as shown in \cref{fig:qualitative-simple-zipnerf-mipnerf360-01}.
    In addition, \cref{tab:quantitative-simple-zipnerf-mipnerf360,fig:qualitative-simple-zipnerf-ablations01} also shows the performance of the augmented model on the MipNeRF360 dataset.
    We observe a significant reduction in distortions in the renders of the augmented model; however, the same does not reflect in the quantitative evaluation due to the blur introduced by the augmented model.

    Further, in \cref{fig:quantitative-simple-zipnerf-increasing-views}, we show the performance of ZipNeRF and Simple-ZipNeRF as the number of input views increases.
    We observe that the performance of ZipNeRF is too low with very few input images, where our augmentation does not help improve the performance.
    As the number of input views increases, the performance of ZipNeRF improves, and our augmentation helps improve the performance significantly.
    Further, with a large number of input views, the performance of ZipNeRF saturates, and our augmentation does not help improve the performance.
    This shows that our augmentations are highly effective when the performance of the base model is moderately good.

    \section{Discussion}\label{sec:discussion}
    \subsection{Computational Complexity}\label{subsec:computational-complexity}
    \tableComparisonSpaceTime
    We report the approximate GPU memory utilization and time taken for training and inference of our family of Simple-RF models in \cref{tab:comparison-space-time}.
    We observe that Simple-NeRF with two augmentations takes only 1.5 times more time than NeRF for training on account of employing augmentations on the coarse NeRF only.
    While coarse NeRF queries the MLPs 64 times, the fine NeRF queries the MLPs 192 times, giving a combined 256 queries per pixel.
    Simple-NeRF queries the coarse MLPs 192 times and the fine MLPs 192 times, with a total of 384 queries per pixel.
    On the other hand, Simple-TensoRF and Simple-ZipNeRF take twice the time as TensoRF and ZipNeRF respectively on account of employing a single augmentation with exactly the same number of queries as the main model.
    We note that it could be possible to further reduce the training time for ZipNeRF by employing the augmentation only on the proposal MLP.
    However, this requires the proposal MLPs to output color and to be trained with the photometric loss instead of the interval loss~\cite{barron2023zipnerf}.
    The effect of such a change is unclear and is left for future work.
    Interestingly, Simple-TensoRF requires only a little more memory than TensoRF during training, perhaps due to the low resolution tensor employed by the augmented model.
    Further, while the NeRF models require significantly less memory during inference, grid based models such as TensoRF and ZipNeRF require more memory due to the use of a voxel grid in place of MLPs.
    Finally, we note that at inference time, Simple-RF models take exactly the same time and memory as the baseline models since the augmentations are disabled during inference.
    All the above experiments are conducted on a single NVIDIA RTX 2080 Ti 11GB GPU\@.

    \subsection{Analysis}\label{subsec:analysis}
    \figureAnalysisSceneComplexityFirst
    \figureAnalysisPerformanceImprovementFirst

    We now analyse the correlation between the performance of our model against the scene complexity and the performance of the base model.
    We measure the spatial complexity of a scene using the average image gradient magnitude across all the training images, normalized by the number of input views.
    The intuition is that our model is good at reducing the floaters in smooth regions, but struggles in highly textured regions as they are more difficult to reconstruct accurately with sparse views.
    However, as the number of input views increases, our model is able to reconstruct more complex scenes.
    In~\autoref{fig:analysis-scene-complexity}, we show a scatter plot between the spatial complexity and the SSIM scores of Simple-NeRF across multiple scenes in NeRF-LLFF and RealEstate10K datasets with 2, 3 and 4 input views.
    We observe that Simple-NeRF achieves a SSIM in excess of 0.8 for scenes with spatial complexity less than $6$ across both the datasets.
    Further, the plots provide a coarse trade-off between the spatial complexity and the achievable SSIM score for a given number of input views.
    Thus, one could determine whether our model is likely to achieve a target SSIM score for a given scene by computing its spatial complexity.

    Next, we analyse the dependence of the improvement achieved by Simple-NeRF over the base NeRF model given its results.
    Specifically, we find that Simple-NeRF achieves a higher improvement when the base model has a moderate performance, but still suffers from floaters.
    We quantify this observation through the following analysis.
    We hypothesize that the improvement achieved by Simple-NeRF depends on both the SSIM score and the amount of floaters in the base model.
    Mathematically, if $s_b$ is the SSIM of the base model and $f_b$ is the fraction of pixels corresponding to floaters in the base model, we show a scatter plot between $\alpha s_b - \beta f_b$ and the improvement in SSIM in \autoref{fig:analysis-performance-improvement01}, where $\alpha$ and $\beta$ are normalization constants.
    We subtract $f_b$ from $s_b$, since they are complementary in nature.
    We determine the floater pixels as those whose depth is less than half of the median depth of the scene.
    We set $\alpha=0.2$ and $\beta=0.5$ through a simple grid search.
    We observe that the improvement achieved by Simple-NeRF increases as the base model performance improves and the amount of floaters decreases.
    The quantum of improvement reduces as the base model performance saturates and the amount of floaters reduces.

    \subsection{Limitations and Future Work}\label{subsec:limitations-future-work}
    Our approach of obtaining reliable depth supervision by learning simpler solutions is limited to the cases where the base model achieves a reasonable performance but suffers from various distortions due to overfitting.
    Our regularizations may not help significantly if the performance of the base model is very poor, as in the case of learning a highly complex scene with very few input views as shown in \cref{fig:qualitative-simple-zipnerf-limitations01}.
    In such cases, we can employ our regularizations on top of other sparse input radiance fields as we do for the NeRF-LLFF dataset.
    \cref{subsec:analysis} provides a couple of approaches to determine if our model is likely to help improve the performance for a given scene.

    Even though our method significantly reduces the floaters, some translucent floaters still persist as seen in the supplementary videos.
    Our approach provides supervision from the augmented models only at the pixels where reliable depth estimates are available.
    We determine the reliable depths by reprojecting to the closest training view and checking for consistency as explained in~\cref{subsubsec:reliable-depth-estimates}.
    Hence, in regions which are visible in only one training view, we do not have reliable depth supervision, which sometimes leads to transparent floaters.
    Further, we observe significant transparent floaters in the TensoRF model.
    Although we employ a mass concentration loss (\cref{eq:loss-tensorf-mass-concentration}) to increase opacity, we observe that the transparent objects are not completely removed.
    We also note that employing this loss in other models did not help in removing transparent floaters.
    Finally, our depth supervision loss is imposed only on the mean depth predicted by the model (\cref{eq:loss-augmentation}) and not on the variance of the distribution.
    We experimented with imposing an additional loss to reduce transparent floaters by minimizing the standard deviation of the predicted depth or a KL divergence loss between the predicted depth distribution and a delta function at the estimated depth prior~\cite{deng2022dsnerf}, we found that it led to several stronger artifacts.
    This perhaps requires a more sophisticated approach and is left for future work.

    Training augmented models adds compute and memory overhead during training.
    It would be interesting to explore deriving augmentations from the main model itself, without training an augmented model separately.
    For example, instead of training a separate augmented model with a lower resolution grid in Simple-TensoRF, one could try downsampling the grid from the main model.
    However, achieving the same with other augmentations is non-trivial and is left for future work.

    While we perform elementary modifications of radiance fields to obtain simpler solutions, it would be interesting to explore more sophisticated augmentations to learn simpler solutions.
    For example, one could explore inducing smoothness through different hash functions in Simple-ZipNeRF\@.
    Such sophisticated augmentations could help obtain larger improvements in performance.
    However, such explorations are beyond the scope of this work.

    Our approach to determining reliable depth estimates for supervision depends on the reprojection error, which may be high for specular objects even if the depth estimates are correct.
    It may be helpful to explore approaches to determine the reliability of depth estimates without employing the reprojection error.


    Our model requires accurate camera poses of the sparse input images.
    While the joint optimization of camera poses and scene geometry is explored when dense input views are available~\cite{park2023camp,jeong2021scnerf}, it would be helpful to explore the same with sparse input views.
    In our experiments, we found that a trivial combination of a pose optimization radiance field such as NeRF- -~\cite{wang2021nerfminusminus} and a sparse input radiance field such as Simple-NeRF or DS-NeRF leads to very poor performances.
    While some approaches~\cite{lin2021barf} for camera pose optimization are limited to NeRF and do not extend to explicit radiance fields, other approaches~\cite{bian2023nopenerf,truong2023sparf,han2024more} require pre-training on a large dataset.
    We believe this problem requires a deeper study and is a very important direction to be pursued in the future.

    Finally, our approach of employing augmentations to obtain better supervision can be explored in sparse input novel view synthesis of highly specular objects~\cite{verbin2022refnerf}, refractive objects~\cite{deng2024ray}, low-light scenes~\cite{mildenhall2022nerfinthedark}, blurred images~\cite{ma2022deblurnerf} and high dynamic range images~\cite{lu2024panonerf}.
    Further, it would be interesting to explore the use of augmentations in related inverse problems such as surface reconstruction~\cite{wang2021neus,long2022sparseneus}, dynamic view synthesis~\cite{pumarola2021dnerf,fridovich2023kplanes,somraj2024rfderf,somraj2022decompnet,choudhary2025bfderf} and style transfer~\cite{wang2024stylizing} when only sparse input viewpoints are available.

    \section{Conclusion}\label{sec:conclusion}
    We address the problem of few-shot radiance fields by obtaining depth supervision from simpler solutions learned by lower capability augmented models that are trained in tandem with the main radiance field model.
    We show that augmentations can be designed for both implicit models, such as NeRF, and explicit radiance fields, such as TensoRF and ZipNeRF\@.
    Since the shortcomings of various radiance fields are different, we design the augmentations appropriately for each model.
    We show that our augmentations improve performance significantly on all three models, and we achieve state-of-the-art performance on forward-facing scenes as well as 360$\degree$ scenes.
    Notably, our models achieve a significant improvement in the depth estimation of the scene, which indicates a superior geometry estimation.

\begin{acks}
    This work was supported in part by a grant from Qualcomm.
    The first author was supported by the Prime Minister’s Research Fellowship awarded by the Ministry of Education, Government of India.
\end{acks}

    \bibliographystyle{ACM-Reference-Format}
    \bibliography{SSLN_Journal}

\appendix
\twocolumn[\subsection*{\centering \fontsize{13}{15}\selectfont Supplement}]

    \noindent The contents of this supplement include
    \begin{enumerate}[label=\Alph*., noitemsep]
        \item Details on evaluation measures.
        \item Video examples on LLFF, RealEstate-10K and MipNeRF360 datasets.
        \item Additional analysis - positional encoding frequency, and depth reliability masks.
        \item Extensive quantitative evaluation reports.
    \end{enumerate}

    \section{Details on Evaluation Measures}\label{sec:details-evaluation}

    \subsection{Masked Evaluation}\label{subsec:masked-evaluation}
    As mentioned in the main paper, we evaluate the model predictions only in the regions visible in the training images.
    We now explain our reasoning behind masking the model predictions for evaluation and then provide the details of how we compute the masks.

    Recall that radiance fields are designed to memorize a scene and are therefore not equipped to predict unseen regions by design.
    Further, many regularization based sparse input NeRF models do not employ pre-trained prior.
    As a result, radiance fields are also ill-equipped to predict the depth of objects seen in only one of the input views, again by design.
    Thus, radiance fields require the objects to be visible in at least two views to estimate their 3D geometry accurately.
    Hence, we generate a mask that denotes the pixels visible in at least two input views, and evaluate the predictions in such regions only.

    We generate the mask by using the depths predicted by the dense input NeRF model as follows.
    For every train view, we warp its depth predicted by Dense-NeRF to every other test view and compare it with the Dense-NeRF predicted depth of the test view.
    Intuitively, if a pixel in a test view is visible in the considered train view, then the two depths should be close to each other.
    Thus, we threshold the depth difference to obtain the mask.
    That is, if the difference in the two depths is less than a threshold, then we mark the pixel in the test view as visible in the considered train view.
    Our final mask for every test view is generated by marking pixels as visible if they are visible in at least two input views.
    We warp the depth maps using depth based reprojection similar to \citet{cho2017hole,kanchana2022ivp} and set the threshold to $0.05$ times the maximum depth of the train view.
    By computing the mask using depths and not color, our approach is robust to the presence of specular objects, as long as the depth estimated by Dense-NeRF is accurate.
    We will release the code used to generate the masks and the masks along with the main code release.

    Nonetheless, we report the performance without masking the unseen regions in \cref{tab:quantitative-all-simplenerf-llff01,tab:quantitative-all-simplenerf-llff02,tab:quantitative-all-simplenerf-llff03,tab:quantitative-all-simplenerf-realestate01,tab:quantitative-all-simplenerf-realestate02,tab:quantitative-all-simplenerf-realestate03,tab:quantitative-all-simpletensorf-llff01,tab:quantitative-all-simpletensorf-realestate01}.

    \subsection{Difference in Evaluation Scores}\label{subsec:difference-in-evaluation-scores}
    We note that the evaluation scores of the comparison models differ from those reported in their respective papers due to the following reasons.
    \begin{itemize}
        \item As noted above, we mask out the regions not visible in at least two of the input views during evaluation.
        \item The number of train and test views used for experiments vary across the different papers.
        While they're uniformly picked in certain cases, they are hard-coded in other cases.
        Further, different papers use different resolution for the input images.
        For a reliable comparison across different models, we fix the same train and test views as well as the resolution and re-run all the benchmarked models using the author-provided source code and report the results.
        \item Finally, the SSIM scores vary across different implementations~\cite{venkataramanan2021hitchhikerssim}.
        Similarly, the LPIPS scores vary depending on the backbone used (Alex-net vs VGG-net).
        We use scikit-image implementation of SSIM and Alex-net backbone for LPIPS\@.
    \end{itemize}

    \section{Video Comparisons}\label{sec:video-comparisons}
    We compare various models by rendering videos along a continuous trajectory.
    For the LLFF dataset, we render the videos along the spiral trajectory that is commonly used in the literature.
    Since RealEstate-10K is a dataset of videos, we combine the train and test frames to get the continuous trajectories.
    For the MipNeRF360 dataset, we render the videos along the elliptical trajectory that is commonly used in the literature.

    We group the video comparisons based on Simple-NeRF, Simple-TensoRF and Simple-ZipNeRF models.
    In each group, we divide the video comparisons into two sets.
    In the first set, we show how our regularizations reduce various artifacts by comparing the videos rendered by our model with those of the competing models and the ablated models.
    In the second set, we compare the videos rendered by our model with those of the competing models.
    Finally, we also include a few videos to support certain arguments made in the main paper.
    The videos are available on our project website \url{https://nagabhushansn95.github.io/publications/2024/Simple-RF.html}

    \section{Additional Analysis}\label{sec:additional-analysis}

    \subsection{Ablation on Positional Encoding Frequency in Simple-NeRF}\label{subsec:ablation-positional-encoding-frequency}
    \begin{figure}
        \includegraphics[width=\linewidth]{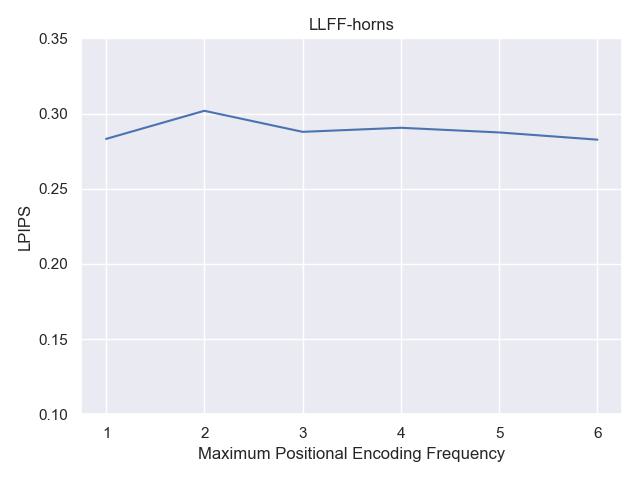}
        \caption{LPIPS scores on the horns scene as $l_p^{\text{ap}}$ of Simple-NeRF is varied.}
        \label{fig:quantitative-simple-nerf-pe-degree}
    \end{figure}
    We analyze the variation in the performance of Simple-NeRF as $l_p^{\text{as}}$ varies.
    We vary $l_p^{\text{as}}$ from $1$ to $6$ and test the performance of Simple-NeRF on the horns scene of the NeRF-LLFF dataset.
    We show the quantitative performance in terms of LPIPS in \cref{fig:quantitative-simple-nerf-pe-degree}.
    We observe only small variations in the performance as $l_p^{\text{ap}}$ is varied, and thus, we conclude that our framework is robust to the choice of $l_p^{\text{ap}}$.
    Further, we note that using $l_p^{\text{ap}} = l_p = 10$ is equivalent to using an identical augmentation.

    \subsection{Visualization of Depth Reliability Masks}\label{subsec:visualization-depth-reliability-masks}
    \begin{figure*}
        \centering
        \begin{subfigure}[t]{0.48\linewidth}
            \centering
            \includegraphics[width=\linewidth]{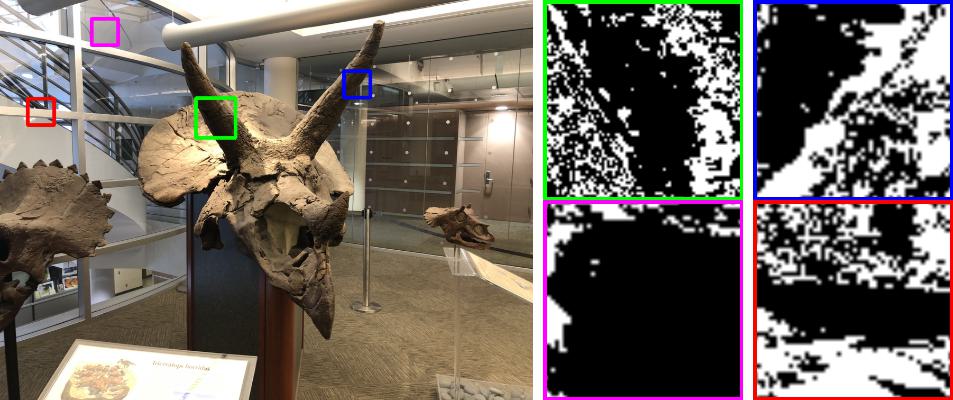}
            \caption{\textbf{Smoothing augmentation:} The green and blue boxes focus on the two horns, where we observe that the augmented model depth is preferred in the depth-wise smooth regions on horns, and the main model depth is preferred at the edges. The magenta box focuses on a completely smooth region, so the augmented model depth is preferred for most pixels. In the red box, augmented model depth is preferred along the horizontal bar. The main model depth is preferred on either side of the bar that contains multiple depth discontinuities.}
            \label{fig:smoothing-augmentation-visualization}
        \end{subfigure}
        \hfill
        \begin{subfigure}[t]{0.48\linewidth}
            \centering
            \includegraphics[width=\linewidth]{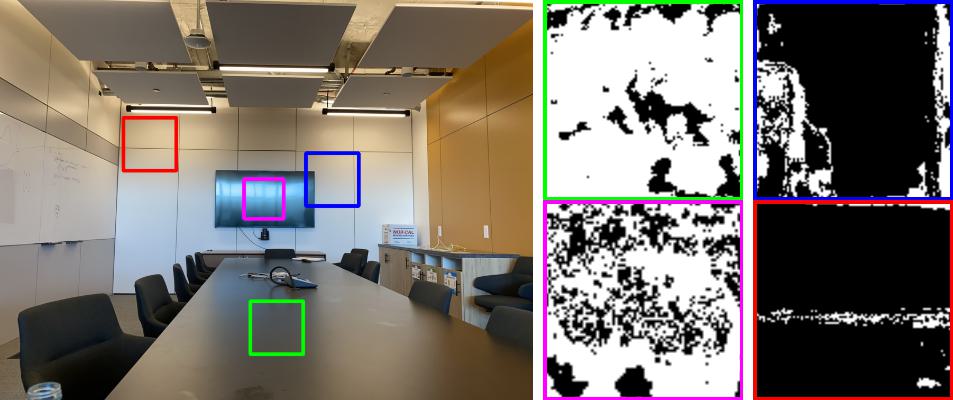}
            \caption{\textbf{Lambertian augmentation:} The green and magenta boxes focus on the TV and the table, respectively, which are highly specular in this scene (please view the supplementary videos of the room scene to observe the specularity of these objects).
            In these regions, the main model depth is determined to be more accurate since the main model can handle specular regions.
            The red and blue boxes focus on Lambertian regions of the scene where the depth estimated by the augmented model is preferred.
            }
            \label{fig:lambertian-augmentation-visualization}
        \end{subfigure}
        \caption{Visualizations of depth reliability mask for the two augmentations of Simple-NeRF.
        White pixels in the mask indicate that the main model depth is determined to be more accurate at the corresponding locations.
        Black pixels indicate that the augmented model depth is determined to be more accurate.}
        \label{fig:augmentations-visualizations}
    \end{figure*}
    In \cref{fig:augmentations-visualizations}, we present visualizations that motivate the design of our augmentations in Simple-NeRF, namely the smoothing and Lambertian augmentations.
    We train our model without augmentations and the individual augmentations separately with only $\lossColor$ and $\lossSparseDepth$ for 100k iterations.
    Using the depth maps predicted by the models for an input training view, we determine the mask that indicates which depth estimates are more accurate, as explained in Sec 4.1.4 of the main paper.
    For two scenes from the LLFF dataset, we show an input training view and focus on a small region to visualize the corresponding masks.

    We observe that the smoothing augmentation is determined to have estimated better depths in smooth regions.
    At edges, the depth estimated by the main model is more accurate.
    Similarly, the Lambertian augmentation estimates better depth in Lambertian regions, while the main model estimates better depth in specular regions.
    We note that the masks shown are not the masks obtained by our final model.
    Since the masks are computed at every iteration, and the training of the main and augmented models are coupled, it is not possible to determine the exact locations where the augmented models help the main model learn better.

    \section{Performance on Individual Scenes}\label{sec:performance-on-individual-scenes}
    For the benefit of follow-up work, where researchers may want to analyze the performance of different models or compare the models on individual scenes, we provide the performance of various models on individual scenes in \cref{tab:quantitative-scene-wise-simplenerf-llff01,tab:quantitative-scene-wise-simplenerf-llff02,tab:quantitative-scene-wise-simplenerf-llff03,tab:quantitative-scene-wise-simplenerf-llff04,tab:quantitative-scene-wise-simplenerf-realestate01,tab:quantitative-scene-wise-simplenerf-realestate02,tab:quantitative-scene-wise-simplenerf-realestate03,tab:quantitative-scene-wise-simplenerf-realestate04,tab:quantitative-scene-wise-simpletensorf-llff01,tab:quantitative-scene-wise-simpletensorf-realestate01,tab:quantitative-scene-wise-simplezipnerf-mipnerf360-01,tab:quantitative-scene-wise-simplezipnerf-mipnerf360-02,tab:quantitative-scene-wise-simplezipnerf-mipnerf360-03,tab:quantitative-scene-wise-simplezipnerf-nerfsynthetic01,tab:quantitative-scene-wise-simplezipnerf-nerfsynthetic02,tab:quantitative-scene-wise-simplezipnerf-nerfsynthetic03}.

    \clearpage
    \begin{table*}
        \centering
        \caption{Quantitative Results of NeRF based models on LLFF dataset with two input views. The values within parenthesis show unmasked scores.}
\begin{tabular}{l|ccc|cc}
            \hline
            \textbf{Model} & \textbf{LPIPS} & \textbf{SSIM} & \textbf{PSNR} & \textbf{Depth MAE}  & \textbf{Depth SROCC} \\
            \hline
            InfoNeRF     &  0.6024(0.7561) &  0.2219(0.2095) &  9.16(9.23) &  0.9797(1.1000) &  -0.0188(-0.0092)  \\
            DietNeRF     &  0.5465(0.7265) &  0.3283(0.3209) &  11.94(11.89) &  0.8886(1.0105) &  -0.0099(-0.0045)  \\
            RegNeRF      &  0.3056(0.4297) &  0.5712(0.4885) &  18.52(16.88) &  -- &  0.7141(0.6513)  \\
            DS-NeRF      &  0.3106(0.4176) &  0.5862(0.5074) &  18.24(16.93) &  0.2074(0.3372) &  0.7230(0.5787)  \\
            DDP-NeRF     &  0.2851(0.3920) &  0.6218(0.5424) &  18.73(17.19) &  0.2048(0.3494) &  0.7480(0.6109)  \\
            FreeNeRF     &  \textbf{0.2638}(0.3760) &  0.6322(0.5432) &  19.52(17.55) &  -- &  0.8066(0.7137)  \\
            ViP-NeRF     &  0.2768(\textbf{0.3725}) &  0.6225(0.5230) &  18.61(16.66) &  0.1999(0.3413) &  0.7344(0.6221)  \\
\hline
        \end{tabular}
        \label{tab:quantitative-all-simplenerf-llff01}
    \end{table*}

    \begin{table*}
        \centering
        \caption{Quantitative Results of NeRF based models on LLFF dataset with three input views. The values within parenthesis show unmasked scores.}
        \begin{tabular}{l|ccc|cc}
            \hline
            \textbf{Model} & \textbf{LPIPS} & \textbf{SSIM} & \textbf{PSNR} & \textbf{Depth MAE}  & \textbf{Depth SROCC} \\
            \hline
            InfoNeRF     &  0.6732(0.7679) &  0.1953(0.1859) &  8.38(8.52) &  1.0012(1.1149) &  -0.0144(-0.0176)  \\
            DietNeRF     &  0.6120(0.7254) &  0.3405(0.3297) &  11.76(11.77) &  0.9093(1.0242) &  -0.0598(-0.0471)  \\
            RegNeRF      &  0.2908(0.3602) &  0.6334(0.5677) &  20.22(18.65) &  -- &  0.8238(0.7589)  \\
            DS-NeRF      &  0.3031(0.3641) &  0.6321(0.5774) &  20.20(18.97) &  0.1787(0.2699) &  0.7852(0.7173)  \\
            DDP-NeRF     &  0.3250(0.3869) &  0.6152(0.5628) &  18.73(17.71) &  0.1941(0.3032) &  0.7433(0.6707)  \\
            FreeNeRF     &  0.2754(0.3415) &  0.6583(0.5960) &  20.93(19.30) &  -- &  0.8379(0.7656)  \\
            ViP-NeRF     &  0.2798(0.3365) &  0.6548(0.5907) &  20.54(18.89) &  0.1721(0.2795) &  0.7891(0.7082)  \\
            InstantSplat &  0.2974(0.3364) &  0.4603(0.4227) &  17.49(16.34) &  -- &  --  \\
            CoR-GS       &  \textbf{0.1860}(\textbf{0.2409}) &  \textbf{0.7332}(\textbf{0.6656}) &  21.25(\textbf{19.49}) &  -- &  --  \\
            \hline
            Simple-NeRF  &  0.2559(0.3259) &  0.6940(0.6222) &  \textbf{21.37}(19.47) &  \textbf{0.1199}(\textbf{0.2201}) &  \textbf{0.8935}(\textbf{0.8153})  \\
            \hline
        \end{tabular}
        \label{tab:quantitative-all-simplenerf-llff02}
    \end{table*}

    \begin{table*}
        \centering
        \caption{Quantitative Results of NeRF based models on LLFF dataset with four input views. The values within parenthesis show unmasked scores.}
        \begin{tabular}{l|ccc|cc}
            \hline
            \textbf{Model} & \textbf{LPIPS} & \textbf{SSIM} & \textbf{PSNR} & \textbf{Depth MAE}  & \textbf{Depth SROCC} \\
            \hline
            InfoNeRF     &  0.6985(0.7701) &  0.2270(0.2188) &  9.18(9.25) &  1.0411(1.1119) &  -0.0394(-0.0390)  \\
            DietNeRF     &  0.6506(0.7396) &  0.3496(0.3404) &  11.86(11.84) &  0.9546(1.0259) &  -0.0368(-0.0249)  \\
            RegNeRF      &  0.2794(0.3227) &  0.6645(0.6159) &  21.32(19.89) &  -- &  0.8933(0.8528)  \\
            DS-NeRF      &  0.2979(0.3376) &  0.6582(0.6135) &  21.23(20.07) &  0.1451(0.2097) &  0.8506(0.8130)  \\
            DDP-NeRF     &  0.3042(0.3467) &  0.6558(0.6121) &  20.17(19.19) &  0.1704(0.2487) &  0.8322(0.7664)  \\
            FreeNeRF     &  0.2848(0.3280) &  0.6764(0.6303) &  21.91(20.45) &  -- &  0.9091(0.8626)  \\
            ViP-NeRF     &  0.2854(0.3203) &  0.6675(0.6182) &  20.75(19.34) &  0.1555(0.2316) &  0.8622(0.8070)  \\
            InstantSplat &  0.3058(0.3314) &  0.4818(0.4488) &  17.81(16.68) &  -- &  --  \\
            CoR-GS       &  \textbf{0.1694}(\textbf{0.2070}) &  \textbf{0.7560}(\textbf{0.7091}) &  \textbf{22.44}(\textbf{20.87}) &  -- &  --  \\
            \hline
            Simple-NeRF  &  0.2633(0.3083) &  0.7016(0.6521) &  21.99(20.44) &  \textbf{0.1110}(\textbf{0.1741}) &  \textbf{0.9355}(\textbf{0.8952})  \\
            \hline
        \end{tabular}
        \label{tab:quantitative-all-simplenerf-llff03}
    \end{table*}

    \begin{table*}
        \centering
        \caption{Quantitative Results of NeRF based models on RealEstate-10K dataset with two input views. The values within parenthesis show unmasked scores.}
        \begin{tabular}{l|ccc|cc}
            \hline
            \textbf{Model} & \textbf{LPIPS} & \textbf{SSIM} & \textbf{PSNR} & \textbf{Depth MAE}  & \textbf{Depth SROCC} \\
            \hline
            InfoNeRF     &  0.5924(0.6384) &  0.4342(0.4343) &  12.27(12.17) &  2.1793(2.2314) &  0.0912(0.0942)  \\
            DietNeRF     &  0.4381(0.4862) &  0.6534(0.6520) &  18.06(17.83) &  2.0247(2.0807) &  0.2339(0.2452)  \\
            RegNeRF      &  0.4129(0.4483) &  0.5916(0.5864) &  17.14(16.87) &  -- &  0.1118(0.1117)  \\
            DS-NeRF      &  0.2709(0.3171) &  0.7983(0.7859) &  26.26(25.44) &  0.7164(0.8323) &  0.6660(0.6433)  \\
            DDP-NeRF     &  0.1290(0.1481) &  0.8640(0.8502) &  27.79(26.15) &  0.4831(0.5944) &  0.7921(0.7663)  \\
            FreeNeRF     &  0.5036(0.5471) &  0.5354(0.5336) &  14.70(14.50) &  -- &  -0.1937(-0.1813)  \\
            ViP-NeRF     &  0.0687(0.0783) &  0.8889(0.8717) &  32.32(29.55) &  0.3856(0.5337) &  0.8446(0.7851)  \\
\hline
        \end{tabular}
        \label{tab:quantitative-all-simplenerf-realestate01}
    \end{table*}

    \begin{table*}
        \centering
        \caption{Quantitative Results of NeRF based models on RealEstate-10K dataset with three input views. The values within parenthesis show unmasked scores.}
        \begin{tabular}{l|ccc|cc}
            \hline
            \textbf{Model} & \textbf{LPIPS} & \textbf{SSIM} & \textbf{PSNR} & \textbf{Depth MAE}  & \textbf{Depth SROCC} \\
            \hline
            InfoNeRF     &  0.6561(0.6846) &  0.3792(0.3780) &  10.57(10.57) &  2.2198(2.2830) &  0.1929(0.1994)  \\
            DietNeRF     &  0.4636(0.4886) &  0.6456(0.6445) &  18.01(17.89) &  2.0355(2.1023) &  0.0240(0.0438)  \\
            RegNeRF      &  0.4171(0.4362) &  0.6132(0.6078) &  17.86(17.73) &  -- &  0.0574(0.0475)  \\
            DS-NeRF      &  0.2893(0.3211) &  0.8004(0.7905) &  26.50(25.94) &  0.5400(0.6524) &  0.8106(0.7910)  \\
            DDP-NeRF     &  0.1518(0.1601) &  0.8587(0.8518) &  26.67(25.92) &  0.4139(0.5222) &  0.8612(0.8331)  \\
            FreeNeRF     &  0.5146(0.5414) &  0.5708(0.5675) &  15.26(15.12) &  -- &  -0.2590(-0.2445)  \\
            ViP-NeRF     &  0.0758(0.0832) &  0.8967(0.8852) &  31.93(30.27) &  0.3365(0.4683) &  0.9009(0.8558)  \\
            InstantSplat &  0.0925(0.1031) &  0.8761(0.8646) &  30.01(27.69) &  -- &  --  \\
            CoR-GS       &  0.2660(0.2777) &  0.7131(0.7050) &  20.58(20.34) &  --  &  --  \\
            \hline
            Simple-NeRF  &  \textbf{0.0726}(\textbf{0.0829}) &  \textbf{0.8984}(\textbf{0.8879}) &  \textbf{33.21}(\textbf{31.40}) &  \textbf{0.2770}(\textbf{0.3885}) &  \textbf{0.9266}(\textbf{0.8931})  \\
            \hline
        \end{tabular}
        \label{tab:quantitative-all-simplenerf-realestate02}
    \end{table*}

    \begin{table*}
        \centering
        \caption{Quantitative Results of NeRF based models on RealEstate-10K dataset with four input views. The values within parenthesis show unmasked scores.}
        \begin{tabular}{l|ccc|cc}
            \hline
            \textbf{Model} & \textbf{LPIPS} & \textbf{SSIM} & \textbf{PSNR} & \textbf{Depth MAE}  & \textbf{Depth SROCC} \\
            \hline
            InfoNeRF     &  0.6651(0.6721) &  0.3843(0.3830) &  10.62(10.59) &  2.1874(2.2742) &  0.2549(0.2594)  \\
            DietNeRF     &  0.4853(0.4954) &  0.6503(0.6475) &  18.01(17.89) &  2.0398(2.1273) &  0.0990(0.1011)  \\
            RegNeRF      &  0.4316(0.4383) &  0.6257(0.6198) &  18.34(18.25) &  -- &  0.1422(0.1396)  \\
            DS-NeRF      &  0.3103(0.3287) &  0.7999(0.7920) &  26.65(26.28) &  0.5154(0.6171) &  0.8145(0.8018)  \\
            DDP-NeRF     &  0.1563(0.1584) &  0.8617(0.8557) &  27.07(26.48) &  0.3832(0.4813) &  0.8739(0.8605)  \\
            FreeNeRF     &  0.5226(0.5323) &  0.6027(0.5989) &  16.31(16.25) &  -- &  -0.2152(-0.2162)  \\
            ViP-NeRF     &  0.0892(0.0909) &  0.8968(0.8894) &  31.95(30.83) &  0.3658(0.4761) &  0.8414(0.8080)  \\
            InstantSplat &  0.1618(0.1682) &  0.8643(0.8590) &  29.60(28.56) &  -- &  --  \\
            CoR-GS       &  0.2953(0.3001) &  0.7119(0.7052) &  20.33(20.18) &  --  &  --  \\
            \hline
            Simple-NeRF  &  \textbf{0.0847}(\textbf{0.0891}) &  \textbf{0.8987}(\textbf{0.8917}) &  \textbf{32.88}(\textbf{31.73}) &  \textbf{0.2692}(\textbf{0.3565}) &  \textbf{0.9209}(\textbf{0.9035})  \\
            \hline
        \end{tabular}
        \label{tab:quantitative-all-simplenerf-realestate03}
    \end{table*}
    \clearpage

    \begin{table*}
        \centering
        \caption{Quantitative Results of TensoRF based models on LLFF dataset with three input views. The values within parenthesis show unmasked scores.}
        \begin{tabular}{l|ccc|cc}
            \hline
            \textbf{Model} & \textbf{LPIPS} & \textbf{SSIM} & \textbf{PSNR} & \textbf{Depth MAE}  & \textbf{Depth SROCC} \\
            \hline
            TensoRF      &  0.5474(0.6177) &  0.3163(0.3002) &  12.29(12.14) &  0.6682(0.7824) &  0.0252(0.0324)  \\
            DS-TensoRF   &  0.2897(0.3549) &  0.6291(0.5702) &  18.58(17.41) &  0.2276(0.3483) &  0.7279(0.6478)  \\
            Simple-TensoRF &  \textbf{0.2461}(\textbf{0.3037}) &  \textbf{0.6749}(\textbf{0.6099}) &  \textbf{20.22}(\textbf{18.71}) &  \textbf{0.1671}(\textbf{0.2897}) &  \textbf{0.8272}(\textbf{0.7451})  \\
            \hline
            \makecell[l]{Simple-TensoRF w/ $R_\sigma^s = R_\sigma$} &  0.2536(0.3136) &  0.6677(0.6020) &  19.85(18.44) &  0.1777(0.2986) &  0.8112(0.7298)  \\
            \makecell[l]{Simple-TensoRF w/ $\numTensorfVoxels[s] = \numTensorfVoxels$} &  0.2568(0.3182) &  0.6579(0.5946) &  19.95(18.54) &  0.1902(0.3106) &  0.7945(0.7128)  \\
            \makecell[l]{Simple-TensoRF w/ $R_\sigma^s = R_\sigma; \numTensorfVoxels[s] = \numTensorfVoxels$} &  0.2728(0.3356) &  0.6424(0.5814) &  19.50(18.25) &  0.2190(0.3425) &  0.7446(0.6625)  \\
            \hline
        \end{tabular}
        \label{tab:quantitative-all-simpletensorf-llff01}
    \end{table*}

    \begin{table*}
        \centering
        \caption{Quantitative Results of TensoRF based models on RealEstate-10K dataset with three input views. The values within parenthesis show unmasked scores.}
        \begin{tabular}{l|ccc|cc}
            \hline
            \textbf{Model} & \textbf{LPIPS} & \textbf{SSIM} & \textbf{PSNR} & \textbf{Depth MAE}  & \textbf{Depth SROCC} \\
            \hline
            TensoRF      &  0.0986(0.1050) &  0.8532(0.8427) &  29.62(28.00) &  0.4394(0.4841) &  0.6314(0.6077)  \\
            DS-TensoRF   &  0.0739(0.0827) &  0.8872(0.8748) &  32.50(30.20) &  0.2720(0.3321) &  0.7527(0.7227)  \\
            Simple-TensoRF &  \textbf{0.0706}(\textbf{0.0780}) &  \textbf{0.8920}(\textbf{0.8809}) &  \textbf{32.70}(\textbf{30.79}) &  0.2229(\textbf{0.2882}) &  0.7983(0.7621)  \\
            \hline
            \makecell[l]{Simple-TensoRF w/ $R_\sigma^s = R_\sigma$} &  0.0850(0.0919) &  0.8821(0.8709) &  30.94(29.51) &  0.2742(0.3317) &  0.7650(0.7238)  \\
            \makecell[l]{Simple-TensoRF w/ $\numTensorfVoxels[s] = \numTensorfVoxels$} &  0.0735(0.0809) &  0.8896(0.8773) &  32.22(30.29) &  \textbf{0.2200}(0.2888) &  \textbf{0.8223}(\textbf{0.7751})  \\
            \makecell[l]{Simple-TensoRF w/ $R_\sigma^s = R_\sigma; \numTensorfVoxels[s] = \numTensorfVoxels$} &  0.0787(0.0867) &  0.8871(0.8743) &  31.73(29.88) &  0.2346(0.3083) &  0.7939(0.7458)  \\
            \hline
        \end{tabular}
        \label{tab:quantitative-all-simpletensorf-realestate01}
    \end{table*}

    \begin{table*}
        \centering
        \caption{Per-scene performance of various NeRF based models with two input views on LLFF dataset.
        The five rows show LPIPS, SSIM, PSNR, Depth MAE and Depth SROCC scores, respectively.
        The values within parenthesis show unmasked scores.}
        \begin{tabular}{l|c|c|c|c|c|c|c|c|c}
            \hline
            \textbf{\makecell{Model \textbackslash \\ Scene Name}} & \textbf{Fern} & \textbf{Flower} & \textbf{Fortress} & \textbf{Horns} & \textbf{Leaves} & \textbf{Orchids} & \textbf{Room} & \textbf{Trex} & \textbf{Average} \\
            \hline
            InfoNeRF     & \makecell{ 0.66(0.78)\\ 0.26(0.26)\\ 10.7(11.0)\\ 0.96(1.05)\\ -0.00(-0.04)} & \makecell{ 0.57(0.69)\\ 0.21(0.19)\\ 10.8(11.0)\\ 0.76(0.89)\\ -0.25(-0.18)} & \makecell{ 0.65(0.82)\\ 0.18(0.18)\\ 6.6(6.5)\\ 1.18(1.41)\\ 0.09(0.06)} & \makecell{ 0.58(0.76)\\ 0.20(0.20)\\ 8.9(8.9)\\ 1.08(1.26)\\ 0.07(0.08)} & \makecell{ 0.45(0.64)\\ 0.15(0.10)\\ 9.7(9.5)\\ 0.85(0.92)\\ -0.14(-0.12)} & \makecell{ 0.60(0.69)\\ 0.17(0.11)\\ 9.1(9.4)\\ 0.87(1.12)\\ 0.00(0.01)} & \makecell{ 0.68(0.80)\\ 0.36(0.38)\\ 10.4(10.8)\\ 1.05(1.06)\\ -0.01(0.00)} & \makecell{ 0.61(0.79)\\ 0.23(0.21)\\ 8.5(8.4)\\ 0.94(0.95)\\ -0.00(0.01)} & \makecell{ 0.60(0.76)\\ 0.22(0.21)\\ 9.2(9.2)\\ 0.98(1.10)\\ -0.02(-0.01)} \\
            \hline
            DietNeRF     & \makecell{ 0.63(0.77)\\ 0.30(0.29)\\ 12.1(12.3)\\ 0.83(0.92)\\ -0.02(-0.05)} & \makecell{ 0.54(0.69)\\ 0.29(0.26)\\ 12.1(12.2)\\ 0.70(0.84)\\ -0.02(0.00)} & \makecell{ 0.50(0.66)\\ 0.44(0.44)\\ 15.2(14.2)\\ 0.94(1.17)\\ 0.13(0.13)} & \makecell{ 0.54(0.76)\\ 0.27(0.28)\\ 10.5(10.7)\\ 1.02(1.20)\\ -0.01(-0.01)} & \makecell{ 0.43(0.67)\\ 0.18(0.12)\\ 10.8(10.6)\\ 0.83(0.90)\\ -0.13(-0.10)} & \makecell{ 0.57(0.73)\\ 0.21(0.15)\\ 10.5(10.6)\\ 0.78(1.03)\\ 0.02(-0.01)} & \makecell{ 0.62(0.77)\\ 0.49(0.52)\\ 13.0(13.1)\\ 0.95(0.97)\\ 0.02(0.03)} & \makecell{ 0.54(0.75)\\ 0.35(0.37)\\ 11.1(11.3)\\ 0.89(0.90)\\ -0.09(-0.07)} & \makecell{ 0.55(0.73)\\ 0.33(0.32)\\ 11.9(11.9)\\ 0.89(1.01)\\ -0.01(-0.00)} \\
\hline
        \end{tabular}
        \label{tab:quantitative-scene-wise-simplenerf-llff01}
    \end{table*}

    \begin{table*}
        \centering
        \caption{Per-scene performance of various NeRF based models with three input views on LLFF dataset.
        The five rows show LPIPS, SSIM, PSNR, Depth MAE and Depth SROCC scores, respectively.
        The values within parenthesis show unmasked scores.}
        \begin{tabular}{l|c|c|c|c|c|c|c|c|c}
            \hline
            \textbf{\makecell{Model \textbackslash \\ Scene Name}} & \textbf{Fern} & \textbf{Flower} & \textbf{Fortress} & \textbf{Horns} & \textbf{Leaves} & \textbf{Orchids} & \textbf{Room} & \textbf{Trex} & \textbf{Average} \\
            \hline
            InfoNeRF     & \makecell{ 0.74(0.86)\\ 0.22(0.21)\\ 7.5(7.5)\\ 0.99(1.08)\\ 0.01(-0.07)} & \makecell{ 0.62(0.69)\\ 0.24(0.23)\\ 10.3(10.8)\\ 0.84(0.94)\\ -0.18(-0.11)} & \makecell{ 0.76(0.83)\\ 0.16(0.17)\\ 5.1(5.2)\\ 1.21(1.42)\\ 0.17(0.17)} & \makecell{ 0.63(0.76)\\ 0.18(0.18)\\ 8.9(8.9)\\ 1.07(1.26)\\ 0.04(0.05)} & \makecell{ 0.50(0.65)\\ 0.16(0.11)\\ 10.2(9.9)\\ 0.87(0.92)\\ -0.12(-0.10)} & \makecell{ 0.68(0.71)\\ 0.12(0.08)\\ 7.9(8.3)\\ 0.95(1.15)\\ 0.07(0.08)} & \makecell{ 0.75(0.83)\\ 0.29(0.29)\\ 8.7(8.9)\\ 1.04(1.07)\\ -0.11(-0.15)} & \makecell{ 0.69(0.79)\\ 0.19(0.18)\\ 8.6(8.7)\\ 0.93(0.96)\\ -0.03(-0.06)} & \makecell{ 0.67(0.77)\\ 0.20(0.19)\\ 8.4(8.5)\\ 1.00(1.11)\\ -0.01(-0.02)} \\
            \hline
            DietNeRF     & \makecell{ 0.68(0.80)\\ 0.33(0.32)\\ 11.9(12.0)\\ 0.86(0.95)\\ -0.06(-0.03)} & \makecell{ 0.58(0.67)\\ 0.33(0.31)\\ 13.2(13.1)\\ 0.73(0.84)\\ -0.16(-0.15)} & \makecell{ 0.63(0.70)\\ 0.40(0.41)\\ 12.6(12.4)\\ 1.00(1.22)\\ -0.08(-0.09)} & \makecell{ 0.60(0.75)\\ 0.28(0.29)\\ 10.9(11.1)\\ 1.03(1.21)\\ 0.03(0.06)} & \makecell{ 0.49(0.67)\\ 0.18(0.13)\\ 10.8(10.6)\\ 0.85(0.90)\\ -0.03(-0.03)} & \makecell{ 0.65(0.75)\\ 0.19(0.15)\\ 10.0(10.1)\\ 0.86(1.06)\\ -0.07(-0.08)} & \makecell{ 0.68(0.76)\\ 0.54(0.55)\\ 12.6(12.6)\\ 0.95(0.97)\\ 0.03(0.09)} & \makecell{ 0.60(0.72)\\ 0.36(0.35)\\ 11.8(11.9)\\ 0.88(0.91)\\ -0.16(-0.17)} & \makecell{ 0.61(0.73)\\ 0.34(0.33)\\ 11.8(11.8)\\ 0.91(1.02)\\ -0.06(-0.05)} \\
\hline
        \end{tabular}
        \label{tab:quantitative-scene-wise-simplenerf-llff02}
    \end{table*}

    \begin{table*}
        \centering
        \caption{Per-scene performance of various NeRF based models with four input views on LLFF dataset.
        The five rows show LPIPS, SSIM, PSNR, Depth MAE and Depth SROCC scores, respectively.
        The values within parenthesis show unmasked scores.}
        \begin{tabular}{l|c|c|c|c|c|c|c|c|c}
            \hline
            \textbf{\makecell{Model \textbackslash \\ Scene Name}} & \textbf{Fern} & \textbf{Flower} & \textbf{Fortress} & \textbf{Horns} & \textbf{Leaves} & \textbf{Orchids} & \textbf{Room} & \textbf{Trex} & \textbf{Average} \\
            \hline
            InfoNeRF     & \makecell{ 0.71(0.80)\\ 0.20(0.19)\\ 9.6(9.8)\\ 0.95(1.03)\\ 0.02(0.00)} & \makecell{ 0.63(0.68)\\ 0.24(0.23)\\ 11.4(11.5)\\ 0.83(0.91)\\ -0.27(-0.21)} & \makecell{ 0.84(0.87)\\ 0.16(0.16)\\ 4.7(4.7)\\ 1.29(1.41)\\ 0.09(0.08)} & \makecell{ 0.68(0.77)\\ 0.19(0.19)\\ 8.9(8.8)\\ 1.18(1.27)\\ 0.08(0.09)} & \makecell{ 0.58(0.66)\\ 0.12(0.09)\\ 9.4(9.3)\\ 0.89(0.92)\\ -0.12(-0.11)} & \makecell{ 0.72(0.76)\\ 0.11(0.09)\\ 7.8(8.1)\\ 1.00(1.18)\\ 0.03(0.04)} & \makecell{ 0.72(0.78)\\ 0.45(0.45)\\ 11.9(11.9)\\ 1.05(1.06)\\ -0.06(-0.06)} & \makecell{ 0.69(0.80)\\ 0.26(0.26)\\ 9.9(10.1)\\ 0.97(0.97)\\ -0.13(-0.17)} & \makecell{ 0.70(0.77)\\ 0.23(0.22)\\ 9.2(9.2)\\ 1.04(1.11)\\ -0.04(-0.04)} \\
            \hline
            DietNeRF     & \makecell{ 0.66(0.80)\\ 0.36(0.35)\\ 12.7(12.9)\\ 0.85(0.93)\\ -0.11(-0.08)} & \makecell{ 0.62(0.69)\\ 0.31(0.29)\\ 12.6(12.6)\\ 0.76(0.84)\\ -0.05(-0.06)} & \makecell{ 0.67(0.70)\\ 0.42(0.42)\\ 12.8(12.6)\\ 1.07(1.19)\\ -0.04(-0.04)} & \makecell{ 0.66(0.78)\\ 0.30(0.30)\\ 10.9(10.8)\\ 1.13(1.22)\\ -0.00(0.02)} & \makecell{ 0.57(0.70)\\ 0.18(0.14)\\ 11.0(10.8)\\ 0.87(0.90)\\ -0.12(-0.10)} & \makecell{ 0.69(0.77)\\ 0.20(0.16)\\ 10.2(10.1)\\ 0.88(1.06)\\ -0.02(-0.02)} & \makecell{ 0.68(0.73)\\ 0.59(0.59)\\ 14.0(13.9)\\ 0.98(0.99)\\ 0.08(0.12)} & \makecell{ 0.64(0.75)\\ 0.34(0.35)\\ 10.9(11.2)\\ 0.91(0.92)\\ -0.10(-0.10)} & \makecell{ 0.65(0.74)\\ 0.35(0.34)\\ 11.9(11.8)\\ 0.95(1.03)\\ -0.04(-0.02)} \\
\hline
        \end{tabular}
        \label{tab:quantitative-scene-wise-simplenerf-llff03}
    \end{table*}

    \begin{table*}
        \centering
        \caption{Per-scene performance of Simple-NeRF ablated models with two input views on LLFF dataset.
        The five rows show LPIPS, SSIM, PSNR, Depth MAE and Depth SROCC scores, respectively.
        The values within parenthesis show unmasked scores.}
        \begin{tabular}{l|c|c|c|c|c|c|c|c|c}
            \hline
            \textbf{\makecell{Model \textbackslash \\ Scene Name}} & \textbf{Fern} & \textbf{Flower} & \textbf{Fortress} & \textbf{Horns} & \textbf{Leaves} & \textbf{Orchids} & \textbf{Room} & \textbf{Trex} & \textbf{Average} \\
            \hline
            Simple-NeRF   & \makecell{ 0.39(0.51)\\ 0.54(0.50)\\ 17.4(17.0)\\ 0.14(0.21)\\ \textbf{0.72}(0.68)} & \makecell{ 0.28(0.43)\\ \textbf{0.65}(0.53)\\ 20.1(16.9)\\ 0.10(0.25)\\ 0.94(0.72)} & \makecell{ 0.16(0.25)\\ 0.72(0.67)\\ \textbf{23.8}(\textbf{22.5})\\ 0.05(0.25)\\ \textbf{0.99}(\textbf{0.99})} & \makecell{ \textbf{0.26}(\textbf{0.42})\\ 0.64(0.54)\\ 19.6(17.1)\\ \textbf{0.18}(0.35)\\ \textbf{0.88}(\textbf{0.76})} & \makecell{ 0.25(0.44)\\ 0.45(0.30)\\ 15.2(13.5)\\ 0.31(0.44)\\ 0.64(0.38)} & \makecell{ \textbf{0.33}(0.41)\\ 0.52(0.37)\\ \textbf{16.3}(\textbf{14.7})\\ \textbf{0.15}(0.34)\\ \textbf{0.85}(\textbf{0.74})} & \makecell{ \textbf{0.27}(\textbf{0.35})\\ \textbf{0.83}(\textbf{0.77})\\ \textbf{22.5}(\textbf{19.5})\\ 0.12(\textbf{0.21})\\ 0.86(0.84)} & \makecell{ 0.28(0.39)\\ 0.68(0.58)\\ 18.3(16.8)\\ 0.13(0.19)\\ 0.78(0.73)} & \makecell{ \textbf{0.27}(\textbf{0.39})\\ \textbf{0.65}(\textbf{0.55})\\ \textbf{19.6}(\textbf{17.6})\\ \textbf{0.14}(0.28)\\ \textbf{0.85}(\textbf{0.75})} \\
\hline
        \end{tabular}
        \label{tab:quantitative-scene-wise-simplenerf-llff04}
    \end{table*}

    \begin{table*}
        \centering
        \caption{Per-scene performance of various NeRF based models with two input views on RealEstate-10K dataset.
        The five rows show LPIPS, SSIM, PSNR, Depth MAE and Depth SROCC scores, respectively.
        The values within parenthesis show unmasked scores.}
        \begin{tabular}{l|c|c|c|c|c|c}
            \hline
            \textbf{Model \textbackslash \ Scene Name} & \textbf{0} & \textbf{1} & \textbf{3} & \textbf{4}  & \textbf{6}  & \textbf{Average} \\
            \hline
            InfoNeRF     & \makecell{ 0.5711(0.6035)\\ 0.4396(0.4356)\\ 11.70(11.68)\\ 2.2791(2.2552)\\ -0.0335(-0.0449)} & \makecell{ 0.4592(0.5016)\\ 0.6275(0.6339)\\ 16.20(15.61)\\ 2.0442(2.0017)\\ 0.1286(0.1617)} & \makecell{ 0.5959(0.6686)\\ 0.2549(0.2469)\\ 10.66(10.64)\\ 2.8117(2.9467)\\ 0.0727(0.0663)} & \makecell{ 0.7184(0.7605)\\ 0.3873(0.3936)\\ 10.70(10.83)\\ 2.3704(2.3495)\\ 0.0867(0.0712)} & \makecell{ 0.6175(0.6580)\\ 0.4616(0.4614)\\ 12.10(12.09)\\ 1.3912(1.6040)\\ 0.2014(0.2166)} & \makecell{ 0.5924(0.6384)\\ 0.4342(0.4343)\\ 12.27(12.17)\\ 2.1793(2.2314)\\ 0.0912(0.0942)} \\
            \hline
            DietNeRF     & \makecell{ 0.4738(0.5162)\\ 0.6335(0.6295)\\ 14.80(14.72)\\ 2.2511(2.2272)\\ 0.0047(0.0032)} & \makecell{ 0.3748(0.4149)\\ 0.8289(0.8387)\\ 21.04(20.58)\\ 1.9828(1.9476)\\ 0.5266(0.5465)} & \makecell{ 0.5748(0.6718)\\ 0.3356(0.3284)\\ 12.93(12.59)\\ 2.8672(3.0125)\\ 0.2821(0.3065)} & \makecell{ 0.4712(0.4986)\\ 0.6972(0.7016)\\ 18.47(18.50)\\ 2.0381(2.0190)\\ -0.1165(-0.0861)} & \makecell{ 0.2956(0.3296)\\ 0.7718(0.7617)\\ 23.05(22.74)\\ 0.9844(1.1973)\\ 0.4725(0.4559)} & \makecell{ 0.4381(0.4862)\\ 0.6534(0.6520)\\ 18.06(17.83)\\ 2.0247(2.0807)\\ 0.2339(0.2452)} \\
            \hline
            RegNeRF      & \makecell{ 0.3357(0.3480)\\ 0.6175(0.6010)\\ 16.77(16.51)\\ -- \\ -0.0420(-0.0269)} & \makecell{ 0.2898(0.3179)\\ 0.8226(0.8290)\\ 21.62(21.04)\\ -- \\ -0.3059(-0.2176)} & \makecell{ 0.4428(0.4944)\\ 0.3103(0.2973)\\ 14.16(13.88)\\ -- \\ 0.1796(0.1485)} & \makecell{ 0.4969(0.5390)\\ 0.6177(0.6148)\\ 17.46(17.13)\\ -- \\ 0.2285(0.1764)} & \makecell{ 0.4992(0.5421)\\ 0.5902(0.5901)\\ 15.68(15.79)\\ -- \\ 0.4986(0.4781)} & \makecell{ 0.4129(0.4483)\\ 0.5916(0.5864)\\ 17.14(16.87)\\ -- \\ 0.1118(0.1117)} \\
\hline
        \end{tabular}
        \label{tab:quantitative-scene-wise-simplenerf-realestate01}
    \end{table*}

    \begin{table*}
        \centering
        \caption{Per-scene performance of various NeRF based models with three input views on RealEstate-10K dataset.
        The five rows show LPIPS, SSIM, PSNR, Depth MAE and Depth SROCC scores, respectively.
        The values within parenthesis show unmasked scores.}
        \begin{tabular}{l|c|c|c|c|c|c}
            \hline
            \textbf{Model \textbackslash \ Scene Name} & \textbf{0} & \textbf{1} & \textbf{3} & \textbf{4}  & \textbf{6}  & \textbf{Average} \\
            \hline
            InfoNeRF     & \makecell{ 0.6722(0.6983)\\ 0.3622(0.3593)\\ 9.74(9.73)\\ 2.3079(2.2937)\\ 0.0640(0.0560)} & \makecell{ 0.6151(0.6280)\\ 0.4916(0.4871)\\ 12.08(11.75)\\ 2.1419(2.1368)\\ 0.3196(0.3379)} & \makecell{ 0.6187(0.6621)\\ 0.1892(0.1848)\\ 9.08(9.23)\\ 2.8991(3.0636)\\ 0.0813(0.0884)} & \makecell{ 0.7524(0.7785)\\ 0.3749(0.3790)\\ 10.19(10.28)\\ 2.3476(2.3454)\\ 0.1210(0.1273)} & \makecell{ 0.6220(0.6562)\\ 0.4779(0.4797)\\ 11.74(11.87)\\ 1.4028(1.5757)\\ 0.3788(0.3875)} & \makecell{ 0.6561(0.6846)\\ 0.3792(0.3780)\\ 10.57(10.57)\\ 2.2198(2.2830)\\ 0.1929(0.1994)} \\
            \hline
            DietNeRF     & \makecell{ 0.4698(0.4893)\\ 0.6226(0.6166)\\ 16.44(16.36)\\ 2.1863(2.1710)\\ 0.0856(0.0728)} & \makecell{ 0.4979(0.5115)\\ 0.8046(0.8158)\\ 18.51(18.29)\\ 2.0866(2.0870)\\ -0.1136(-0.0664)} & \makecell{ 0.5759(0.6289)\\ 0.3235(0.3233)\\ 13.44(13.30)\\ 2.8594(3.0383)\\ 0.3096(0.3401)} & \makecell{ 0.4600(0.4719)\\ 0.7144(0.7151)\\ 18.98(18.95)\\ 1.9957(1.9918)\\ -0.2370(-0.2100)} & \makecell{ 0.3142(0.3413)\\ 0.7630(0.7514)\\ 22.65(22.54)\\ 1.0496(1.2234)\\ 0.0754(0.0823)} & \makecell{ 0.4636(0.4886)\\ 0.6456(0.6445)\\ 18.01(17.89)\\ 2.0355(2.1023)\\ 0.0240(0.0438)} \\
            \hline
            RegNeRF      & \makecell{ 0.3877(0.3952)\\ 0.6084(0.5950)\\ 16.22(15.99)\\ -- \\ -0.0854(-0.0849)} & \makecell{ 0.3179(0.3220)\\ 0.8178(0.8237)\\ 21.06(20.89)\\ -- \\ 0.2210(0.2376)} & \makecell{ 0.4885(0.5302)\\ 0.3008(0.2893)\\ 14.07(13.87)\\ -- \\ -0.0298(-0.0517)} & \makecell{ 0.5406(0.5635)\\ 0.6227(0.6221)\\ 17.61(17.60)\\ -- \\ 0.3273(0.2715)} & \makecell{ 0.3506(0.3699)\\ 0.7162(0.7087)\\ 20.35(20.28)\\ -- \\ -0.1459(-0.1349)} & \makecell{ 0.4171(0.4362)\\ 0.6132(0.6078)\\ 17.86(17.73)\\ -- \\ 0.0574(0.0475)} \\
\hline
        \end{tabular}
        \label{tab:quantitative-scene-wise-simplenerf-realestate02}
    \end{table*}

    \begin{table*}
        \centering
        \caption{Per-scene performance of various NeRF based models with four input views on RealEstate-10K dataset.
        The five rows show LPIPS, SSIM, PSNR, Depth MAE and Depth SROCC scores, respectively.
        The values within parenthesis show unmasked scores.}
        \begin{tabular}{l|c|c|c|c|c|c}
            \hline
            \textbf{Model \textbackslash \ Scene Name} & \textbf{0} & \textbf{1} & \textbf{3} & \textbf{4}  & \textbf{6}  & \textbf{Average} \\
            \hline
            InfoNeRF     & \makecell{ 0.5948(0.6030)\\ 0.5023(0.5002)\\ 12.80(12.74)\\ 2.2648(2.2692)\\ 0.2109(0.2053)} & \makecell{ 0.7054(0.6927)\\ 0.3621(0.3621)\\ 8.76(8.60)\\ 2.0969(2.1482)\\ 0.2990(0.3040)} & \makecell{ 0.6889(0.6976)\\ 0.1767(0.1725)\\ 8.68(8.71)\\ 2.8542(3.0882)\\ 0.2874(0.2812)} & \makecell{ 0.7918(0.8068)\\ 0.3119(0.3176)\\ 8.76(8.79)\\ 2.3608(2.3658)\\ 0.1548(0.1696)} & \makecell{ 0.5447(0.5604)\\ 0.5683(0.5625)\\ 14.12(14.12)\\ 1.3605(1.4993)\\ 0.3227(0.3369)} & \makecell{ 0.6651(0.6721)\\ 0.3843(0.3830)\\ 10.62(10.59)\\ 2.1874(2.2742)\\ 0.2549(0.2594)} \\
            \hline
            DietNeRF     & \makecell{ 0.5095(0.5189)\\ 0.6163(0.6125)\\ 15.78(15.63)\\ 2.2313(2.2346)\\ 0.0064(0.0055)} & \makecell{ 0.4664(0.4628)\\ 0.8352(0.8399)\\ 19.91(19.64)\\ 2.0104(2.0647)\\ 0.1436(0.1283)} & \makecell{ 0.6182(0.6399)\\ 0.3243(0.3198)\\ 13.29(13.27)\\ 2.7825(3.0206)\\ 0.3925(0.3897)} & \makecell{ 0.4881(0.4966)\\ 0.7157(0.7156)\\ 19.06(19.01)\\ 1.9930(1.9969)\\ -0.1838(-0.1637)} & \makecell{ 0.3442(0.3589)\\ 0.7598(0.7495)\\ 21.99(21.93)\\ 1.1820(1.3199)\\ 0.1363(0.1458)} & \makecell{ 0.4853(0.4954)\\ 0.6503(0.6475)\\ 18.01(17.89)\\ 2.0398(2.1273)\\ 0.0990(0.1011)} \\
            \hline
            RegNeRF      & \makecell{ 0.4215(0.4252)\\ 0.5922(0.5824)\\ 16.29(16.09)\\ -- \\ -0.0032(-0.0018)} & \makecell{ 0.3498(0.3461)\\ 0.8249(0.8256)\\ 21.14(20.98)\\ -- \\ 0.2630(0.2456)} & \makecell{ 0.5737(0.5903)\\ 0.3036(0.2936)\\ 13.93(13.91)\\ -- \\ 0.2565(0.2652)} & \makecell{ 0.5450(0.5581)\\ 0.6484(0.6479)\\ 18.49(18.48)\\ -- \\ 0.4133(0.4104)} & \makecell{ 0.2678(0.2719)\\ 0.7595(0.7497)\\ 21.86(21.78)\\ -- \\ -0.2187(-0.2214)} & \makecell{ 0.4316(0.4383)\\ 0.6257(0.6198)\\ 18.34(18.25)\\ -- \\ 0.1422(0.1396)} \\
\hline
        \end{tabular}
        \label{tab:quantitative-scene-wise-simplenerf-realestate03}
    \end{table*}

    \begin{table*}
        \centering
        \caption{Per-scene performance of Simple-NeRF ablated models with two input views on RealEstate-10K dataset.
        The five rows show LPIPS, SSIM, PSNR, Depth MAE and Depth SROCC scores, respectively.
        The values within parenthesis show unmasked scores.}
        \begin{tabular}{l|c|c|c|c|c|c}
            \hline
            \textbf{Model \textbackslash \ Scene Name} & \textbf{0} & \textbf{1} & \textbf{3} & \textbf{4}  & \textbf{6}  & \textbf{Average} \\
            \hline
            Simple-NeRF   & \makecell{ \textbf{0.0276}(\textbf{0.0369})\\ \textbf{0.9662}(\textbf{0.9532})\\ 34.94(31.89)\\ \textbf{0.3299}(\textbf{0.3764})\\ \textbf{0.8490}(\textbf{0.8007})} & \makecell{ \textbf{0.0273}(\textbf{0.0392})\\ \textbf{0.9823}(\textbf{0.9728})\\ \textbf{38.36}(\textbf{33.80})\\ \textbf{0.2283}(\textbf{0.3202})\\ \textbf{0.9733}(\textbf{0.9309})} & \makecell{ 0.1913(0.2130)\\ 0.5968(0.5590)\\ 19.23(18.65)\\ 0.7286(\textbf{1.0194})\\ 0.9414(0.8582)} & \makecell{ 0.0300(0.0336)\\ 0.9592(0.9511)\\ 38.51(34.93)\\ \textbf{0.0730}(\textbf{0.0960})\\ \textbf{0.9624}(\textbf{0.9363})} & \makecell{ \textbf{0.0413}(\textbf{0.0500})\\ \textbf{0.9663}(\textbf{0.9553})\\ 34.47(32.24)\\ \textbf{0.2747}(\textbf{0.4803})\\ \textbf{0.8812}(\textbf{0.8647})} & \makecell{ \textbf{0.0635}(\textbf{0.0745})\\ \textbf{0.8942}(\textbf{0.8783})\\ \textbf{33.10}(30.30)\\ \textbf{0.3269}(\textbf{0.4584})\\ \textbf{0.9215}(\textbf{0.8781})} \\
            \hline
            \makecell[l]{Simple-NeRF w/o \\ Smoothing Augmentation} & \makecell{ 0.0334(0.0426)\\ 0.9618(0.9487)\\ 34.26(31.31)\\ 0.3632(0.4080)\\ 0.8279(0.7849)} & \makecell{ 0.0387(0.0543)\\ 0.9768(0.9672)\\ 37.15(32.80)\\ 0.3145(0.3992)\\ 0.9211(0.8817)} & \makecell{ 0.1988(0.2215)\\ 0.5927(0.5540)\\ 19.28(18.76)\\ 0.7602(1.0816)\\ 0.9074(0.7931)} & \makecell{ 0.0315(0.0365)\\ 0.9582(0.9492)\\ 38.27(34.50)\\ 0.0862(0.1146)\\ 0.9601(0.9296)} & \makecell{ 0.0736(0.0898)\\ 0.9533(0.9418)\\ 33.75(31.87)\\ 0.3733(0.5510)\\ 0.8699(0.8540)} & \makecell{ 0.0752(0.0889)\\ 0.8886(0.8722)\\ 32.54(29.85)\\ 0.3795(0.5109)\\ 0.8973(0.8487)} \\
\hline
        \end{tabular}
        \label{tab:quantitative-scene-wise-simplenerf-realestate04}
    \end{table*}
    \clearpage

    \begin{table*}
        \centering
        \caption{Per-scene performance of various TensoRF based models with three input views on LLFF dataset.
        The five rows show LPIPS, SSIM, PSNR, Depth MAE and Depth SROCC scores, respectively.
        The values within parenthesis show unmasked scores.}
        \begin{tabular}{l|c|c|c|c|c|c|c|c|c}
            \hline
            \textbf{\makecell{Model \textbackslash \\ Scene Name}} & \textbf{Fern} & \textbf{Flower} & \textbf{Fortress} & \textbf{Horns} & \textbf{Leaves} & \textbf{Orchids} & \textbf{Room} & \textbf{Trex} & \textbf{Average} \\
            \hline
            TensoRF      & \makecell{ 0.61(0.66)\\ 0.28(0.26)\\ 12.4(12.4)\\ 0.47(0.56)\\ 0.13(0.12)} & \makecell{ 0.47(0.52)\\ 0.37(0.33)\\ 14.4(14.2)\\ 0.59(0.70)\\ -0.18(-0.10)} & \makecell{ 0.46(0.51)\\ 0.37(0.38)\\ 14.4(13.8)\\ 0.67(0.88)\\ 0.05(0.05)} & \makecell{ 0.55(0.64)\\ 0.25(0.25)\\ 11.7(11.6)\\ 0.84(1.02)\\ 0.02(0.03)} & \makecell{ 0.43(0.57)\\ 0.20(0.14)\\ 11.8(11.3)\\ 0.74(0.79)\\ -0.04(-0.02)} & \makecell{ 0.56(0.60)\\ 0.23(0.17)\\ 11.8(11.8)\\ 0.54(0.75)\\ -0.08(-0.09)} & \makecell{ 0.66(0.70)\\ 0.50(0.51)\\ 12.2(12.2)\\ 0.79(0.81)\\ -0.00(-0.00)} & \makecell{ 0.62(0.70)\\ 0.28(0.27)\\ 10.3(10.5)\\ 0.54(0.58)\\ 0.23(0.21)} & \makecell{ 0.55(0.62)\\ 0.32(0.30)\\ 12.3(12.1)\\ 0.67(0.78)\\ 0.03(0.03)} \\
            \hline
            DS-TensoRF   & \makecell{ 0.40(0.46)\\ 0.47(0.44)\\ 17.8(17.5)\\ \textbf{0.19}(\textbf{0.27})\\ \textbf{0.55}(\textbf{0.49})} & \makecell{ 0.24(0.30)\\ 0.68(0.59)\\ 20.8(18.8)\\ 0.19(0.32)\\ 0.80(0.59)} & \makecell{ 0.13(0.17)\\ 0.80(0.76)\\ 23.3(21.2)\\ 0.12(0.33)\\ 0.92(0.85)} & \makecell{ 0.29(0.38)\\ 0.62(0.55)\\ 17.2(15.8)\\ 0.31(0.49)\\ 0.62(0.55)} & \makecell{ 0.31(0.46)\\ 0.37(0.28)\\ 14.6(13.3)\\ 0.47(0.55)\\ 0.54(0.41)} & \makecell{ 0.47(0.52)\\ 0.35(0.26)\\ 14.8(14.2)\\ 0.30(0.47)\\ 0.57(0.49)} & \makecell{ 0.31(0.33)\\ 0.79(0.77)\\ 19.8(19.3)\\ 0.20(0.23)\\ 0.80(0.83)} & \makecell{ 0.28(0.34)\\ 0.70(0.65)\\ 18.4(17.5)\\ 0.12(0.17)\\ 0.85(0.77)} & \makecell{ 0.29(0.35)\\ 0.63(0.57)\\ 18.6(17.4)\\ 0.23(0.35)\\ 0.73(0.65)} \\
\hline
        \end{tabular}
        \label{tab:quantitative-scene-wise-simpletensorf-llff01}
    \end{table*}

    \begin{table*}
        \centering
        \caption{Per-scene performance of various TensoRF based models with three input views on RealEstate-10K dataset.
        The five rows show LPIPS, SSIM, PSNR, Depth MAE and Depth SROCC scores, respectively.
        The values within parenthesis show unmasked scores.}
        \begin{tabular}{l|c|c|c|c|c|c}
            \hline
            \textbf{Model \textbackslash \ Scene Name} & \textbf{0} & \textbf{1} & \textbf{3} & \textbf{4}  & \textbf{6}  & \textbf{Average} \\
            \hline
            TensoRF      & \makecell{ 0.0385(0.0407)\\ 0.9580(0.9428)\\ 32.66(29.91)\\ 0.2546(0.2708)\\ 0.6713(0.6488)} & \makecell{ 0.0498(0.0563)\\ 0.9709(0.9642)\\ 32.19(30.83)\\ 0.3033(0.3398)\\ 0.8540(0.8215)} & \makecell{ 0.3072(0.3196)\\ 0.4348(0.4228)\\ 16.16(15.91)\\ 0.6816(0.7829)\\ 0.3840(0.3944)} & \makecell{ 0.0590(0.0628)\\ 0.9378(0.9279)\\ 31.18(29.52)\\ 0.0969(0.1215)\\ 0.8204(0.7474)} & \makecell{ \textbf{0.0386}(\textbf{0.0458})\\ \textbf{0.9644}(\textbf{0.9559})\\ \textbf{35.90}(\textbf{33.81})\\ 0.8604(0.9056)\\ 0.4272(0.4263)} & \makecell{ 0.0986(0.1050)\\ 0.8532(0.8427)\\ 29.62(28.00)\\ 0.4394(0.4841)\\ 0.6314(0.6077)} \\
            \hline
            DS-TensoRF   & \makecell{ \textbf{0.0333}(\textbf{0.0390})\\ \textbf{0.9616}(\textbf{0.9476})\\ \textbf{34.15}(\textbf{31.09})\\ \textbf{0.1580}(\textbf{0.1822})\\ \textbf{0.7528}(\textbf{0.7036})} & \makecell{ \textbf{0.0303}(\textbf{0.0391})\\ \textbf{0.9801}(0.9731)\\ 38.01(\textbf{34.85})\\ 0.2152(0.2471)\\ 0.9446(\textbf{0.9348})} & \makecell{ 0.2288(0.2480)\\ 0.5777(0.5531)\\ 18.79(18.29)\\ 0.4200(0.5451)\\ 0.7748(0.7419)} & \makecell{ 0.0346(0.0381)\\ 0.9533(0.9464)\\ 36.35(33.48)\\ \textbf{0.0363}(0.0516)\\ \textbf{0.9258}(0.8734)} & \makecell{ 0.0423(0.0489)\\ 0.9633(0.9538)\\ 35.21(33.32)\\ 0.5304(0.6343)\\ 0.3654(0.3596)} & \makecell{ 0.0739(0.0827)\\ 0.8872(0.8748)\\ 32.50(30.20)\\ 0.2720(0.3321)\\ 0.7527(0.7227)} \\
\hline
        \end{tabular}
        \label{tab:quantitative-scene-wise-simpletensorf-realestate01}
    \end{table*}
    \clearpage

    \begin{table*}
        \centering
        \caption{Per-scene performance of various ZipNeRF based models with twelve input views on the MipNeRF360 dataset.
        The five rows show LPIPS, SSIM, PSNR, Depth MAE and Depth SROCC scores, respectively.}
        \begin{tabular}{l|c|c|c|c|c|c|c|c}
            \hline
            \textbf{Model \textbackslash \ Scene Name} & \textbf{Bicycle} & \textbf{Bonsai} & \textbf{Counter} & \textbf{Garden}  & \textbf{Kitchen} & \textbf{Room} & \textbf{Stump} & \textbf{Average} \\
            \hline
            ZipNeRF      & \makecell{0.6617\\0.2044\\14.77\\\textbf{8.4860}\\-0.3807} & \makecell{0.5722\\0.5486\\15.79\\\textbf{2.9808}\\0.4405} & \makecell{0.7258\\0.4330\\12.41\\15.1470\\-0.0017} & \makecell{\textbf{0.4130}\\0.4385\\18.65\\1.5437\\\textbf{0.8949}} & \makecell{0.5927\\0.4072\\14.05\\15.9720\\-0.0473} & \makecell{0.3978\\0.7121\\18.65\\\textbf{0.4371}\\\textbf{0.7352}} & \makecell{0.6249\\0.2591\\17.19\\8.7365\\\textbf{0.0952}} & \makecell{0.5614\\0.4616\\15.86\\7.4260\\0.2755} \\
            \hline
            InstantSplat & \makecell{\textbf{0.5135}\\\textbf{0.2955}\\\textbf{16.97}\\ -- \\ -- } & \makecell{0.5791\\0.5101\\14.64\\ -- \\ -- } & \makecell{\textbf{0.4629}\\0.5863\\17.60\\ -- \\ -- } & \makecell{0.5373\\0.2275\\15.51\\ -- \\ -- } & \makecell{\textbf{0.2515}\\\textbf{0.7314}\\\textbf{20.53}\\ -- \\ -- } & \makecell{0.4105\\0.6702\\18.45\\ -- \\ -- } & \makecell{0.4320\\0.3852\\19.64\\ -- \\ -- } & \makecell{0.4503\\0.5204\\17.57\\ -- \\ -- } \\
            \hline
            CoR-GS       & \makecell{0.8728\\0.2300\\13.48\\ -- \\ -- } & \makecell{0.4922\\0.6409\\18.86\\ -- \\ -- } & \makecell{0.4684\\0.6529\\18.59\\ -- \\ -- } & \makecell{0.4591\\0.4926\\18.99\\ -- \\ -- } & \makecell{0.3128\\0.7175\\18.61\\ -- \\ -- } & \makecell{0.3925\\0.7236\\19.65\\ -- \\ -- } & \makecell{0.8008\\0.3518\\18.93\\ -- \\ -- } & \makecell{0.5039\\0.5834\\18.32\\ -- \\ -- } \\
\hline
            \end{tabular}
        \label{tab:quantitative-scene-wise-simplezipnerf-mipnerf360-01}
    \end{table*}

    \begin{table*}
        \centering
        \caption{Per-scene performance of various ZipNeRF based models with twenty input views on the MipNeRF360 dataset.
        The five rows show LPIPS, SSIM, PSNR, Depth MAE and Depth SROCC scores, respectively.}
        \begin{tabular}{l|c|c|c|c|c|c|c|c}
            \hline
            \textbf{Model \textbackslash \ Scene Name} & \textbf{Bicycle} & \textbf{Bonsai} & \textbf{Counter} & \textbf{Garden}  & \textbf{Kitchen} & \textbf{Room} & \textbf{Stump} & \textbf{Average} \\
            \hline
            ZipNeRF      & \makecell{0.6554\\0.2293\\14.97\\\textbf{6.1012}\\-0.1092} & \makecell{0.3793\\0.6912\\18.43\\5.3864\\0.6476} & \makecell{0.7201\\0.4716\\13.45\\2.1826\\0.0093} & \makecell{0.2472\\0.6358\\\textbf{22.49}\\\textbf{0.5384}\\\textbf{0.9838}} & \makecell{0.2799\\0.7406\\20.23\\\textbf{0.9845}\\0.7730} & \makecell{0.3114\\0.7889\\23.15\\\textbf{0.3474}\\0.8285} & \makecell{0.6076\\0.2732\\17.58\\\textbf{5.5524}\\\textbf{-0.0517}} & \makecell{0.4350\\0.5911\\18.89\\\textbf{2.7528}\\0.5032} \\
            \hline
            InstantSplat & \makecell{0.6465\\0.2893\\16.30\\ -- \\ -- } & \makecell{0.7485\\0.4929\\13.86\\ -- \\ -- } & \makecell{0.4544\\0.6418\\18.04\\ -- \\ -- } & \makecell{0.4018\\0.3837\\18.49\\ -- \\ -- } & \makecell{0.2525\\0.7423\\20.53\\ -- \\ -- } & \makecell{0.4018\\0.7290\\20.02\\ -- \\ -- } & \makecell{0.4729\\0.3670\\19.15\\ -- \\ -- } & \makecell{0.4816\\0.5545\\18.02\\ -- \\ -- } \\
            \hline
            CoR-GS       & \makecell{0.6851\\0.3991\\19.57\\ -- \\ -- } & \makecell{0.2874\\\textbf{0.8061}\\\textbf{24.03}\\ -- \\ -- } & \makecell{\textbf{0.3047}\\\textbf{0.7622}\\\textbf{21.23}\\ -- \\ -- } & \makecell{0.2723\\\textbf{0.6454}\\21.95\\ -- \\ -- } & \makecell{0.1875\\\textbf{0.8160}\\21.91\\ -- \\ -- } & \makecell{\textbf{0.2819}\\\textbf{0.8124}\\\textbf{23.50}\\ -- \\ -- } & \makecell{0.6948\\0.3730\\18.84\\ -- \\ -- } & \makecell{0.3500\\0.7008\\21.98\\ -- \\ -- } \\
\hline
            \end{tabular}
        \label{tab:quantitative-scene-wise-simplezipnerf-mipnerf360-02}
    \end{table*}

    \begin{table*}
        \centering
        \caption{Per-scene performance of various ZipNeRF based models with thirty-six input views on the MipNeRF360 dataset.
        The five rows show LPIPS, SSIM, PSNR, Depth MAE and Depth SROCC scores, respectively.}
        \begin{tabular}{l|c|c|c|c|c|c|c|c}
            \hline
            \textbf{Model \textbackslash \ Scene Name} & \textbf{Bicycle} & \textbf{Bonsai} & \textbf{Counter} & \textbf{Garden}  & \textbf{Kitchen} & \textbf{Room} & \textbf{Stump} & \textbf{Average} \\
            \hline
            ZipNeRF      & \makecell{0.5717\\0.2810\\16.09\\13.6503\\0.0187} & \makecell{0.1689\\0.8687\\25.54\\2.6188\\\textbf{0.9025}} & \makecell{0.7272\\0.4546\\13.01\\\textbf{2.1970}\\0.0558} & \makecell{\textbf{0.1485}\\0.7571\\\textbf{24.52}\\0.4649\\\textbf{0.9905}} & \makecell{0.1416\\0.8636\\25.15\\\textbf{0.5189}\\0.9455} & \makecell{0.2159\\0.8515\\25.56\\\textbf{0.6145}\\\textbf{0.8953}} & \makecell{0.5632\\0.2733\\17.75\\\textbf{4.6859}\\0.2225} & \makecell{0.3316\\0.6737\\21.78\\\textbf{3.0695}\\0.6353} \\
            \hline
            InstantSplat & \makecell{0.9922\\0.2500\\12.83\\ -- \\ -- } & \makecell{0.8490\\0.4673\\12.38\\ -- \\ -- } & \makecell{0.6799\\0.5957\\15.89\\ -- \\ -- } & \makecell{0.4130\\0.5020\\19.57\\ -- \\ -- } & \makecell{0.2707\\0.7523\\18.81\\ -- \\ -- } & \makecell{0.4545\\0.7097\\18.48\\ -- \\ -- } & \makecell{0.9624\\0.3138\\16.71\\ -- \\ -- } & \makecell{0.6268\\0.5461\\16.37\\ -- \\ -- } \\
            \hline
            CoR-GS       & \makecell{0.4229\\0.5395\\21.31\\ -- \\ -- } & \makecell{0.1763\\0.8873\\27.01\\ -- \\ -- } & \makecell{0.1972\\0.8444\\25.18\\ -- \\ -- } & \makecell{0.1535\\0.7612\\23.98\\ -- \\ -- } & \makecell{0.1410\\0.8661\\25.32\\ -- \\ -- } & \makecell{0.2022\\0.8707\\26.85\\ -- \\ -- } & \makecell{0.4133\\0.5440\\21.84\\ -- \\ -- } & \makecell{0.2239\\0.7907\\24.98\\ -- \\ -- } \\
\hline
            \end{tabular}
        \label{tab:quantitative-scene-wise-simplezipnerf-mipnerf360-03}
    \end{table*}

    \begin{table*}
        \centering
        \caption{Per-scene performance of various ZipNeRF based models with four input views on NeRF-Synthetic dataset.
        The five rows show LPIPS, SSIM, and PSNR, respectively.}
        \begin{tabular}{l|c|c|c|c|c|c|c|c|c}
            \hline
            \textbf{Model \textbackslash \ Scene Name} & \textbf{Chair} & \textbf{Drums} & \textbf{Ficus} & \textbf{Hotdog}  & \textbf{Lego} & \textbf{Materials} & \textbf{Mic} & \textbf{Ship} & \textbf{Average} \\
            \hline
            ZipNeRF      & \makecell{0.3172\\0.8372\\14.12} & \makecell{0.4762\\\textbf{0.7501}\\10.75} & \makecell{0.2500\\0.8430\\14.47} & \makecell{0.4280\\0.7869\\11.33} & \makecell{0.5128\\0.6811\\9.39} & \makecell{0.4689\\0.7340\\8.63} & \makecell{0.3186\\0.8562\\13.20} & \makecell{0.6389\\0.5496\\6.43} & \makecell{0.4263\\0.7548\\11.04} \\
            \hline
            Simple-ZipNeRF & \makecell{\textbf{0.2848}\\\textbf{0.8454}\\\textbf{14.54}} & \makecell{\textbf{0.4635}\\0.7446\\\textbf{11.28}} & \makecell{\textbf{0.2114}\\\textbf{0.8500}\\\textbf{15.17}} & \makecell{\textbf{0.4083}\\\textbf{0.8087}\\\textbf{11.38}} & \makecell{\textbf{0.5021}\\\textbf{0.6867}\\\textbf{9.40}} & \makecell{\textbf{0.4085}\\\textbf{0.7592}\\\textbf{9.05}} & \makecell{\textbf{0.2903}\\\textbf{0.8609}\\\textbf{14.63}} & \makecell{\textbf{0.5331}\\\textbf{0.6162}\\\textbf{6.54}} & \makecell{\textbf{0.3878}\\\textbf{0.7715}\\\textbf{11.50}} \\
            \hline
            \end{tabular}
        \label{tab:quantitative-scene-wise-simplezipnerf-nerfsynthetic01}
    \end{table*}

    \begin{table*}
        \centering
        \caption{Per-scene performance of various ZipNeRF based models with eight input views on NeRF-Synthetic dataset.
        The five rows show LPIPS, SSIM, and PSNR, respectively.}
        \begin{tabular}{l|c|c|c|c|c|c|c|c|c}
            \hline
            \textbf{Model \textbackslash \ Scene Name} & \textbf{Chair} & \textbf{Drums} & \textbf{Ficus} & \textbf{Hotdog}  & \textbf{Lego} & \textbf{Materials} & \textbf{Mic} & \textbf{Ship} & \textbf{Average} \\
            \hline
            ZipNeRF      & \makecell{0.2227\\0.8471\\16.86} & \makecell{0.3508\\0.7706\\11.90} & \makecell{\textbf{0.0813}\\\textbf{0.8980}\\\textbf{21.91}} & \makecell{0.2773\\0.8391\\14.21} & \makecell{\textbf{0.4562}\\\textbf{0.6991}\\\textbf{9.72}} & \makecell{0.4464\\0.7388\\8.67} & \makecell{0.0453\\\textbf{0.9571}\\\textbf{26.99}} & \makecell{\textbf{0.4216}\\\textbf{0.6282}\\\textbf{9.82}} & \makecell{0.2877\\0.7973\\15.01} \\
            \hline
            Simple-ZipNeRF & \makecell{\textbf{0.1552}\\\textbf{0.8687}\\\textbf{18.53}} & \makecell{\textbf{0.2572}\\\textbf{0.8084}\\\textbf{14.17}} & \makecell{0.0879\\0.8964\\21.68} & \makecell{\textbf{0.2318}\\\textbf{0.8481}\\\textbf{14.56}} & \makecell{0.4900\\0.6871\\9.59} & \makecell{\textbf{0.2418}\\\textbf{0.7801}\\\textbf{12.30}} & \makecell{\textbf{0.0405}\\0.9541\\26.87} & \makecell{0.4641\\0.6075\\9.31} & \makecell{\textbf{0.2461}\\\textbf{0.8063}\\\textbf{15.88}} \\
            \hline
            \end{tabular}
        \label{tab:quantitative-scene-wise-simplezipnerf-nerfsynthetic02}
    \end{table*}

    \begin{table*}
        \centering
        \caption{Per-scene performance of various ZipNeRF based models with twelve input views on NeRF-Synthetic dataset.
        The five rows show LPIPS, SSIM, and PSNR, respectively.}
        \begin{tabular}{l|c|c|c|c|c|c|c|c|c}
            \hline
            \textbf{Model \textbackslash \ Scene Name} & \textbf{Chair} & \textbf{Drums} & \textbf{Ficus} & \textbf{Hotdog}  & \textbf{Lego} & \textbf{Materials} & \textbf{Mic} & \textbf{Ship} & \textbf{Average} \\
            \hline
            ZipNeRF      & \makecell{0.1012\\\textbf{0.9062}\\\textbf{23.40}} & \makecell{0.2181\\0.8143\\14.72} & \makecell{\textbf{0.0474}\\\textbf{0.9336}\\\textbf{25.18}} & \makecell{0.1278\\\textbf{0.9055}\\21.09} & \makecell{0.1149\\\textbf{0.8718}\\\textbf{20.24}} & \makecell{0.1659\\0.8403\\18.52} & \makecell{0.0402\\0.9613\\27.39} & \makecell{\textbf{0.4844}\\\textbf{0.5898}\\\textbf{10.39}} & \makecell{0.1625\\0.8528\\20.12} \\
            \hline
            Simple-ZipNeRF & \makecell{\textbf{0.1010}\\0.9013\\22.27} & \makecell{\textbf{0.1458}\\\textbf{0.8475}\\\textbf{18.27}} & \makecell{0.0569\\0.9236\\24.01} & \makecell{\textbf{0.1227}\\0.9055\\\textbf{21.24}} & \makecell{\textbf{0.1116}\\0.8602\\20.12} & \makecell{\textbf{0.1123}\\\textbf{0.8617}\\\textbf{20.59}} & \makecell{\textbf{0.0342}\\\textbf{0.9614}\\\textbf{28.32}} & \makecell{0.5411\\0.5634\\9.27} & \makecell{\textbf{0.1532}\\\textbf{0.8531}\\\textbf{20.51}} \\
            \hline
            \end{tabular}
        \label{tab:quantitative-scene-wise-simplezipnerf-nerfsynthetic03}
    \end{table*}

\end{document}